\pdfoutput=1
\documentclass[11pt]{article}
\usepackage[preprint]{coling}
\usepackage{times}
\usepackage{latexsym}
\usepackage[T1]{fontenc}
\newcommand{\pink}[1]{\textcolor[rgb]{1.00,0.00,1.00}{#1}}
\usepackage[utf8]{inputenc}

\usepackage{microtype}
\usepackage[whole]{bxcjkjatype}
\usepackage{comment}

\usepackage{graphicx}
\usepackage{amsmath}
\usepackage{amssymb}
\usepackage{multirow}
\usepackage{array}
\usepackage{booktabs}
\usepackage{subcaption}
\usepackage{adjustbox}
\DeclareMathOperator{\argmax}{argmax}
\usepackage{stmaryrd}
\usepackage{siunitx}

\title{Revisiting Cosine Similarity via Normalized ICA-transformed Embeddings}

\author{
  Hiroaki Yamagiwa$^{1}$ \qquad Momose Oyama$^{1,2}$ \qquad Hidetoshi Shimodaira$^{1,2}$ \\
  ${}^1$Kyoto University \qquad ${}^2$RIKEN\\
  \texttt{hiroaki.yamagiwa@sys.i.kyoto-u.ac.jp,}\\
  \texttt{oyama.momose@sys.i.kyoto-u.ac.jp, shimo@i.kyoto-u.ac.jp}\\
}

\newcommand{\colora}[1]{\textcolor[rgb]{1.00,0.00,0.00}{#1}}
\newcommand{\colorb}[1]{\textcolor[rgb]{1.00,0.34,0.17}{#1}}
\newcommand{\colorc}[1]{\textcolor[rgb]{1.00,0.64,0.34}{#1}}
\newcommand{\colord}[1]{\textcolor[rgb]{0.83,0.87,0.50}{#1}}
\newcommand{\colore}[1]{\textcolor[rgb]{0.61,0.98,0.64}{#1}}
\newcommand{\colorf}[1]{\textcolor[rgb]{0.39,0.98,0.77}{#1}}
\newcommand{\colorg}[1]{\textcolor[rgb]{0.17,0.87,0.87}{#1}}
\newcommand{\colorh}[1]{\textcolor[rgb]{0.06,0.64,0.94}{#1}}
\newcommand{\colori}[1]{\textcolor[rgb]{0.28,0.34,0.98}{#1}}
\newcommand{\colorj}[1]{\textcolor[rgb]{0.50,0.00,1.00}{#1}}

\begin{document}
\maketitle
\begin{abstract}
Cosine similarity is widely used to measure the similarity between two embeddings, while interpretations based on angle and correlation coefficient are common. In this study, we focus on the interpretable axes of embeddings transformed by Independent Component Analysis (ICA), and propose a novel interpretation of cosine similarity as the sum of semantic similarities over axes. The normalized ICA-transformed embeddings exhibit sparsity, enhancing the interpretability of each axis, and the semantic similarity defined by the product of the components represents the shared meaning between the two embeddings along each axis. The effectiveness of this approach is demonstrated through intuitive numerical examples and thorough numerical experiments. By deriving the probability distributions that govern each component and the product of components, we propose a method for selecting statistically significant axes.
\end{abstract}

\section{Introduction}\label{sec:intro}
Cosine similarity is widely used to measure the similarity between two embeddings~\cite{DBLP:journals/tacl/BojanowskiGJM17,DBLP:conf/emnlp/ReimersG19,sitikhu2019comparison} and can be computed efficiently~\cite{DBLP:conf/ideal/LiH13,DBLP:journals/isci/XiaZL15,DBLP:conf/emnlp/GaoYC21}. 
For word embeddings, the norm represents the importance of the word and the direction represents the meaning of the word~\cite{DBLP:conf/emnlp/YokoiTASI20,DBLP:conf/emnlp/OyamaYS23}.
Therefore, cosine similarity, which is the inner product of the normalized embeddings, makes sense as word similarity. 
Studies dealing with cosine tend to focus on the angle~\cite{DBLP:conf/cvpr/DengGXZ19,DBLP:journals/corr/abs-2309-12871} or interpret it as a correlation coefficient~\cite{DBLP:journals/corr/abs-1208-3145}.

\begin{figure}[t!]
    \centering
    \includegraphics[width=0.92\columnwidth]{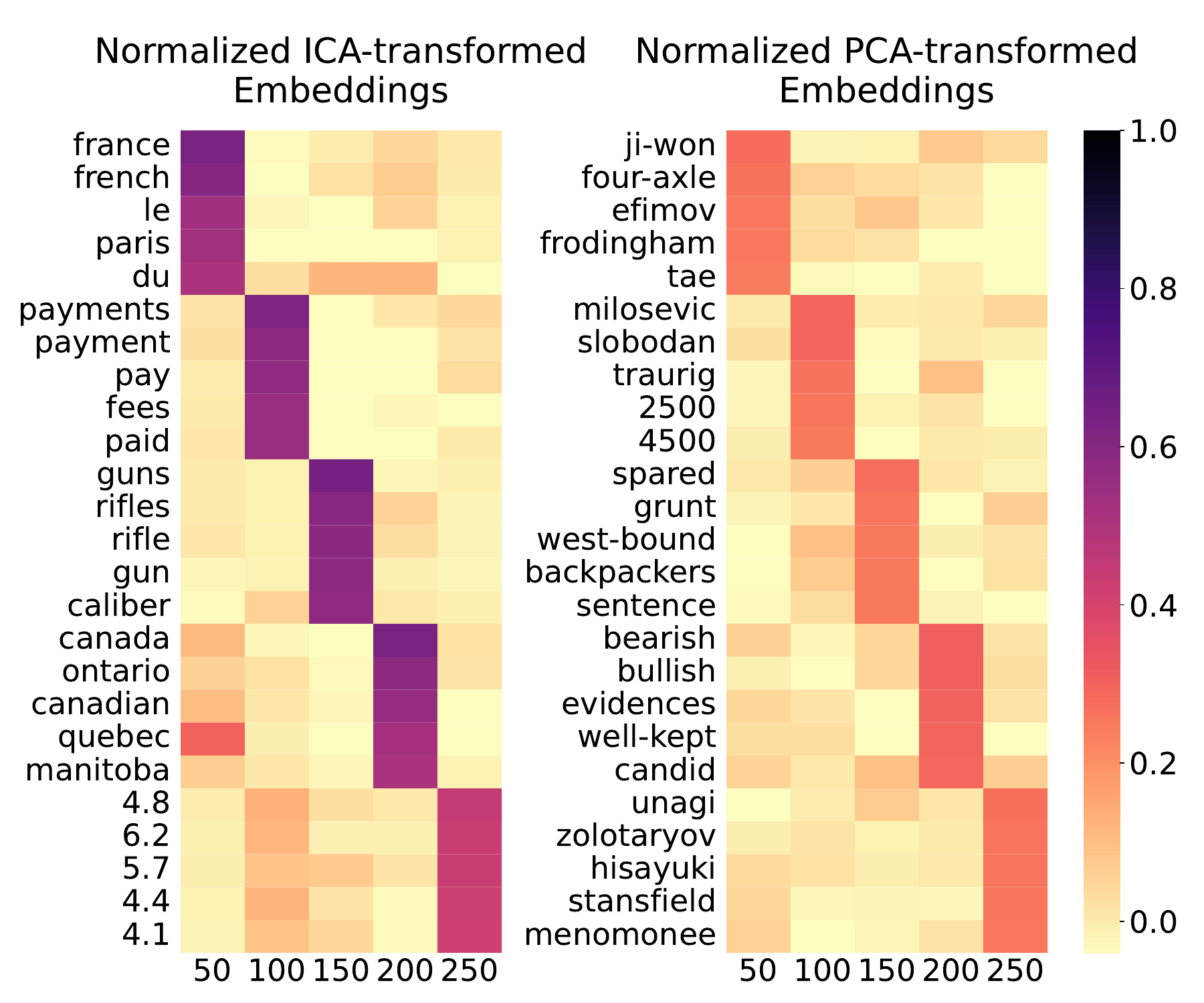}
    \caption{
Heatmaps of 300-dimensional GloVe embeddings transformed by (left) Independent Component Analysis (ICA) and (right) Principal Component Analysis (PCA), with embeddings normalized to unit length following the transformations.
We select five specific axes (50th, 100th, etc.) and display the top five words by component values for each axis.
For the normalized ICA-transformed embeddings, the maximum component values on the axes are substantial, highlighting significant features, while the remaining values are typically small, resulting in a sparse representation.
Conversely, for the normalized PCA-transformed embeddings, even the maximum values are not large, making it difficult to interpret the meanings of the axes.
}
    \label{fig:intro}
\end{figure}

\begin{figure*}[p]
\centering
\begin{subfigure}{\textwidth}
\centering
    \includegraphics[width=\textwidth]{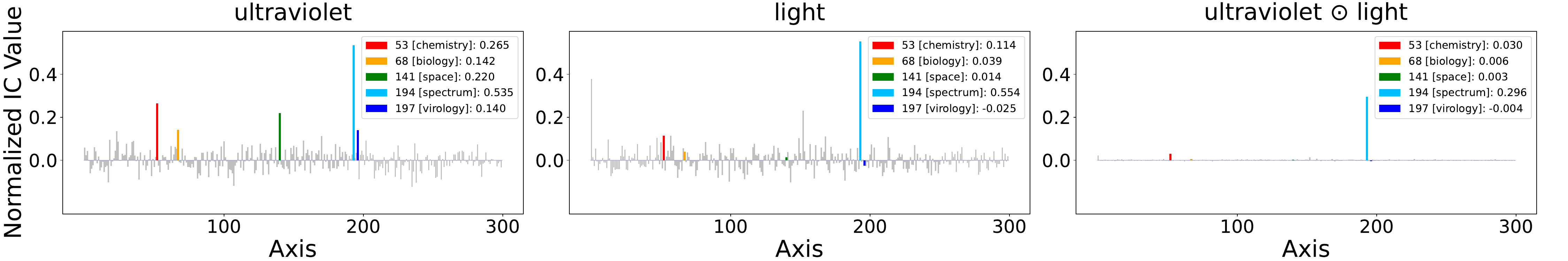}
    \subcaption{
Normalized ICA-transformed GloVe embeddings of \textit{ultraviolet} and \textit{light} and their component-wise products.
}
    \label{fig:cosine_ica}
\end{subfigure}
\par\medskip
\begin{subfigure}{\textwidth}
\centering
    \includegraphics[width=\textwidth]{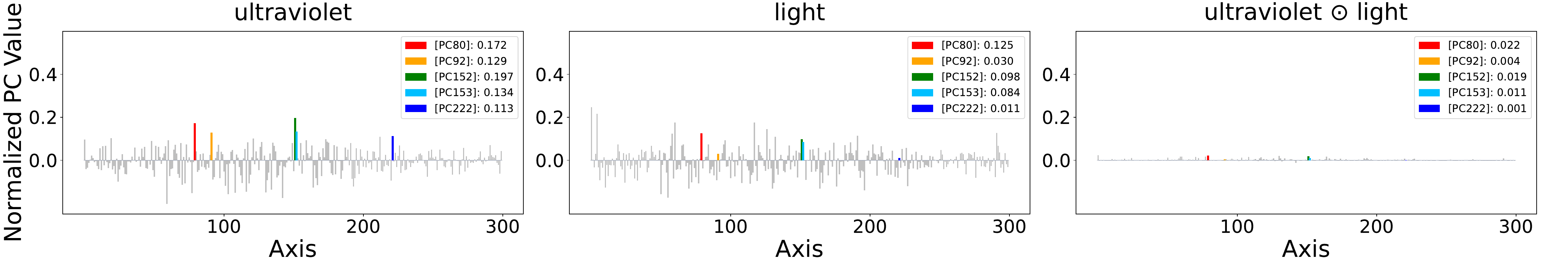}
    \subcaption{
Normalized PCA-transformed GloVe embeddings of \textit{ultraviolet} and \textit{light} and their component-wise products.
}
    \label{fig:cosine_pca}
\end{subfigure}
    \caption{
For the (a) ICA and (b) PCA transformations, bar graphs are displayed for each, plotting the component values of the normalized GloVe embeddings: (left) \textit{ultraviolet}, (middle) \textit{light}, and (right) their component-wise products.
The axes with the top five component values in the \textit{ultraviolet} embedding are highlighted, and these same axes are consistently colored across the other two plots.
For the normalized ICA-transformed embedding of \textit{ultraviolet}, the meanings of the top five axes are \textit{[chemistry]}, \textit{[biology]}, \textit{[space]}, \textit{[spectrum]}, and \textit{[virology]} in the order of their indices.
See Table~\ref{tab:intro-topwords} in Appendix~\ref{app:cosine} for the top words of the axes.
The component \textit{[spectrum]} of the normalized ICA-transformed embeddings should be much more emphasized in the component-wise products than in the component values. This is because the standard deviation of the probability distribution for the component-wise products is $1/d$, which is smaller than the standard deviation of $1/\sqrt{d}$ for the component values.
See Appendix~\ref{app:cosine} for more descriptions and Appendix~\ref{app:distribution-theory} for details of the distribution theory.
}
\label{fig:cosine}
\end{figure*}

\begin{table*}[p]
\centering
\begin{subtable}{.33\textwidth}
  \centering
  \begin{adjustbox}{max width=0.93\linewidth}
 \begin{tabular}{rcrrr}
  \toprule
  Axis & Meaning & Value & $p$-value & Bonferroni \\
  \midrule
{\color{cyan}\num{194}}  & {\color{cyan}\textit{[spectrum]}}   & \num{0.535} & \num{9.97e-21} & \num{2.99e-18} \\
{\color{red}\num{53}}    & {\color{red}\textit{[chemistry]}}   & \num{0.265} & \num{2.17e-06} & \num{6.51e-04} \\
{\color{green}\num{141}} & {\color{green}\textit{[space]}}     & \num{0.220} & \num{6.97e-05} & \num{2.09e-02} \\
{\color{orange}\num{68}} & {\color{orange}\textit{[biology]}}  & \num{0.142} & \num{6.88e-03} & \num{1.00e+00} \\
{\color{blue}\num{197}}  & {\color{blue}\textit{[virology]}}   & \num{0.140} & \num{7.53e-03} & \num{1.00e+00} \\
\bottomrule
  \end{tabular}
  \end{adjustbox}
\caption{\textit{ultraviolet}}
\end{subtable}\renewcommand{\arraystretch}{1.1}\begin{subtable}{.33\textwidth}
  \centering
  \begin{adjustbox}{max width=0.93\linewidth}
 \begin{tabular}{rcrrr}
  \toprule
  Axis & Meaning & Value & $p$-value & Bonferroni \\
  \midrule
{\color{cyan}\num{194}} & {\color{cyan}\textit{[spectrum]}} & \num{0.554} & \num{4.44e-22} & \num{1.33e-19} \\
\num{1}                 & \textit{[function words]}         & \num{0.379} & \num{2.67e-11} & \num{8.02e-09} \\
\num{153}               & \textit{[boxing]}                 & \num{0.230} & \num{3.35e-05} & \num{1.00e-02} \\
{\color{red}\num{53}}   & {\color{red}\textit{[chemistry]}} & \num{0.114} & \num{2.37e-02} & \num{1.00e+00} \\
\num{58}                & \textit{[ordinal]}                & \num{0.114} & \num{2.46e-02} & \num{1.00e+00} \\
\bottomrule
  \end{tabular}
  \end{adjustbox}
\caption{\textit{light}}
\end{subtable}
\begin{subtable}{.33\textwidth}
  \centering
  \begin{adjustbox}{max width=0.93\linewidth}
 \begin{tabular}{rcrrr}
  \toprule
  Axis & Meaning & Value & $p$-value & Bonferroni \\
  \midrule
{\color{cyan}\num{194}} & {\color{cyan}\textit{[spectrum]}} & \num{0.296} & \num{1.12e-40} & \num{3.36e-38} \\
{\color{red}\num{53}}   & {\color{red}\textit{[chemistry]}} & \num{0.030} & \num{1.38e-05} & \num{4.15e-03} \\
\num{1}                 & \textit{[function words]}         & \num{0.022} & \num{1.69e-04} & \num{5.07e-02} \\
\num{153}               & \textit{[boxing]}                 & \num{0.014} & \num{2.56e-03} & \num{7.68e-01} \\
\num{158}               & \textit{[police]}                 & \num{0.007} & \num{2.76e-02} & \num{1.00e+00} \\
\bottomrule
  \end{tabular}
  \end{adjustbox}
  \caption{\textit{ultraviolet} $\odot$ \textit{light}}
\end{subtable}
  \caption{(a, b) show the observed component values of normalized ICA-transformed GloVe embeddings in Fig.~\ref{fig:cosine_ica} and (c) shows their component-wise products.
The $p$-values and their Bonferroni-corrected values are shown for the top five axes in each table.
Refer to Section~\ref{sec:statistical-analysis-axis-selection} for details on the $p$-value calculations. Appendix~\ref{app:cosine} presents the top words of these axes and Table~\ref{tab:pvalues_pca} in Appendix~\ref{app:distribution-theory} shows the results of normalized PCA-transformed embeddings.
}
\label{tab:table2_5columns}
\end{table*}
\renewcommand{\arraystretch}{1.0} 

\begin{figure*}[p]
    \centering
    \begin{minipage}{0.33\linewidth}
        \centering
        \includegraphics[width=\linewidth]{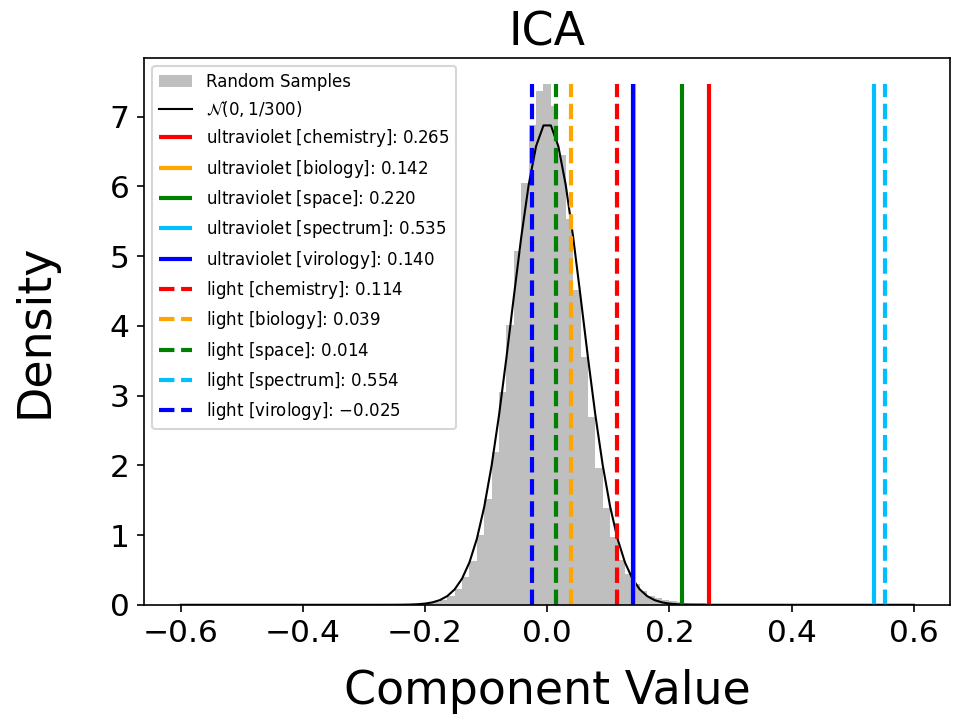}
        \subcaption{Components}
        \label{fig:comp_hist_maintext}
    \end{minipage}\hfill
    \begin{minipage}{0.33\linewidth}
        \centering
        \includegraphics[width=\linewidth]{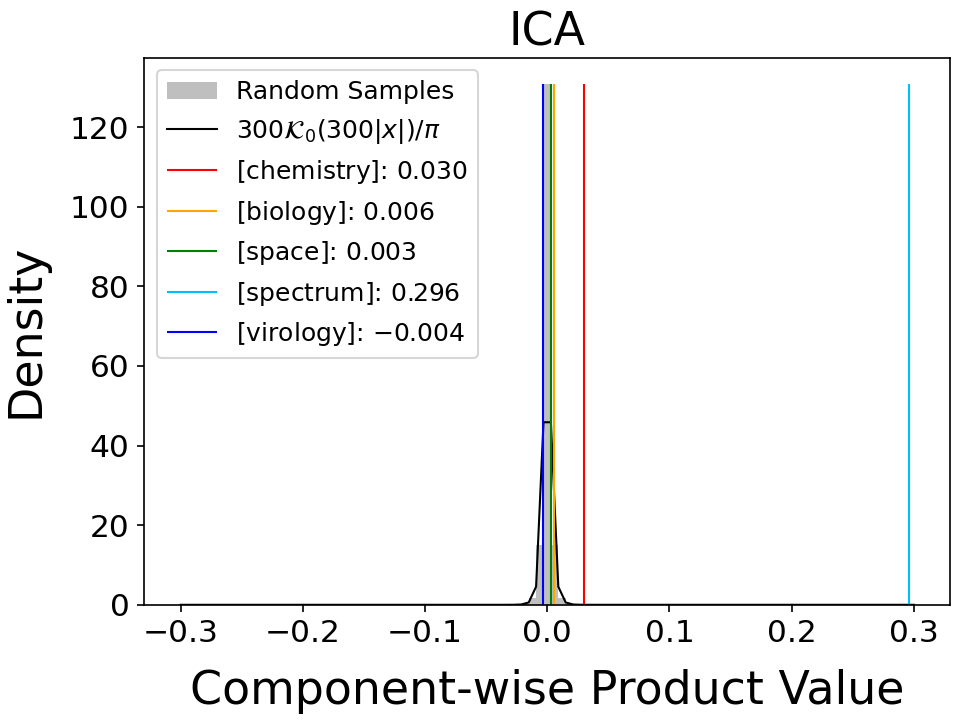}
        \subcaption{Component-wise products}
        \label{fig:comp_prod_hist_maintext}
    \end{minipage}
    \begin{minipage}{0.33\linewidth}
        \centering
        \includegraphics[width=\linewidth]{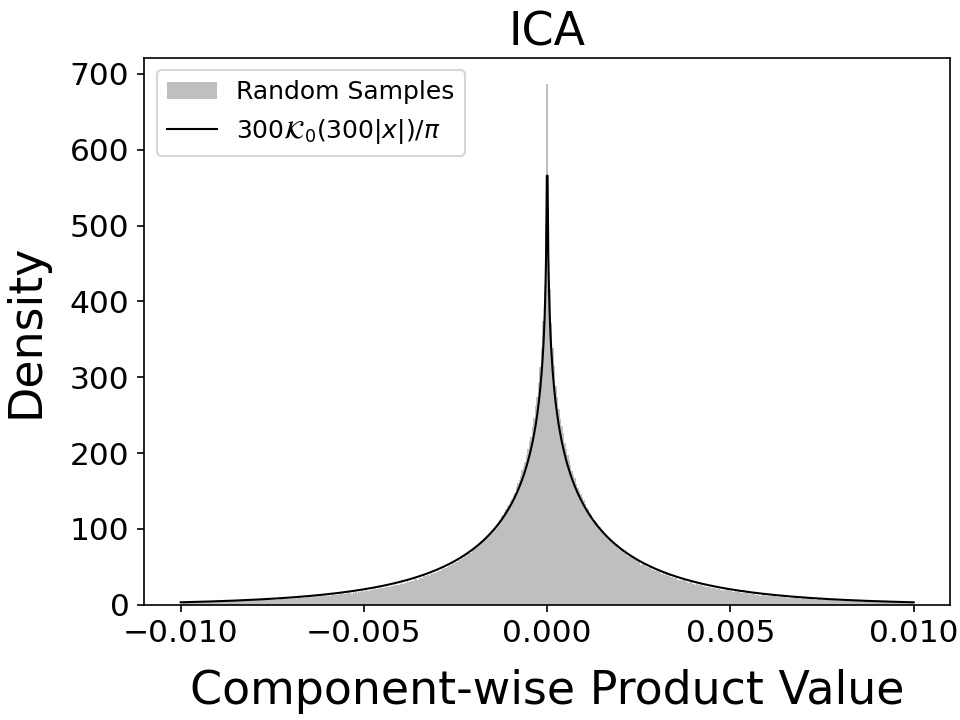}
        \subcaption{Component-wise products (magnified)}
        \label{fig:comp_prod_hist_zoom_maintext}
    \end{minipage}
    \caption{
For {10,000} randomly sampled pairs of normalized ICA-transformed GloVe embeddings 
$ \hat{\mathbf{s}},  \hat{\mathbf{s}}' \in \mathbb{R}^d$, (a)~the histogram of the components $\hat s^{(\ell)}$ and (b, c)~the histograms of the products of the components $\hat s^{(\ell)} \hat s^{(\ell)\prime}$  are displayed. 
The observed component values in Fig.~\ref{fig:cosine_ica} are also indicated as vertical lines.
Appendix~\ref{app:cosine} presents additional descriptions and Fig.~\ref{fig:histgram-pca} in Appendix~\ref{app:distribution-theory} shows the results for the normalized PCA-transformed GloVe embeddings.
The theoretical probability density of (\ref{eq:dist-component}) for the components and that of (\ref{eq:dist-product}) for the component-wise products in Appendix~\ref{app:distribution-theory} 
are almost identical to their observed histograms.
The theory in Appendix~\ref{app:distribution-theory} is also supported by the inverse of the observed variance, ${300.005}\approx d$ in (a) and $89{,}917.992\approx d^2$ in (b, c) for $d=300$.
}
    \label{fig:histgram-ica_maintext}
\end{figure*}

Unlike existing studies, our research introduces a novel interpretation of cosine similarity, focusing on embeddings transformed by Independent Component Analysis (ICA)~\cite{DBLP:journals/nn/HyvarinenO00}, which aims to maximize the independence of components.
FastICA~\cite{DBLP:journals/tnn/Hyvarinen99} is widely used as an implementation of ICA, where it further rotates the whitened embeddings from Principal Component Analysis (PCA)~\cite{hotelling1933analysis} to make the components of the embeddings closer to independent random variables.
ICA-transformed embeddings are known to have interpretable axes~\cite{marecek-etal-2020,DBLP:conf/coling/MusilM24,DBLP:conf/emnlp/YamagiwaOS23}. 
Specifically, \citet{DBLP:conf/emnlp/YamagiwaOS23} determined the meanings of the axes by examining the top words with the highest component values in the normalized ICA-transformed embeddings.
Figure~\ref{fig:intro} shows heatmaps of the normalized GloVe embeddings after ICA and PCA transformations. 
Here, as an example, the meaning of an axis is denoted as \textit{[decimals]}, and this notation applies to other axes as well.
These axes of the ICA-transformed embeddings can be interpreted as \textit{[france]}, \textit{[payments]}, \textit{[guns]}, \textit{[canada]}, and \textit{[decimals]}, whereas those of the PCA-transformed embeddings remain uninterpretable.
As demonstrated in the experiments in Section~\ref{sec:experiments}, ICA provides better interpretability than PCA, and normalization further enhances this interpretability.
Hereafter, a word will be denoted as \textit{paris}, as an example. 
For the indices of an axis and a word, we use $\ell_{\text{\textit{[decimals]}}}$ and $i_{\text{\textit{paris}}}$, respectively, as examples. The same notation applies to other axes and words.

The inner product of normalized embeddings represents the cosine similarity. 
Figure~\ref{fig:cosine} shows the normalized GloVe embeddings transformed by ICA and PCA for \textit{ultraviolet} and \textit{light}, along with their component-wise products, shown in a bar graph.
The sum of the component-wise products forms the inner product, yielding an identical cosine similarity value of $0.485$ for both transformations. This equivalence arises because the ICA embeddings are obtained by rotating the PCA embeddings.
However, a closer examination of the component-wise products shows distinct differences: In the ICA-transformed embeddings, both \textit{ultraviolet} and \textit{light} exhibit a large component value in \textit{[spectrum]}, resulting in a significant product value and sparse values elsewhere. Conversely, the PCA-transformed embeddings do not have any axes with large component values, resulting in a uniformly dense vector. These differences illustrate that, while the overall cosine similarity is the same, the underlying structural contributions to this similarity vary significantly between the two transformations.

Based on these observations, we define the semantic similarity on the $\ell$-th axis for words $w_i$ and $w_j$ as the component-wise product of the normalized ICA-transformed embeddings, denoted as $\text{sem}_\ell(w_i, w_j)$.
For example, in Fig.~\ref{fig:cosine_ica}, 
\[ \text{sem}_{\ell_{\text{\textit{[spectrum]}}}}(\text{\textit{ultraviolet}}, \text{\textit{light}})=0.296. \]
A large value of $\text{sem}_\ell(w_i, w_j)$ indicates that both $w_i$ and $w_j$ have the meaning represented by $\ell$-th axis.
The cosine similarity can be interpreted as the sum of the semantic similarities across all axes:
\begin{align}
    \cos(w_i,w_j) = \sum_{\ell=1}^d \text{sem}_\ell(w_i,w_j),\label{eq:cosine-sem}
\end{align}
which represents the ``additive compositionality'' of semantic similarities, decomposing overall similarity into the component-wise similarities.

As shown in parts (a, b) of Table~\ref{tab:table2_5columns}, selecting the axes with large component values for each word allows us to interpret the meaning of that word. Similarly, as shown in part (c) of Table~\ref{tab:table2_5columns}, selecting the axes with large $\text{sem}_\ell(w_i, w_j)$ values for a word pair allows us to interpret the meaning shared by both words.

The number of selected axes can be determined by statistical methods. As shown in Fig.~\ref{fig:histgram-ica_maintext}, we theoretically derived the probability distributions for both the component values and their products.
Therefore, as  will be discussed in Section~\ref{sec:statistical-analysis-axis-selection}, we can calculate the $p$-value to determine whether the observed values are significantly greater than zero.
We select the axes where the Bonferroni-corrected $p$-value, accounting for the multiplicity of hypothesis testing, is smaller than the significance level $\alpha$.
For example, with $\alpha = 0.05$, \textit{[spectrum]}, \textit{[chemistry]}, and \textit{[space]} are selected for \textit{ultraviolet}, while \textit{[spectrum]} and \textit{[chemistry]} are selected for the pair \textit{ultraviolet} and \textit{light}.

\section{Background: Independent Components in Embeddings}\label{sec:background}
In this section, we explain PCA and ICA transformations for embeddings based on \citet{DBLP:books/wi/HyvarinenKO01,DBLP:conf/emnlp/YamagiwaOS23}.
Let $\mathbf{X}\in\mathbb{R}^{n\times d}$ be pre-trained embeddings with vocabulary size $n$ and dimension $d$. We assume that $\mathbf{X}$ is centered, i.e., the mean of components in each column is zero.

\subsection{PCA-transformed embeddings}
PCA, typically implemented using algorithms such as SVD, transforms the embeddings so that their components align with the directions of maximum variance. 
The PCA-transformed embeddings $\mathbf{Z}\in\mathbb{R}^{n\times d}$ of $\mathbf{X}$ are given by the transformation matrix $\mathbf{A}\in\mathbb{R}^{d\times d}$ as
\begin{align}
    \mathbf{Z} = \mathbf{X}\mathbf{A}.\label{eq:Z_XA}
\end{align}
The columns of $\mathbf{Z}$ are called principal components (PC), and the matrix $\mathbf{Z}$ is whitened; i.e., the variance of components in each column is $1$ and the columns are uncorrelated with each other. 
Whitening generally improves the quality of the embeddings~\cite{DBLP:journals/corr/abs-2103-15316,DBLP:journals/ipm/SasakiHSI23}.

\subsection{ICA-transformed embeddings}
ICA transforms the embeddings so that their components are as statistically independent as possible. 
Statistical independence, a stronger property than uncorrelatedness or whitening, ensures that random variables are not only uncorrelated but also independent in a probabilistic sense~\citep{DBLP:books/wi/HyvarinenKO01}.
The ICA-transformed embeddings $\mathbf{S}\in\mathbb{R}^{d\times d}$ of $\mathbf{X}$ are given by the transformation matrix $\mathbf{B}\in\mathbb{R}^{d\times d}$ as
\begin{align}
    \mathbf{S} = \mathbf{X}\mathbf{B}.\label{eq:S_XB}
\end{align}
The columns of $\mathbf{S}$ are called independent components (IC). 

In particular, FastICA~\cite{DBLP:journals/tnn/Hyvarinen99} uses $\mathbf{Z}$ in (\ref{eq:Z_XA}) to compute $\mathbf{S}$ as follows:
\begin{align}
    \mathbf{S} = \mathbf{Z}\mathbf{R}_{\text{ica}},\label{eq:S_ZR}
\end{align}
where $\mathbf{R}_{\text{ica}}\in\mathbb{R}^{d\times d}$ is an orthogonal matrix that maximizes the statistical independence of the columns of $\mathbf{S}$. 
Similar to $\mathbf{Z}$, the matrix $\mathbf{S}$ is also whitened.

\subsection{Normalized ICA-transformed embeddings}\label{sec:hat-s}
The ICA-transformed embedding of a word $w_i$, denoted by $\mathbf{s}_i\in\mathbb{R}^d$, is normalized to $\hat{\mathbf{s}}_i\in\mathbb{R}^d$:
\begin{align}
    \hat{\mathbf{s}}_i:= \mathbf{s}_i/\|\mathbf{s}_i\| = (\hat{s}_i^{(1)},\ldots,\hat{s}_i^{(\ell)},\ldots,\hat{s}_i^{(d)}).
\end{align}
The $\ell$-th component $\hat{s}_i^{(\ell)}$ of the normalized ICA-transformed embedding $\hat{\mathbf{s}}_i$ for the word $w_i$ can be interpreted as the semantic component of $w_i$ along the $\ell$-th axis.
For example, as shown in Fig.~\ref{fig:cosine_ica}, the normalized ICA-transformed embedding of \textit{ultraviolet} has large semantic components of \textit{[chemistry]},  \textit{[biology]}, \textit{[space]}, \textit{[spectrum]}, and \textit{[virology]}.

\section{Decomposition and Interpretation of Cosine Similarity}\label{sec:cosine-similarity}
We define the semantic similarity mentioned in Section~\ref{sec:intro} using the notation introduced in Section~\ref{sec:background} and explain how ICA improves interpretability compared to PCA.

\begin{figure*}[t!]
    \centering
    \includegraphics[width=0.7\textwidth]{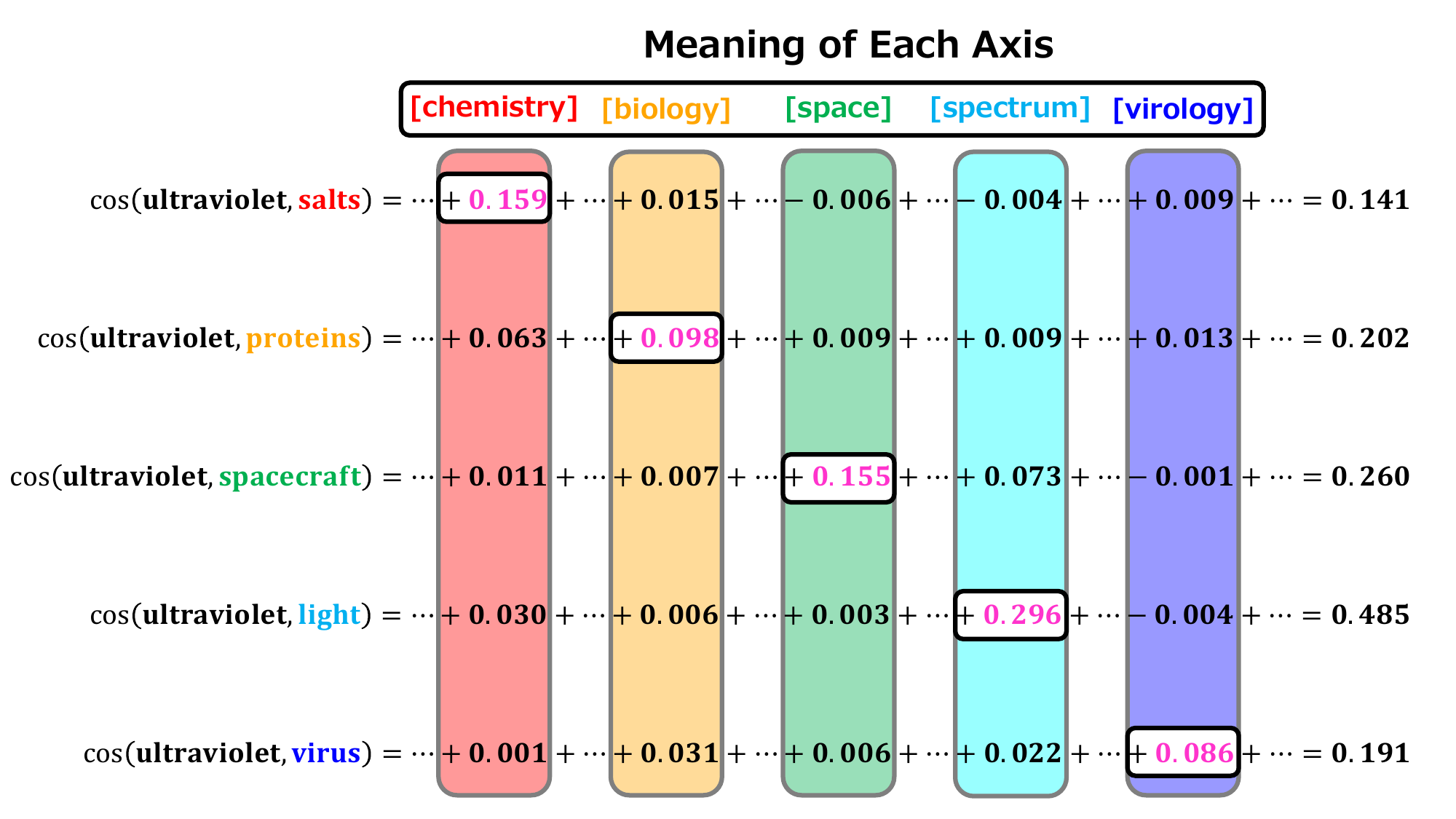}
    \caption{
Cosine similarity interpretation.
For the normalized ICA-transformed GloVe embedding of \textit{ultraviolet}, the meanings of the axes of the top five components are \textit{[chemistry]},  \textit{[biology]}, \textit{[space]}, \textit{[spectrum]}, and \textit{[virology]}.
For the top words on these axes, see Table~\ref{tab:intro-topwords} in Appendix~\ref{app:cosine}.
The cosine similarities are computed between \textit{ultraviolet} and the top word of each axis: \textit{salts}, \textit{proteins}, \textit{spacecraft}, \textit{light}, and \textit{virus}, respectively.
The inner products (i.e., cosine similarities) of their normalized ICA-transformed GloVe embeddings are computed, and the semantic similarities of these five axes are displayed.
For example, the semantic similarity on the \textit{[space]} axis between  \textit{ultraviolet} and \textit{spacecraft} is $0.155$, which is more than half of the cosine similarity of $0.260$.
}
\label{fig:cosine_examples}
\end{figure*}

\subsection{Semantic similarity on axes}\label{sec:semantic-similarity}
Cosine similarity is widely used to measure the similarity between words. 
The cosine similarity between words $w_i$ and $w_j$ can be expressed as the inner product of their normalized ICA-transformed embeddings $\hat{\mathbf{s}}_i$ and $\hat{\mathbf{s}}_j$:
\begin{align}
    \cos(w_i,w_j) = \hat{\mathbf{s}}_i^\top\hat{\mathbf{s}}_j=\sum_{\ell=1}^d \hat{s}_i^{(\ell)}\hat{s}_j^{(\ell)}.\label{eq:cosine}
\end{align}
As seen in Section~\ref{sec:background}, $\hat{s}_i^{(\ell)}$ can be interpreted as the semantic component of a word $w_i$ on the $\ell$-th axis. 
Therefore, the semantic similarity for words $w_i$ and $w_j$ on the $\ell$-th axis, $\text{sem}_\ell(w_i,w_j)$, is defined as:
\begin{align}
    \text{sem}_\ell(w_i,w_j) := \hat{s}_i^{(\ell)}\hat{s}_j^{(\ell)}.\label{eq:sem}
\end{align}
Using the element-wise product (i.e., Hadamard product), $\text{sem}_\ell(w_i, w_j)$, $\ell = 1, \ldots, d$, can also be interpreted as the $\ell$-th component of the vector\footnote{For words $w_i$ and $w_j$, we abbreviate $\hat{\mathbf{s}}_i \odot \hat{\mathbf{s}}_j$ as $w_i \odot w_j$.}
\begin{align*}
\hat{\mathbf{s}}_i \odot \hat{\mathbf{s}}_j = (\hat{s}_i^{(1)}\hat{s}_j^{(1)},\ldots, \hat{s}_i^{(\ell)}\hat{s}_j^{(\ell)}, \dots,\hat{s}_i^{(d)}\hat{s}_j^{(d)}).
\end{align*}
The expression for the cosine similarity in (\ref{eq:cosine}) can be rewritten as in (\ref{eq:cosine-sem}) with the definition (\ref{eq:sem}).
Thus, the cosine similarity can be interpreted as the sum of the semantic similarities over all axes.

Figure~\ref{fig:cosine_examples} shows the cosine similarity computations for \textit{ultraviolet} with \textit{salts}, \textit{proteins}, \textit{spacecraft}, \textit{light}, and \textit{virus}, and semantic similarities on the axes of \textit{[chemistry]}, \textit{[biology]}, \textit{[space]}, \textit{[spectrum]}, and \textit{[virology]}. 
The cosine similarity values, which is the sum of the all semantic similarities, can be interpreted from these semantic similarity values.

\subsection{ICA improves interpretability} \label{sec:cosine-pca-ica}

Note that the cosine similarity between two embeddings is the same for the ICA-transformed embeddings and the PCA-transformed embeddings.
As seen in (\ref{eq:S_ZR}), since $\mathbf{S}$ is $\mathbf{Z}$ multiplied by the orthogonal matrix $\mathbf{R}_\text{ica}$, $\hat{\mathbf{z}}_i=\mathbf{R}_\text{ica}\hat{\mathbf{s}}_i$, where $\hat{\mathbf{z}}_i$ is the normalized PCA-transformed embedding of $w_i$.
Then $\sum_{\ell=1}^d \hat{z}_i^{(\ell)}\hat{z}_j^{(\ell)} = \hat{\mathbf{z}}_i^\top\hat{\mathbf{z}}_j=\hat{\mathbf{s}}_i^\top\hat{\mathbf{s}}_j=\cos(w_i,w_j)$, meaning that $\hat{z}_i^{(\ell)}\hat{z}_j^{(\ell)}$ can also be interpreted as the semantic similarity for the normalized PCA-transformed embeddings.

However, the PCA-based semantic similarity lacks interpretability.
Figure~\ref{fig:cosine} shows bar graphs for the normalized ICA-transformed and PCA-transformed GloVe embeddings of \textit{ultraviolet}, \textit{light}, and their component-wise products.
The sum of the component-wise products is equal to the cosine similarity value ($0.485$) for both transformations.
In the normalized ICA-transformed embeddings, the semantic components of \textit{[spectrum]} are large ($0.535$ for \textit{ultraviolet} and $0.554$ for \textit{light}), and the semantic similarity is $\text{sem}_{\ell_\text{\textit{[spectrum]}}}(\textit{ultraviolet},\textit{light})=0.296$. 
Other semantic similarities are close to zero. 
However, in the normalized PCA-transformed embeddings, although no axis has a component-wise product as large as in the normalized ICA-transformed embeddings, the sum of the component-wise products is still equal to the cosine similarity. 
Therefore, the component-wise products are not close to zero compared to ICA, resulting in a dense vector.

\begin{figure*}[!t]
\centering
\begin{subfigure}{\textwidth}
\centering
    \includegraphics[width=0.95\textwidth]{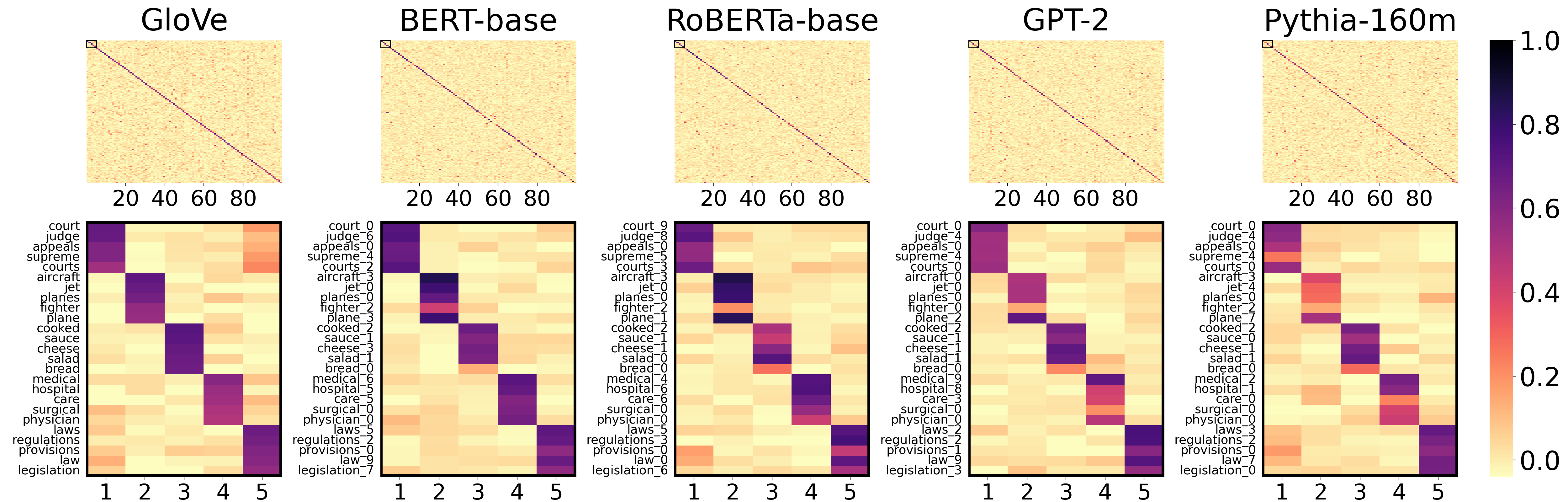}
    \caption{Normalized ICA-transformed embeddings for GloVe and contextualized embedding models.}
    \label{fig:cross-embedding_heatmap_ica}
\end{subfigure}
\par\medskip
\begin{subfigure}{\textwidth}
\centering
    \includegraphics[width=0.95\textwidth]{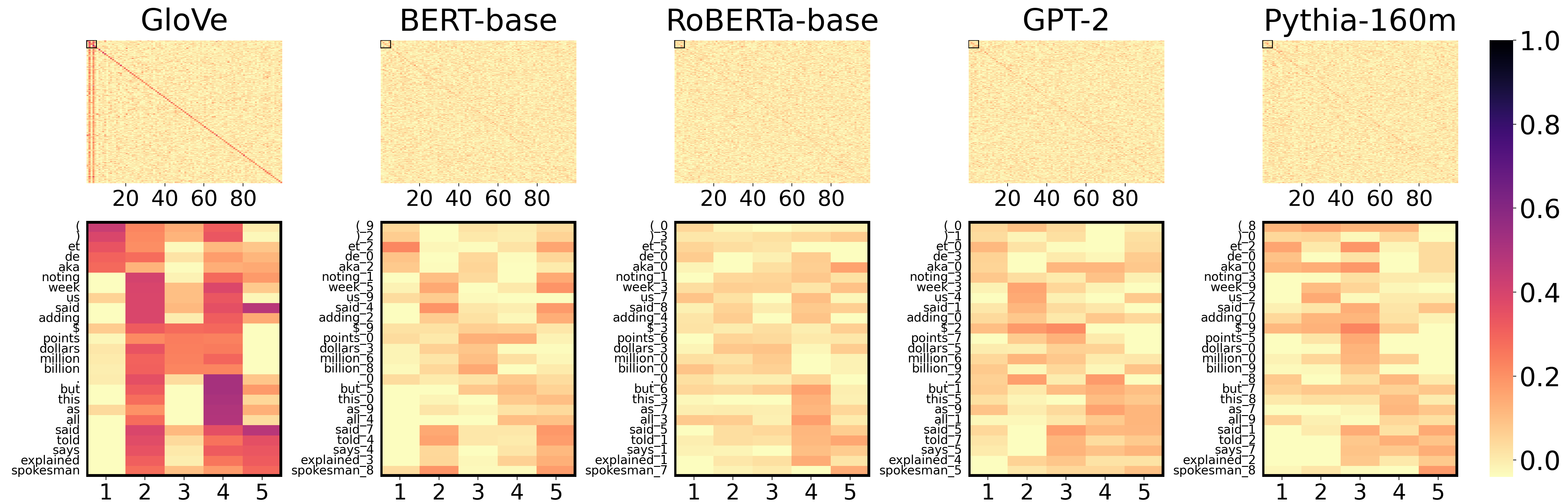}
    \caption{Normalized PCA-transformed embeddings for GloVe and contextualized embedding models.}
    \label{fig:cross-embedding_heatmap_pca}
\end{subfigure}
\caption{
Heatmaps of normalized (a) ICA-transformed and (b) PCA-transformed embeddings for GloVe and four contextualized embedding models, with the axes aligned by the procedure of \citet{DBLP:conf/emnlp/YamagiwaOS23}.
(a) In the top panel, the components of the first 100 axes for 500 embeddings are shown, while the bottom panel magnifies the components of the first 5 axes.
For each GloVe axis, we selected five top words for which corresponding contextualized embeddings exist, and displayed the embeddings along with the contextualized embeddings corresponding to each top word. The embeddings selected for each axis have large component values on that axis, confirming the existence of common semantic axes across the five models.
(b) In contrast to the ICA-transformed embeddings, the PCA-transformed embeddings do not reveal such common semantic axes across the five models.
}
\label{fig:cross-embedding_heatmap}
\end{figure*}

\section{Statistical Analysis of Axis Selection}
\label{sec:statistical-analysis-axis-selection}
For the normalized ICA-transformed embeddings of words $w_i$ and $w_j$, the criteria for selecting axes $\ell$ where the component values $\hat s_i^{(\ell)}$ or the component-wise product values $\hat s_i^{(\ell)} \hat s_j^{(\ell)}$ become large can be determined based on the theory of statistical hypothesis testing. 
Using the probability distribution provided in Appendix~\ref{app:distribution-theory}, the observed component values can be converted into $p$-values, allowing for a probabilistic interpretation. This is illustrated with numerical examples in Fig.~\ref{fig:histgram-ica_maintext} and Table~\ref{tab:table2_5columns}.

\subsection{Component values}
\label{sec:statistical-analysis-component}
As given by equation (\ref{eq:dist-component}) in Appendix~\ref{app:dist-components}, the probability distribution that $\hat s_i^{(\ell)}$ follows is a normal distribution with mean 0 and variance $1/d$,
\[\hat s_i^{(\ell)} \sim \mathcal{N}(0,1/d).\]
Looking at Fig.~\ref{fig:comp_hist_maintext}, we can see that the histogram of the observed $\hat s_i^{(\ell)}$ fits well with the theoretical curve of the normal distribution. The $p$-value is given by the upper tail probability of this probability distribution, and if it is smaller than the pre-determined significance level $\alpha$ (for example, 0.05), that axis is selected. The explicit formula for the $p$-value is expressed using the cumulative distribution function\footnote{\texttt{pnorm} in \texttt{R} or \texttt{norm.cdf} in \texttt{scipy.stats}.}  $\Phi(x)$ of $\mathcal{N}(0,1)$ as $p_1 = \Phi(-\sqrt{d} \hat s_i^{(\ell)})$. 

The $p$-value described above is correct for a single predetermined axis, but in practice, since we are selecting from $d$ axes, it is necessary to account for the multiplicity of tests to avoid false positives. To address this, the Bonferroni correction can be applied by using $p_d = \min\{dp_1, 1\}$. Thus, as a conservative and safe approach, we select axes where $p_d < \alpha$.

\subsection{Product of two component values}
\label{sec:statistical-analysis-product}
As given by equation (\ref{eq:dist-product}) in Appendix~\ref{app:dist-products}, the probability density function of $\hat s_i^{(\ell)} \hat s_j^{(\ell)}$ is
\[
\hat s_i^{(\ell)} \hat s_j^{(\ell)}
 \sim (d/\pi) K_0(d |\hat s_i^{(\ell)} \hat s_j^{(\ell)}|),
\]
where $K_0(\cdot)$ is the modified Bessel function of the second kind of order zero. This probability distribution is particularly intriguing from the perspective of mathematical statistics, and since it is a relatively novel result, the probability distribution has not yet been given a specific name.
From Figs.~\ref{fig:comp_prod_hist_maintext} and \ref{fig:comp_prod_hist_zoom_maintext}, we can see that the histogram of the observed $\hat s_i^{(\ell)} \hat s_j^{(\ell)}$ closely fits the theoretical curve. The upper tail probability of this distribution, denoted as $p_1$, was numerically calculated by integrating the probability density function\footnote{Using Mathematica, we first derived a formula involving the modified Bessel function and the modified Struve function, and then numerically evaluated it for computing $p_1$.}. As a conservative approach, similar to Section~\ref{sec:statistical-analysis-component}, we apply the Bonferroni correction using $p_d = \min\{dp_1, 1\}$.

\begin{figure*}[t!]
    \centering
    \includegraphics[width=\textwidth]{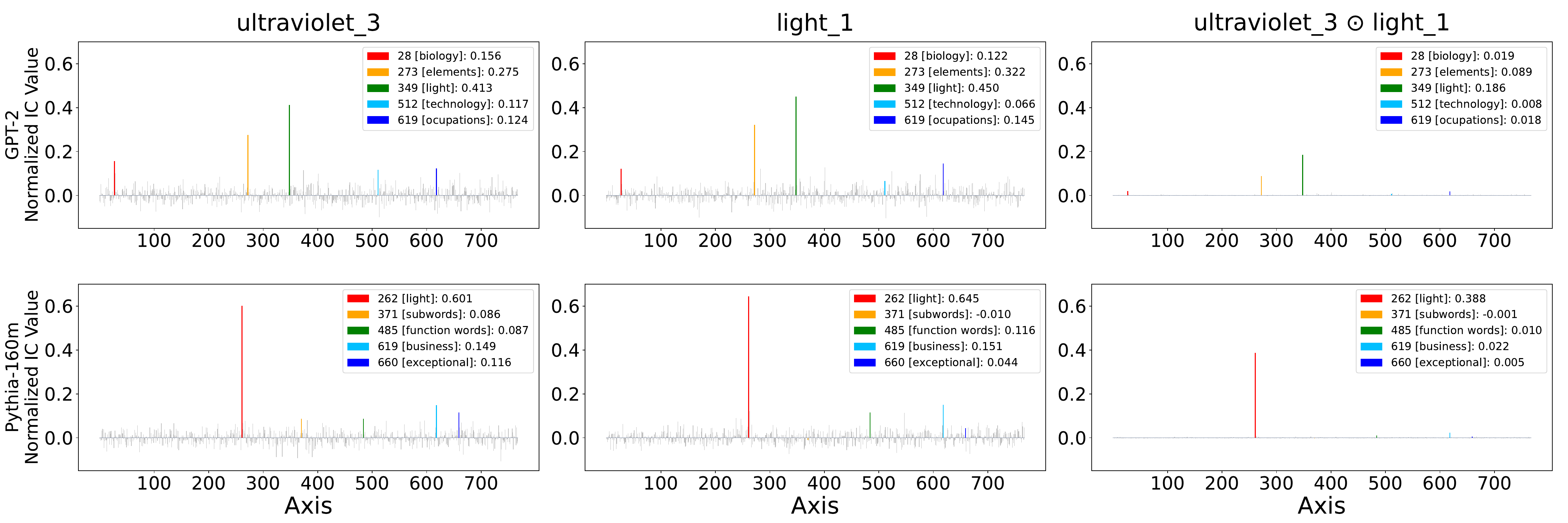}
    \caption{
Similar to Fig.~\ref{fig:cosine_ica}, the component values and their component-wise products for the normalized ICA-transformed contextualized embeddings of \textit{ultraviolet\_3} and \textit{light\_1} are shown in bar graphs. The cosine similarity is 0.576 for GPT-2 and 0.699 for Pythia-160m.
In Appendix~\ref{sec:details_cross_embedding}, Fig.~\ref{fig:cross-embedding_bargraph_ica_bert_roberta} shows results for BERT and RoBERTa, and Fig.~\ref{fig:cross-embedding_bargraph_pca} for normalized PCA-transformed contextualized embeddings.
Table~\ref{tab:4models_topwords} in Appendix~\ref{sec:details_cross_embedding} shows the top words of the axes corresponding to the top 5 components of these normalized embeddings.
}
\label{fig:cross-embedding_bargraph_ica}
\end{figure*}

\begin{table*}[t!]
\centering
\begin{subtable}{.33\textwidth}
  \centering
  \begin{adjustbox}{max width=0.93\linewidth}
 \begin{tabular}{rcrrr}
  \toprule
  Axis & Meaning & Value & $p$-value & Bonferroni \\
  \midrule
{\color{green}\num{349}} & {\color{green}\textit{[light]}} & \num{0.413} & \num{1.28e-30} & \num{9.86e-28} \\
{\color{orange}\num{273}} & {\color{orange}\textit{[elements]}} & \num{0.275} & \num{1.26e-14} & \num{9.70e-12} \\
{\color{red}\num{28}} & {\color{red}\textit{[biology]}} & \num{0.156} & \num{7.62e-06} & \num{5.85e-03} \\
{\color{blue}\num{619}} & {\color{blue}\textit{[occupations]}} & \num{0.124} & \num{3.02e-04} & \num{2.32e-01} \\
{\color{cyan}\num{512}} & {\color{cyan}\textit{[technology]}} & \num{0.117} & \num{5.85e-04} & \num{4.49e-01} \\
\bottomrule
  \end{tabular}
  \end{adjustbox}
\caption{\textit{ultraviolet\_3}}
\end{subtable}\begin{subtable}{.33\textwidth}
  \centering
  \begin{adjustbox}{max width=0.93\linewidth}
 \begin{tabular}{rcrrr}
  \toprule
  Axis & Meaning & Value & $p$-value & Bonferroni \\
  \midrule
{\color{green}\num{349}} & {\color{green}\textit{[light]}} & \num{0.450} & \num{5.88e-36} & \num{4.51e-33} \\
{\color{orange}\num{273}} & {\color{orange}\textit{[elements]}} & \num{0.322} & \num{2.31e-19} & \num{1.78e-16} \\
{\color{blue}\num{619}} & {\color{blue}\textit{[occupations]}} & \num{0.145} & \num{2.97e-05} & \num{2.28e-02} \\
\num{402} & \textit{[rainstorm]} & \num{0.128} & \num{1.89e-04} & \num{1.45e-01} \\
\num{284} & \textit{[challenges]} & \num{0.122} & \num{3.58e-04} & \num{2.75e-01} \\
\bottomrule
  \end{tabular}
  \end{adjustbox}
\caption{\textit{light\_1}}
\end{subtable}
\begin{subtable}{.33\textwidth}
  \centering
  \begin{adjustbox}{max width=0.93\linewidth}
 \begin{tabular}{rcrrr}
  \toprule
  Axis & Meaning & Value & $p$-value & Bonferroni \\
  \midrule
{\color{green}\num{349}} & {\color{green}\textit{[light]}} & \num{0.186} & \num{3.84e-64} & \num{2.95e-61} \\
{\color{orange}\num{273}} & {\color{orange}\textit{[elements]}} & \num{0.089} & \num{1.43e-31} & \num{1.10e-28} \\
{\color{red}\num{28}} & {\color{red}\textit{[biology]}} & \num{0.019} & \num{4.60e-08} & \num{3.53e-05} \\
{\color{blue}\num{619}} & {\color{blue}\textit{[occupations]}} & \num{0.018} & \num{1.08e-07} & \num{8.27e-05} \\
\num{402} & \textit{[rainstorm]} & \num{0.013} & \num{6.89e-06} & \num{5.29e-03} \\
\bottomrule
  \end{tabular}
  \end{adjustbox}
  \caption{\textit{ultraviolet\_3} $\odot$ \textit{light\_1}}
\end{subtable}
  \caption{
(a, b) show the observed component values of normalized ICA-transformed GPT-2 embeddings in Fig.~\ref{fig:cross-embedding_bargraph_ica} and (c) shows their component-wise products with the $p$-values and Bonferroni-corrected values in each table.
}
\label{tab:table2_5columns_gpt2_ica}
\end{table*}

\begin{table*}[t!]
\centering
\begin{subtable}{.33\textwidth}
  \centering
  \begin{adjustbox}{max width=0.93\linewidth}
 \begin{tabular}{rcrrr}
  \toprule
  Axis & Meaning & Value & $p$-value & Bonferroni \\
  \midrule
{\color{red}\num{262}} & {\color{red}\textit{[light]}} & \num{0.601} & \num{1.32e-62} & \num{1.02e-59} \\
{\color{cyan}\num{619}} & {\color{cyan}\textit{[business]}} & \num{0.149} & \num{1.78e-05} & \num{1.36e-02} \\
{\color{blue}\num{660}} & {\color{blue}\textit{[exceptional]}} & \num{0.116} & \num{6.80e-04} & \num{5.22e-01} \\
{\color{green}\num{485}} & {\color{green}\textit{[function words]}} & \num{0.087} & \num{7.70e-03} & \num{1.00e+00} \\
{\color{orange}\num{371}} & {\color{orange}\textit{[subwords]}} & \num{0.086} & \num{8.35e-03} & \num{1.00e+00} \\
\bottomrule
  \end{tabular}
  \end{adjustbox}
\caption{\textit{ultraviolet\_3}}
\end{subtable}\begin{subtable}{.33\textwidth}
  \centering
  \begin{adjustbox}{max width=0.93\linewidth}
 \begin{tabular}{rcrrr}
  \toprule
  Axis & Meaning & Value & $p$-value & Bonferroni \\
  \midrule
{\color{red}\num{262}} & {\color{red}\textit{[light]}} & \num{0.645} & \num{7.92e-72} & \num{6.09e-69} \\
{\color{cyan}\num{619}} & {\color{cyan}\textit{[business]}} & \num{0.151} & \num{1.51e-05} & \num{1.16e-02} \\
\num{264} & \textit{[elements]} & \num{0.122} & \num{3.54e-04} & \num{2.72e-01} \\
{\color{green}\num{485}} & {\color{green}\textit{[function words]}} & \num{0.116} & \num{6.40e-04} & \num{4.92e-01} \\
\num{548} & \textit{[subwords]} & \num{0.114} & \num{7.82e-04} & \num{6.00e-01} \\
\bottomrule
  \end{tabular}
  \end{adjustbox}
\caption{\textit{light\_1}}
\end{subtable}
\begin{subtable}{.33\textwidth}
  \centering
  \begin{adjustbox}{max width=0.93\linewidth}
 \begin{tabular}{rcrrr}
  \toprule
  Axis & Meaning & Value & $p$-value & Bonferroni \\
  \midrule
{\color{red}\num{262}} & {\color{red}\textit{[light]}} & \num{0.388} & \num{9.66e-132} & \num{7.42e-129} \\
{\color{cyan}\num{619}} & {\color{cyan}\textit{[business]}} & \num{0.022} & \num{2.99e-09} & \num{2.29e-06} \\
{\color{green}\num{485}} & {\color{green}\textit{[function words]}} & \num{0.010} & \num{5.44e-05} & \num{4.18e-02} \\
\num{264} & \textit{[elements]} & \num{0.008} & \num{2.68e-04} & \num{2.06e-01} \\
\num{336} & \textit{[creation]} & \num{0.006} & \num{1.23e-03} & \num{9.42e-01} \\
\bottomrule
  \end{tabular}
  \end{adjustbox}
  \caption{\textit{ultraviolet\_3} $\odot$ \textit{light\_1}}
\end{subtable}
\caption{
(a, b) show the observed component values of normalized ICA-transformed Pythia-160m embeddings in Fig.~\ref{fig:cross-embedding_bargraph_ica} and (c) shows their component-wise products with the $p$-values and Bonferroni-corrected values in each table.
}
\label{tab:table2_5columns_pythia}
\end{table*}

\section{Consistency of ICA Transformation for Contextualized Embeddings} \label{sec:various_embeddings}
In this section, we show the consistency of the interpretability of ICA components and their products across different contextualized embeddings.
See Appendix~\ref{sec:details_cross_embedding} for the details of the experiments.

\paragraph{Consistency of component values.} 
For contextualized embeddings, we used BERT-base~\cite{DBLP:conf/naacl/DevlinCLT19}, RoBERTa-base~\cite{radford2019language}, GPT-2~\cite{radford2019language}, and Pythia-160m~\cite{DBLP:conf/icml/BidermanSABOHKP23}, and computed 50,000 embeddings for each model.

Figure~\ref{fig:cross-embedding_heatmap_ica} shows the result of axis matching using correlation coefficients for these ICA-transformed embeddings.
Despite applying ICA transformations to the embeddings of different models individually, we observed over 100 axes representing common semantic content across the five models.
In contrast, Fig.~\ref{fig:cross-embedding_heatmap_pca} shows that PCA-transformed embeddings do not exhibit such common semantic axes among the models.
To improve clarity, prefixes such as \#\# and Ġ were removed in Fig.~\ref{fig:cross-embedding_heatmap}.

\paragraph{Consistency of the product of two components.} 
Figure~\ref{fig:cross-embedding_bargraph_ica} shows the component values of the normalized ICA-transformed embeddings for GPT-2 and Pythia-160m, as well as the product of those components, for \textit{ultraviolet\_3} and \textit{light\_1} in the same sentence. Similar to Fig.~\ref{fig:cosine_ica}, the component values of the embeddings are sparse overall, but when looking at the component-wise products, the common semantic components between the two embeddings (e.g., \textit{[light]}) stand out more clearly.
Additionally, Tables~\ref{tab:table2_5columns_gpt2_ica} and~\ref{tab:table2_5columns_pythia} present the top 5 values for bar graphs in Fig.~\ref{fig:cross-embedding_bargraph_ica}, along with the $p$-values and Bonferroni-corrected values. As in Table~\ref{tab:table2_5columns}, our method identifies statistically significant axes.

\section{Quantitative Experiments}\label{sec:experiments}
We quantitatively evaluate that ICA provides better interpretability than PCA, and that normalization further enhances interpretability (Sec.~\ref{sec:intruder}).
We confirm that ICA tends to yield larger component values compared to PCA, resulting in more dimensions that can be interpreted with specific meanings (Sec.~\ref{sec:ica-vs-pca}). Then, we quantitatively verify the sparsity of the semantic similarities on axes in the normalized ICA-transformed embeddings (Sec.~\ref{sec:downstream}).

\subsection{General settings} \label{sec:settings}
\paragraph{Word embedding.}
We use GloVe~\cite{DBLP:conf/emnlp/PenningtonSM14} embeddings with $d=300$ dimensions and vocabulary size $n=400{,}000$.
\paragraph{ICA-transformation.}
To compute the ICA transformation, we utilized the FastICA implementation available in \texttt{scikit-learn}~\cite{DBLP:journals/jmlr/PedregosaVGMTGBPWDVPCBPD11}.
The hyperparameters for FastICA were configured with a maximum number of iterations set to $10{,}000$ and a convergence tolerance of \num{1e-10}.
As post-processing, we flip the signs of the axes in $\mathbf{S}$ if necessary to ensure positive skewness, and sort the axes in descending order of their skewness.

\subsection{Normalization enhances interpretability} \label{sec:intruder}
We conducted a word intrusion task~\cite{chang-2009-reading, DBLP:conf/ijcai/SunGLXC16} to quantitatively evaluate that ICA provides better interpretability than PCA and that normalization further enhances this interpretability.
The word intrusion task assesses the semantic coherence of a set of the top $k$ words by measuring the ability to identify an intruder.

\paragraph{Settings.} 
For the four embedding types, we compute a semantic coherence score for each dimension $\ell = 1, \ldots, d$.
The interpretability score for each embedding type is the median of these $d$ scores.
Details of the word intrusion task and scoring method are provided in Appendix~\ref{app:intruder}.

\begin{figure}[t!]
    \centering
    \includegraphics[width=0.72\columnwidth]{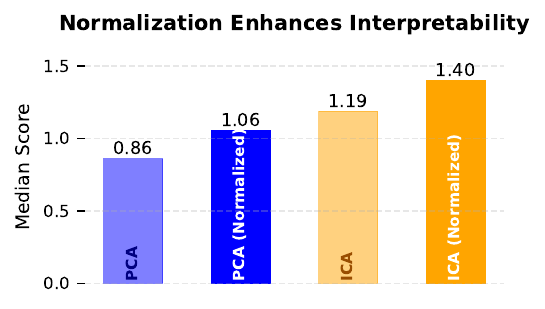}
    \caption{
    Result of word intrusion task. A large value indicates high consistency in the semantic components represented by each axis of the embeddings. We used the GloVe embeddings transformed by ICA and PCA, and their respective normalized versions. For each dimension, we select the top $k=5$ words with the largest component values and compute the average score of $10$ different intruder choices. The final result shown is the median of these scores across $d=300$ dimensions.
    }
\label{fig:intruder}
\end{figure}

\paragraph{Results and discussion.} 
Figure~\ref{fig:intruder} shows that ICA consistently outperforms PCA in interpretability scores, with normalized embeddings achieving higher scores for both methods.
This indicates that ICA components offer better interpretability than PCA, and that normalization further enhances the interpretability.
As an intuitive explanation, normalizing embeddings emphasizes each word's association with specific ICA-defined semantic axes. 
This allows words with focused meanings to stand out along relevant dimensions, while words with mixed semantics exhibit less pronounced values across axes.
This improvement in interpretability may be related to the experimental results showing that the fidelity of rankings along axes and embeddings increases with normalization (Appendix~\ref{app:normalization}).

\subsection{ICA: Larger component values}\label{sec:ica-vs-pca}
By sorting the components, we investigate whether normalized ICA-transformed or PCA-transformed embeddings give greater emphasis on larger component values to improve interpretability.

\paragraph{Settings.}
To compare the ICA and PCA transformations, we employed two methods: 
(a) We sorted the $n$ component values {\bf along embeddings} for each axis in descending order and averaged them over the $d$ axes.
(b) We sorted the $d$ component values {\bf along axes} for each embedding in descending order and averaged them over the $n$ embeddings.

\begin{figure}[t!]
    \centering
    \begin{minipage}{0.48\linewidth}
        \centering
        \includegraphics[width=\linewidth]{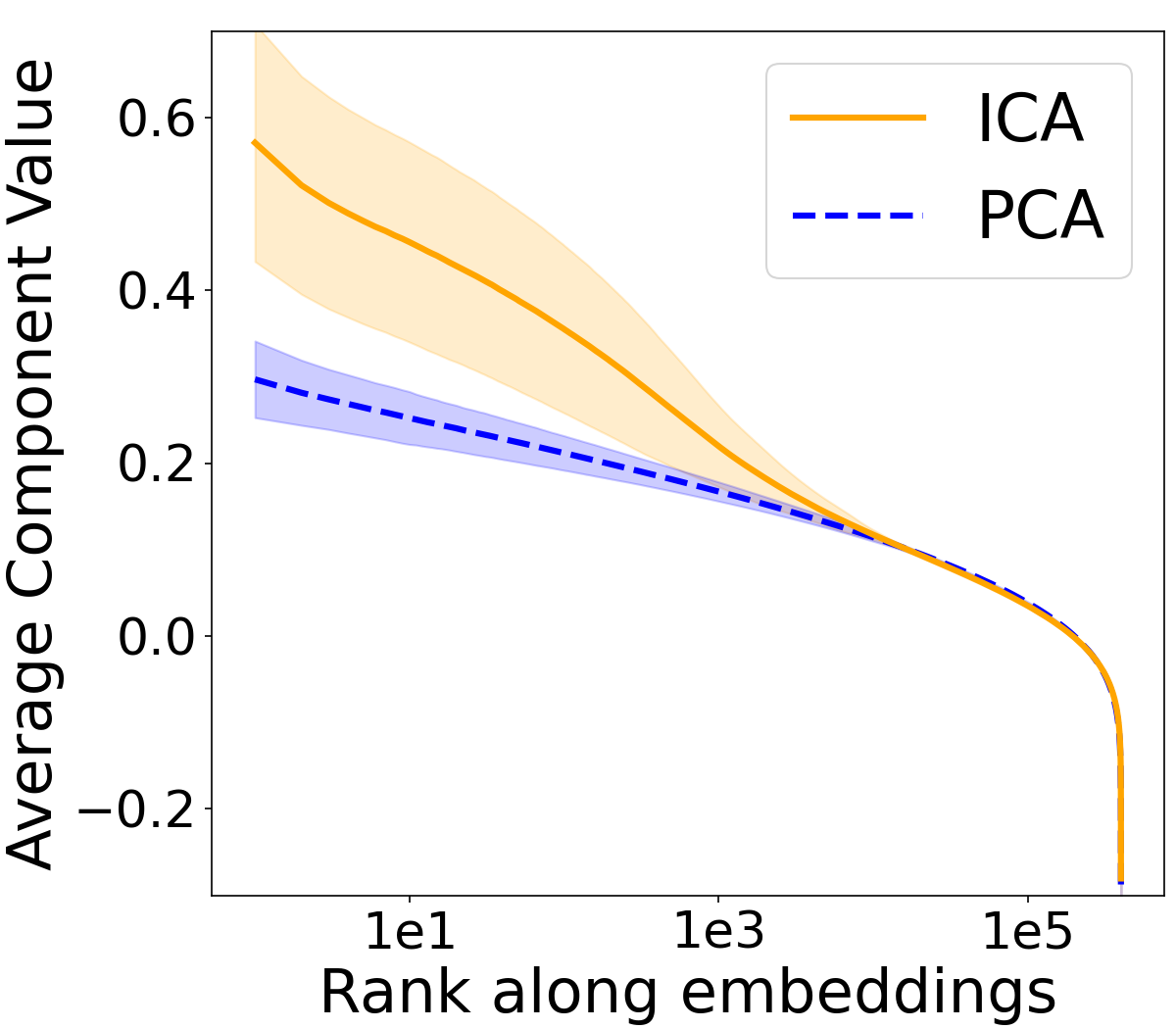}
        \subcaption{Sorted along embeddings}
        \label{fig:sort_emb_component}
    \end{minipage}\hfill
    \begin{minipage}{0.48\linewidth}
        \centering
        \includegraphics[width=\linewidth]{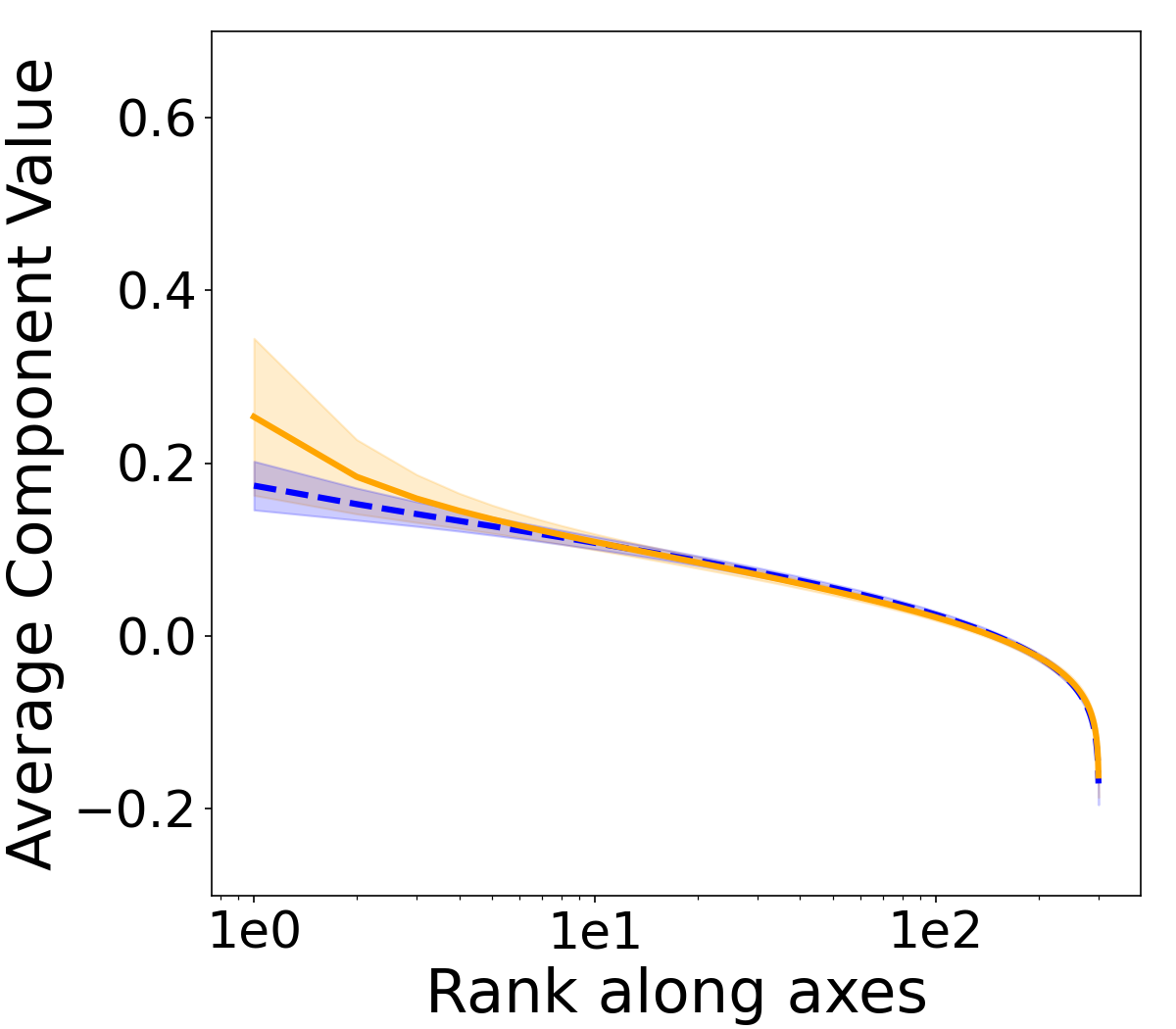}
        \subcaption{Sorted along axes}
        \label{fig:sort_axis_component}
    \end{minipage}
    \caption{
Comparison of component values of the normalized GloVe embeddings transformed by ICA and PCA. The component values are averaged after being sorted in descending order (a) along embeddings for each axis, and  (b) along axes for each embedding. The range of $\pm1\sigma$ is shown, where $\sigma$ is the standard deviation of the component values.
See Fig.~\ref{fig:sorted_along_4models} in Appendix~\ref{app:ica-vs-pca} for the results of contextualized embeddings.
}
    \label{fig:sort_component}
\end{figure}

\paragraph{Results and discussion.}
The sorted component values are shown in Fig.~\ref{fig:sort_component}.
In Fig.~\ref{fig:sort_emb_component}, ICA component values are larger than those of PCA up to approximately the {10,000}th embedding, and in Fig.~\ref{fig:sort_axis_component}, ICA component values are larger for the first few axes.
These results demonstrate that ICA tends to emphasize larger component values compared to PCA.

\subsection{ICA: Sparsity of semantic similarities}\label{sec:downstream}

\begin{figure}[t!]
    \centering
    \includegraphics[width=0.92\columnwidth]{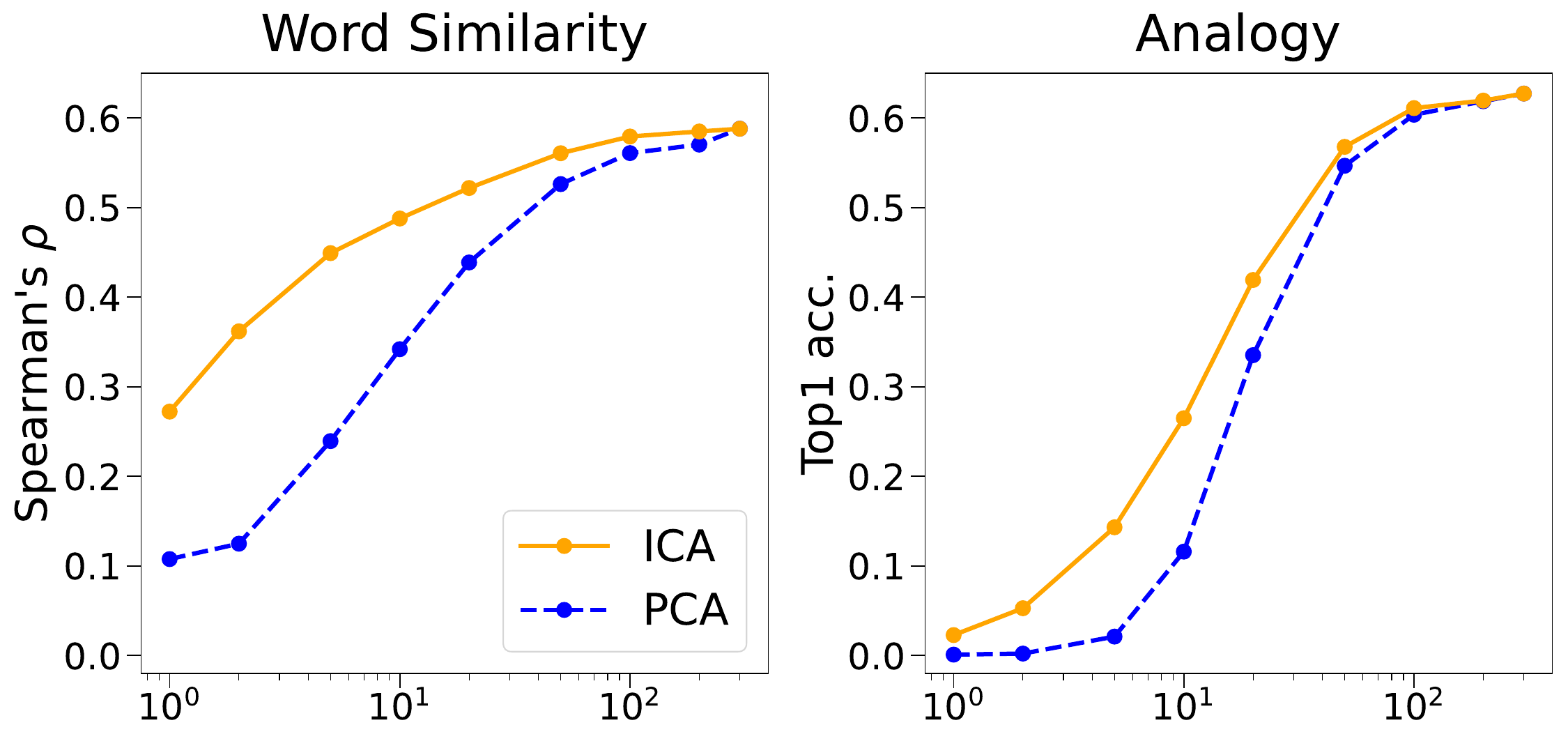}
    \caption{
Performance comparison between ICA and PCA for the GloVe embeddings when reducing non-zero normalized component-wise products and computing cosine similarity. Each value represents the average of 8 word similarity tasks or 30 analogy tasks.
}
\label{fig:downstream}
\end{figure}

We investigate whether the sparsity of component-wise products observed in Fig.~\ref{fig:cosine} generalizes to other word pairs.
To this end, we perform word similarity and analogy tasks, comparing ICA-transformed and PCA-transformed embeddings.

\paragraph{Settings.}
We evaluate the performance degradation as component-wise products are replaced by zero in ascending order until only $p$ non-zero products remain\footnote{In Section 6.2 of~\citet{DBLP:conf/emnlp/YamagiwaOS23}, small component values are zeroed out before computing cosine similarity. In contrast, we first compute element-wise products between normalized embeddings and then zero out small components.}.
For this analysis, we use the Word Embedding Benchmark~\cite{DBLP:journals/corr/JastrzebskiLC17}.
See Appendix~\ref{app:downstream} for more details.

\paragraph{Results and discussion.}
In Fig.~\ref{fig:downstream},
for both word similarity and analogy tasks, the ICA-transformed embeddings consistently outperform the PCA-transformed embeddings, even when the number of non-zero products is small.
This indicates that ICA can represent cosine similarity with fewer dimensions than PCA.

\paragraph{Applications.}
Based on these results, we present potential applications of ICA-transformed embeddings. 
In Appendix~\ref{app:ideal-embeddings}, we consider sparse embeddings that contain only the semantic components of \textit{[food]}, \textit{[animals]}, or \textit{[plants]} and show that word retrieval works effectively. 
Additionally, in Appendix~\ref{app:woman-girl-man}, we show that ablating the \textit{[female]} semantic component from the ICA-transformed embedding of \textit{woman} results in greater similarity to embeddings such as \textit{man} and \textit{boy}.
Furthermore, Appendix~\ref{app:case_study} shows case studies on the noun and verb senses of \textit{shore}, as well as multilingual analyses.

\section{Related Work}\label{sec:related-work}
\paragraph{Cosine similarity.}
Cosine similarity is widely used to measure the similarity between two embeddings, such as word~\cite{DBLP:journals/corr/MikolovLS13,DBLP:conf/nips/MikolovSCCD13}, token~\cite{DBLP:conf/iclr/ZhangKWWA20,DBLP:conf/acl/BommasaniDC20}, and sentence embeddings~\cite{DBLP:conf/emnlp/ReimersG19,DBLP:conf/emnlp/GaoYC21}. 
Cross-lingual alignment methods based on cosine similarity have also been proposed~\cite{DBLP:conf/naacl/XingWLL15,DBLP:conf/emnlp/Alvarez-MelisJ18,DBLP:conf/iclr/LampleCRDJ18}. 

There are studies that question the effectiveness of cosine similarity. 
For example, \citet{DBLP:conf/emnlp/SchnabelLMJ15} showed that word frequency can affect cosine similarity. 
\citet{DBLP:journals/corr/abs-2403-05440} used linear models to show cases where cosine similarity fails.

\paragraph{Interpretability of embeddings.}
Research on the interpretability of embeddings has used various methods, including non-negative matrix factorization~\cite{DBLP:conf/coling/MurphyTM12}, sparse coding~\cite{DBLP:conf/acl/FaruquiTYDS15}, methods for learning interpretable embeddings~\cite{DBLP:conf/emnlp/LuoLLS15,DBLP:conf/ijcai/SunGLXC16}, rotation of embeddings~\cite{DBLP:conf/emnlp/ParkBO17}, Singular Value Decomposition (SVD) ~\cite{shin2018interpreting}, autoencoders~\cite{DBLP:conf/aaai/SubramanianPJBH18,huben2024sparse}, and PCA~\cite{DBLP:conf/tsd/Musil19}. 

Some research has applied ICA to embeddings.
For example, \citet{chagnaa2007feature} showed that the features of similar verbs have a large same independent component through bar graphs. 
The axes of the ICA-transformed embeddings are known to be interpretable~\cite{marecek-etal-2020,DBLP:conf/coling/MusilM24}, and such axes are also observed in embeddings for other languages, dynamic models, and images~\cite{DBLP:conf/emnlp/YamagiwaOS23}.
\citet{DBLP:conf/emnlp/LiMY24} statistically validated their existence.
The dependencies between axes are explored via semantic continuity optimization~\citep{DBLP:journals/corr/abs-2401-06112} and higher-order correlations~\citep{DBLP:conf/emnlp/OyamaYS24}.

\section{Conclusion}
Inspired by previous studies showing that ICA transformation can produce sparse and interpretable embeddings from dense distributed representations, we made the following contributions:

(1)~We experimentally demonstrated that normalized ICA-transformed embeddings exhibit sparsity, enhancing the interpretability of each axis. In particular, ICA provides better interpretability than PCA, and normalization further improves this interpretability. These findings were tested using static embeddings (GloVe) and four contextualized embeddings (BERT, RoBERTa, GPT-2, and Pythia).
    
(2)~We proposed interpreting cosine similarity as the sum of semantic similarities across axes, where the component-wise product of two embeddings represents the semantic similarity on each axis. The experiments showed that ICA results in greater sparsity in the component-wise products compared to PCA, allowing cosine similarity to be represented with fewer dimensions.
    
(3)~By deriving the probability distribution that governs each component in the normalized ICA-transformed embeddings, we proposed a method to statistically select significant axes from an embedding. Similarly, by newly deriving the probability distribution that governs the component-wise products, we proposed a method to select statistically significant axes shared between two embeddings.

\clearpage
\section*{Limitations}
\begin{itemize}
\item This study explains the interpretation of cosine similarity using centered and whitened embeddings. These embeddings are different from the original embeddings.
\item The meanings of the axes of the ICA-transformed embeddings are manually interpreted based on their top words after normalization.
In addition, it is not always possible to interpret the meaning of an axis from its top words. Note that the top words used to interpret the meanings of the axes in this study can all be found in the Appendix sections.
\item We need to pay attention to the signs of the axes of the ICA-transformed embeddings. In this study, following~\citet{DBLP:conf/emnlp/YamagiwaOS23}, we ensure that all axes have positive skewness by flipping their signs when necessary. We then assume that larger component values are more representative of the meanings of the axes. 
\item To explain our interpretation of cosine similarity, we mainly use the GloVe\footnote{\url{https:// nlp.stanford.edu/data/glove.6B.zip.}} embeddings for which cosine similarity works well. Although cosine similarity may not be effective for all embeddings~\cite{DBLP:journals/corr/abs-2403-05440}, this study does not cover which specific types of embeddings are suitable for cosine similarity. 
\item The Bonferroni correction, used to adjust for multiple testing when selecting statistically significant axes from the $d$ axes, is safe but conservative, and tends to result in fewer axes being selected. Other approaches, such as the False Discovery Rate (FDR), should also be considered.
\item The axes selected from the component-wise product of the two embeddings do not necessarily have logical consistency with the axes selected from the components of each embedding. For example, in the numerical example in Table~\ref{tab:table2_5columns}, where axes are selected with $\alpha=0.05$, \textit{[spectrum]}, \textit{[function words]}, and \textit{[boxing]} are selected for \textit{light}, and \textit{[chemistry]} is not selected.
However, as shown in Section~\ref{sec:intro}, \textit{[chemistry]} is selected for the pair \textit{ultraviolet} and \textit{light}.
\item ICA is a more complex method than PCA; the computation of the orthogonal matrix $\mathbf{R}_{\text{ica}}$ in (\ref{eq:S_ZR}) for ICA takes considerably more time than that of the matrix $\mathbf{A}$ in (\ref{eq:Z_XA}) for PCA.
The computation time for FastICA using \texttt{scikit-learn} depends mainly on the embedding dimension $d$, vocabulary size $n$, the maximum number of iterations, and the convergence tolerance. For example, it takes several hours for GloVe embeddings with the settings in Section~\ref{sec:settings}.
\end{itemize}

\section*{Ethics Statement}
This study complies with the \href{https://www.aclweb.org/portal/content/acl-code-ethics}{ACL Ethics Policy}.

\section*{Acknowledgments}
We would like to thank the anonymous reviewers for their helpful comments and suggestions.
This study was partially supported by JSPS KAKENHI 22H05106, 23H03355, JST CREST JPMJCR21N3, JST SPRING JPMJSP2110, JST BOOST JPMJBS2407.

\section*{Code availability}
Code is available at \url{https://github.com/ymgw55/Cosine-Similarity-via-ICA}.

\bibliography{custom}

\appendix

\renewcommand{\arraystretch}{1.15} 
\begin{table*}[t!]
\tiny
\centering
\begin{tabular}{@{\hspace{0.5em}}c@{\hspace{0.5em}}|@{\hspace{0.5em}}r@{\hspace{0.5em}}|@{\hspace{0.5em}}c@{\hspace{0.15em}}c@{\hspace{0.15em}}c@{\hspace{0.15em}}c@{\hspace{0.15em}}c@{\hspace{0.5em}}c@{\hspace{0.15em}}c@{\hspace{0.15em}}c@{\hspace{0.15em}}c@{\hspace{0.15em}}c@{\hspace{0.5em}}|@{\hspace{0.5em}}c@{\hspace{0.5em}}}
\toprule
 & Axis & Top1 & Top2 & Top3 & Top4 & Top5 & Top6 & Top7 & Top8 & Top9 & Top10 & Meaning\\
\midrule
\multirow{5}{*}{\rotatebox{90}{Normalized ICA}}  & {\color{red}53} & \textit{salts} & \textit{solvents} & \textit{chlorine} & \textit{hydrogen} & \textit{inorganic} & \textit{ammonia} & \textit{chloride} & \textit{flammable} & \textit{sulfide} & \textit{sulfur} & {\color{red}\textit{[chemistry]}} \\
 & {\color{orange}68} & \textit{proteins} & \textit{protein} & \textit{genes} & \textit{gene} & \textit{mrna} & \textit{receptor} & \textit{transcription} & \textit{activation} & \textit{p53} & \textit{rna} & {\color{orange}\textit{[biology]}} \\
 & {\color{green}141} & \textit{spacecraft} & \textit{astronauts} & \textit{orbit} & \textit{nasa} & \textit{astronaut} & \textit{orbiter} & \textit{space} & \textit{orbiting} & \textit{mars} & \textit{atlantis} & {\color{green}\textit{[space]}} \\
 & {\color{cyan}194} & \textit{light} & \textit{ultraviolet} & \textit{infrared} & \textit{sunlight} & \textit{uv} & \textit{shadows} & \textit{bright} & \textit{illumination} & \textit{glow} & \textit{illuminate} & {\color{cyan}\textit{[spectrum]}} \\
 & {\color{blue}197} & \textit{virus} & \textit{h5n1} & \textit{influenza} & \textit{flu} & \textit{contagious} & \textit{outbreak} & \textit{swine} & \textit{avian} & \textit{viruses} & \textit{pandemic} & {\color{blue}\textit{[virology]}} \\
\midrule
\multirow{5}{*}{\rotatebox{90}{Normalized PCA}} 
 & {\color{red}80} & \textit{zhongshan} & \textit{solar} & \textit{optical} & \textit{shyh} & \textit{electrics} & \textit{wafers} & \textit{nerpa} & \textit{selden} & \textit{wbut} & \textit{reutemann} & {\color{red}\textit{[PC80]}} \\
 & {\color{orange}92} & \textit{woodcuts} & \textit{natrun} & \textit{eriboll} & \textit{linocuts} & \textit{shamva} & \textit{cellblock} & \textit{hafslund} & \textit{g2} & \textit{heidelberg} & \textit{venter} & {\color{orange}\textit{[PC92]}} \\
 & {\color{green}152} & \textit{sidebar} & \textit{maternal} & \textit{smoker} & \textit{customizer} & \textit{non-qualified} & \textit{sufia} & \textit{foundresses} & \textit{frosting} & \textit{traudl} & \textit{romm} & {\color{green}\textit{[PC152]}} \\
 & {\color{cyan}153} & \textit{monogamous} & \textit{lifespan} & \textit{necessitate} & \textit{loyals} & \textit{supplementation} & \textit{skrall} & \textit{zng} & \textit{gietzen} & \textit{remnant} & \textit{well-meaning} & {\color{cyan}\textit{[PC153]}} \\
 & {\color{blue}222} & \textit{replication} & \textit{isley} & \textit{guaporé} & \textit{bobos} & \textit{pawhuska} & \textit{foss} & \textit{rigoberto} & \textit{angara} & \textit{laporta} & \textit{200-250} & {\color{blue}\textit{[PC222]}} \\
\bottomrule
\end{tabular}
\caption{
For the normalized ICA-transformed and PCA-transformed GloVe embeddings of \textit{ultraviolet} in Fig.~\ref{fig:cosine}, the axes of the top 5 component values are focused on, and their top 10 words are shown.
For ICA, the meanings of the axes are interpreted from these listed words and labeled such as \textit{[chemistry]}.
For PCA, however, since it is difficult to interpret the meanings of the axes, they are simply labeled such as \textit{[PC80]}.
}
\label{tab:intro-topwords}
\end{table*}
\renewcommand{\arraystretch}{1.0}

\renewcommand{\arraystretch}{1.25} 
\begin{table*}[t!]
\tiny
\centering
\begin{tabular}{@{\hspace{0.5em}}c@{\hspace{0.5em}}|@{\hspace{0.5em}}r@{\hspace{0.5em}}|@{\hspace{0.5em}}c@{\hspace{0.5em}}c@{\hspace{0.5em}}c@{\hspace{0.5em}}c@{\hspace{0.5em}}c@{\hspace{0.5em}}c@{\hspace{0.5em}}c@{\hspace{0.5em}}c@{\hspace{0.5em}}c@{\hspace{0.5em}}c@{\hspace{0.5em}}|@{\hspace{0.5em}}c@{\hspace{0.5em}}}
\toprule
 & Axis & Top1 & Top2 & Top3 & Top4 & Top5 & Top6 & Top7 & Top8 & Top9 & Top10 & Meaning\\
\midrule
\multirow{5}{*}{\rotatebox{90}{\textit{light}}}  
 & {\color{cyan}194} & \textit{light} & \textit{ultraviolet} & \textit{infrared} & \textit{sunlight} & \textit{uv} & \textit{shadows} & \textit{bright} & \textit{illumination} & \textit{glow} & \textit{illuminate} & {\color{cyan}\textit{[spectrum]}} \\
 & 1 & \textit{.} & \textit{but} & \textit{though} & \textit{although} & \textit{,} & \textit{even} & \textit{that} & \textit{instance} & \textit{both} & \textit{however} & \textit{[function words]} \\
 & 153 & \textit{welterweight} & \textit{heavyweight} & \textit{middleweight} & \textit{featherweight} & \textit{ibf} & \textit{wbc} & \textit{holyfield} & \textit{wba} & \textit{cruiserweight} & \textit{bout} & \textit{[boxing]} \\
 & {\color{red}53} & \textit{salts} & \textit{solvents} & \textit{chlorine} & \textit{hydrogen} & \textit{inorganic} & \textit{ammonia} & \textit{chloride} & \textit{flammable} & \textit{sulfide} & \textit{sulfur} & {\color{red}\textit{[chemistry]}} \\
 & 58 & \textit{23rd} & \textit{35th} & \textit{27th} & \textit{22nd} & \textit{26th} & \textit{39th} & \textit{24th} & \textit{36th} & \textit{31st} & \textit{37th} & \textit{[ordinal]} \\
\midrule
\multirow{5}{*}{\rotatebox{90}{\textit{ultraviolet} $\odot$ \textit{light}}}
 & {\color{cyan}194} & \textit{light} & \textit{ultraviolet} & \textit{infrared} & \textit{sunlight} & \textit{uv} & \textit{shadows} & \textit{bright} & \textit{illumination} & \textit{glow} & \textit{illuminate} & {\color{cyan}\textit{[spectrum]}} \\
 & {\color{red}53} & \textit{salts} & \textit{solvents} & \textit{chlorine} & \textit{hydrogen} & \textit{inorganic} & \textit{ammonia} & \textit{chloride} & \textit{flammable} & \textit{sulfide} & \textit{sulfur} & {\color{red}\textit{[chemistry]}} \\
 & 1 & \textit{.} & \textit{but} & \textit{though} & \textit{although} & \textit{,} & \textit{even} & \textit{that} & \textit{instance} & \textit{both} & \textit{however} & \textit{[function words]} \\
 & 153 & \textit{welterweight} & \textit{heavyweight} & \textit{middleweight} & \textit{featherweight} & \textit{ibf} & \textit{wbc} & \textit{holyfield} & \textit{wba} & \textit{cruiserweight} & \textit{bout} & \textit{[boxing]} \\
 & 158 & \textit{gendarmerie} & \textit{force} & \textit{contingent} & \textit{contingents} & \textit{detachments} & \textit{500-strong} & \textit{detachment} & \textit{constabulary} & \textit{5,000-strong} & \textit{constables} & \textit{[police]} \\
\bottomrule
\end{tabular}
\caption{
For the normalized ICA-transformed GloVe embeddings of \textit{light} and the component-wise products \textit{ultraviolet $\odot$ light} in Table~\ref{tab:table2_5columns}, the axes of the top 5 values are focused on, and their top 10 words are shown.
Axes are sorted by component values.
The meanings of the axes are interpreted from these listed words.
}
\label{tab:intro-topwords_table1}
\end{table*}
\renewcommand{\arraystretch}{1.0}

\begin{figure}[t!]
\centering
\begin{subfigure}{\columnwidth}
\centering
    \includegraphics[width=\columnwidth]{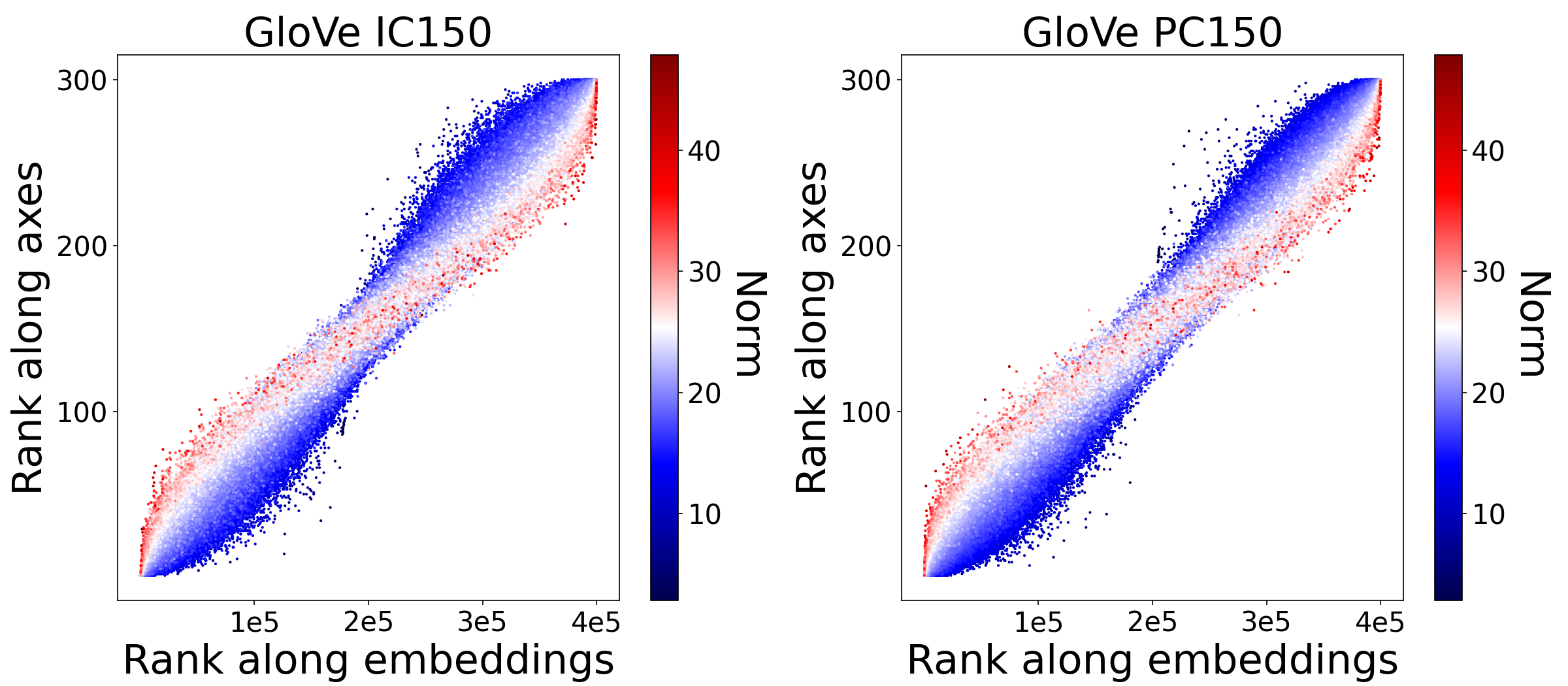}
    \subcaption{Before Normalization}
    \label{fig:inner_rank_before_149}
\end{subfigure}
\par\bigskip
\begin{subfigure}{\columnwidth}
\centering
    \includegraphics[width=\columnwidth]{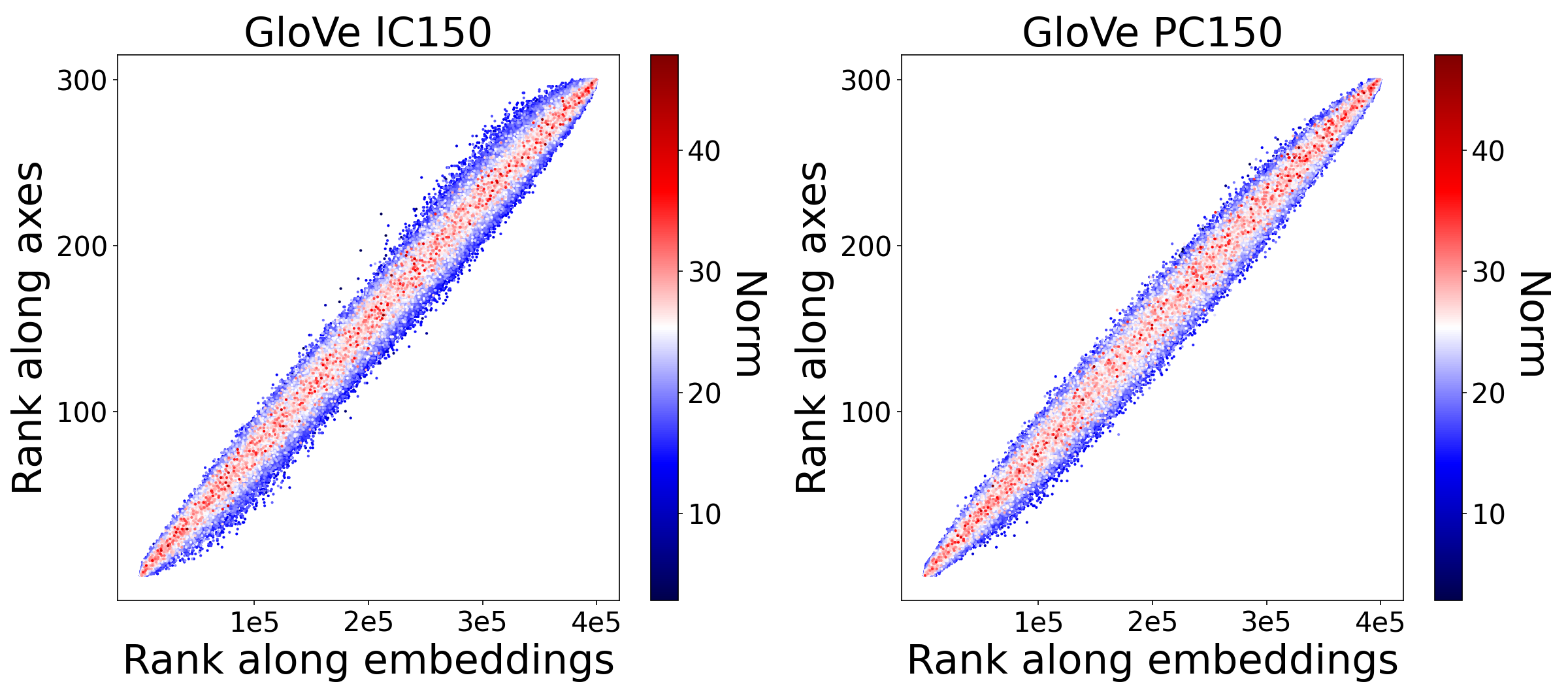}
    \subcaption{After Normalization}
    \label{fig:inner_rank_after_149}
\end{subfigure}
    \caption{
Scatterplots for the 150th axis of the (left) ICA-transformed and (right) PCA-transformed GloVe embeddings, (a) before and (b) after normalization, showing the rankings of component values along the embeddings and those along the axes, colored by the norms.
The larger the norm of an embedding, the more it is plotted in the foreground.
See Appendix~\ref{app:comp_cross_embeds_normalization} for contextualized embeddings.
}
\label{fig:inner_rank_149}
\end{figure}

\section{Interpretation of ICA Components for \textit{ultraviolet}, \textit{light}, and \textit{ultraviolet} \texorpdfstring{$\odot$}{odot} \textit{light}}\label{app:cosine}
\paragraph{\textit{ultraviolet}.}
For the normalized ICA-transformed and PCA-transformed GloVe embeddings of \textit{ultraviolet} in Fig.~\ref{fig:cosine}, Table~\ref{tab:intro-topwords} shows the top 10 words for the axes of the top 5 component values. The semantic interpretations of the axes are provided for the ICA-transformed embeddings, but such interpretations are challenging for the PCA-transformed embeddings.

Based on Table~\ref{tab:intro-topwords}, Fig.~\ref{fig:cosine_examples} illustrates our interpretation of cosine similarity between \textit{ultraviolet} and the top word of each axis: \textit{salts}, \textit{proteins}, \textit{spacecraft}, \textit{light}, and \textit{virus}.

\paragraph{\textit{light} and \textit{ultraviolet} $\odot$ \textit{light}.}
Similar to Table~\ref{tab:intro-topwords}, Table~\ref{tab:intro-topwords_table1} shows the interpretation of ICA components for \textit{light} and \textit{ultraviolet} $\odot$ \textit{light}.
The meaning of each axis in Table~\ref{tab:table2_5columns} is derived from Tables~\ref{tab:intro-topwords} and~\ref{tab:intro-topwords_table1}.

\section{Normalization of Embeddings: Fidelity of Component Rankings}\label{app:normalization}
As shown in Fig.~\ref{fig:intro}, although the meanings of the axes of the ICA-transformed and PCA-transformed embeddings can be interpreted from the top words, the rankings of component values along the embeddings can change before and after normalization. 
Therefore, we compare the rankings along the embeddings with those along the axes, both before and after normalization.

Figure~\ref{fig:inner_rank_149} shows scatterplots of these two rankings for the component values of the 150th axis of the ICA-transformed and PCA-transformed GloVe embeddings, both before and after normalization. 
Before normalization, as shown in Fig.~\ref{fig:inner_rank_before_149}, embeddings with high rankings along the embeddings tend to have lower rankings along the axes as their norms increase. 
After normalization, as shown in Fig.~\ref{fig:inner_rank_after_149}, the norm-derived artifacts observed in Fig.~\ref{fig:inner_rank_before_149} disappear. 
These results support the improvement in interpretability through normalization, as discussed in Section~\ref{sec:intruder}.

\section{Distribution of Cosine Similarity, Component Values and Their Products} \label{app:distribution-theory}

In this section, vectors are denoted simply as $X$ instead of in boldface as $\mathbf{x}$, and the elements of the vector $X$ are denoted using the subscript $X_\ell$ rather than the superscript $X^{(\ell)}$.

\subsection{Cosine similarity} \label{app:dist-cosine-similarity}
Let us consider two random vectors $X=(X_1,\ldots,X_d), Y=(Y_1,\ldots,Y_d) \in \mathbb{R}^d$ with elements of mean zero $\mathbb{E}(X_\ell)=\mathbb{E}(Y_\ell)=0$ and variance one $\mathbb{E}(X_\ell^2)=\mathbb{E}(Y_\ell^2)=1$. 
We assume that the elements $X_1,\ldots,X_d$ and $Y_1,\ldots,Y_d$ are independent.
Then, for sufficiently large $d$, the cosine similarity $\cos(X,Y)$ asymptotically follows $\mathcal{N}(0,1/d)$, the normal distribution with mean 0 and variance $1/d$,
\begin{equation} \label{eq:clt-cosine-similarity}
\cos(X,Y) \sim \mathcal{N}(0,1/d).
\end{equation}
This is easily shown as follows; a more general argument can be found in Appendix~C of \citet{DBLP:journals/corr/abs-2401-06112}.
First note that $\mathbb{E}(X_\ell Y_\ell)=\mathbb{E}(X_\ell)\mathbb{E}(Y_\ell)=0$, $\mathbb{E}(X_\ell^2Y_\ell^2)=\mathbb{E}(X_\ell^2)\mathbb{E}(Y_\ell^2)=1$. Thus the inner product, if scaled by dimension, $d^{-1/2} \langle X, Y \rangle = d^{-1/2}  \sum_{\ell=1}^d X_\ell Y_\ell$ has mean zero and variance one.
Thus, according to the central limit theorem, 
\begin{equation}  \label{eq:clt-inner-product}
  d^{-1/2} \langle X, Y \rangle \sim \mathcal{N}(0,1)
\end{equation}
for sufficiently large $d$.
It also follows from the law of large numbers that $d^{-1} \| X \|^2 = d^{-1} \sum_{\ell=1}^d X_\ell^2$ converges in probability to $\mathbb{E}(X_\ell^2)=1$, and similarly $d^{-1} \| Y \|^2 \to 1$ in probability. Therefore,
\[
\sqrt{d} \cos(X,Y) =\frac{d^{-1/2} \langle X, Y \rangle } {\sqrt{d^{-1}\|X\|^2}\sqrt{d^{-1}\|Y\|^2}}
\]
converges to (\ref{eq:clt-inner-product}), thereby concluding (\ref{eq:clt-cosine-similarity}).

\begin{figure}[t!]
    \centering
    \includegraphics[width=\linewidth]{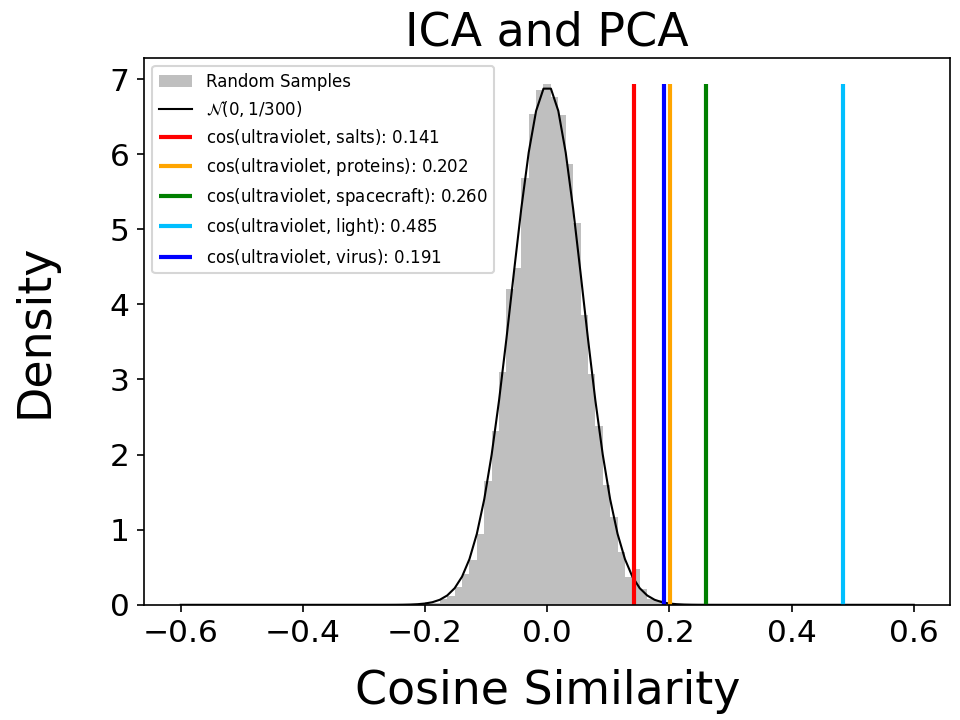}
    \caption{For {10,000} randomly sampled pairs of ICA-transformed embeddings $\mathbf{s}, \mathbf{s}' \in \mathbb{R}^d$, the histogram of the cosine similarity $\cos(\mathbf{s},\mathbf{s}')$ is displayed. Since cosine similarities are invariant under the orthogonal transformation, exactly the same plot is also obtained from PCA-transformed embeddings.
The theoretical probability density of (\ref{eq:clt-cosine-similarity}) is almost identical to the observed histogram.
The theory in Appendix~\ref{app:distribution-theory} is also supported by the inverse of the variance, ${309.663}\approx d$ for $d=300$.
The observed cosine similarities in Fig.~\ref{fig:cosine_examples} are also indicated as vertical lines.}
    \label{fig:cossim_hist}
\end{figure}

\subsection{Component values}\label{app:dist-components}
Let $S=(S_1,\ldots,S_d) \in \mathbb{R}^d$ be a random vector representing ICA-transformed embeddings, and let $e_\ell = (0,\ldots,0,1,0,\ldots,0) \in \mathbb{R}^d$ be the one-hot vector with a one at the $\ell$-th element.
Then, $\hat S_\ell = \cos(S,e_\ell)$ represents the $\ell$-th component of the normalized ICA-transformed embeddings.
Although $e_\ell$ is not a random vector, formally letting $X=S$ and $Y=e_\ell$ in (\ref{eq:clt-cosine-similarity}) gives
\begin{equation}  \label{eq:dist-component}
\hat S_\ell \sim \mathcal{N}(0,1/d)
\end{equation}
for sufficiently large $d$.
Considering $S$ and $e_\ell$ in the original coordinate system before the ICA-transformation, they can be regarded as almost random, which suggests that the formal argument above is valid.

\subsection{Product of two component values}\label{app:dist-products}

\begin{figure*}[t!]
    \centering
    \begin{minipage}{0.33\linewidth}
        \centering
        \includegraphics[width=\linewidth]{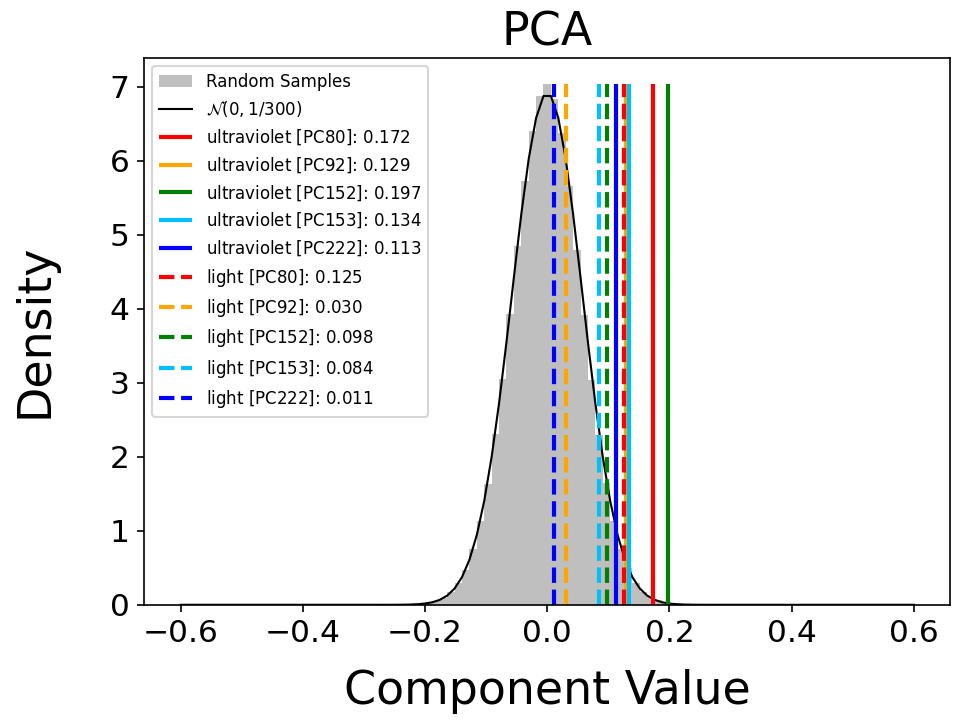}
        \subcaption{Components}
        \label{fig:comp_hist_pca}
    \end{minipage}\hfill
    \begin{minipage}{0.33\linewidth}
        \centering
        \includegraphics[width=\linewidth]{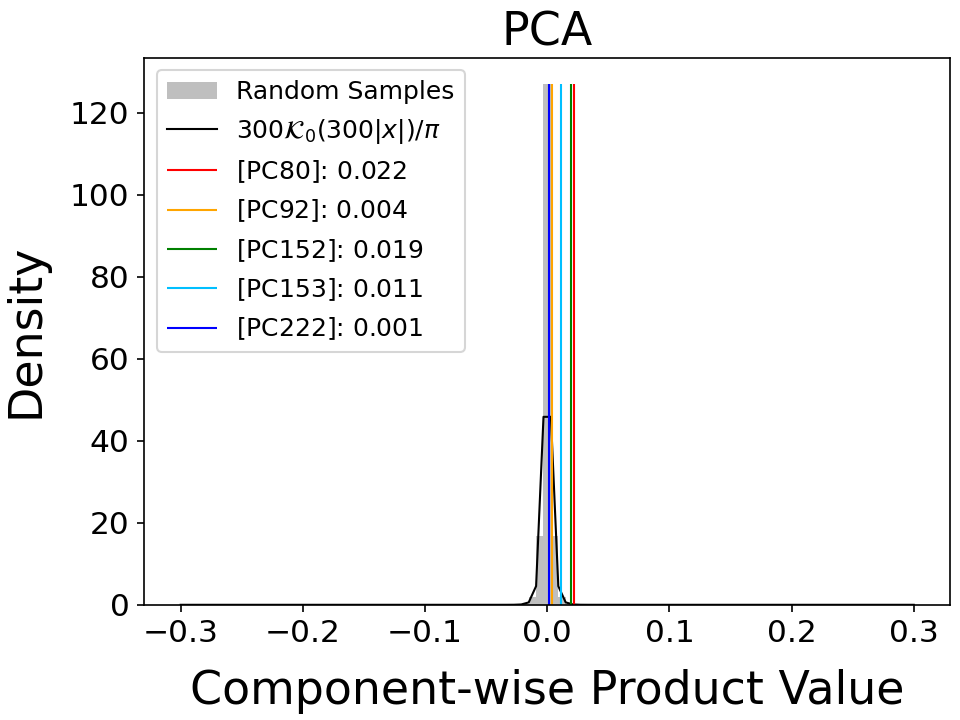}
        \subcaption{Component-wise products}
        \label{fig:comp_prod_hist_pca}
    \end{minipage}
    \begin{minipage}{0.33\linewidth}
        \centering
        \includegraphics[width=\linewidth]{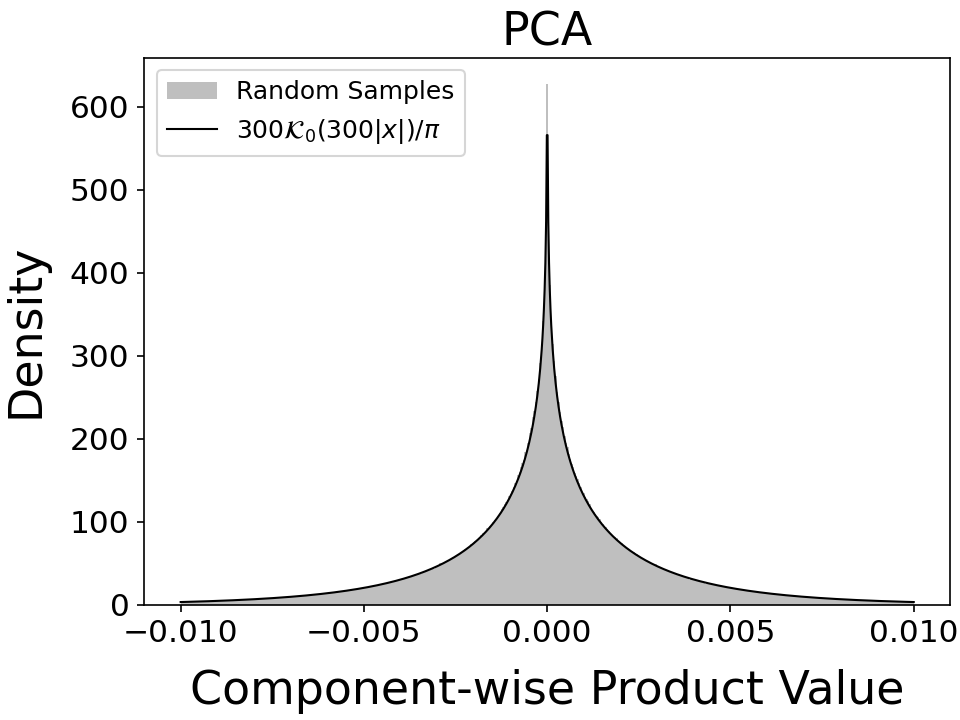}
        \subcaption{Component-wise products (magnified)}
        \label{fig:comp_prod_hist_zoom_pca}
    \end{minipage}
    \caption{
    The plots for the normalized PCA-transformed GloVe embeddings are displayed in the same manner as
    in Fig.~\ref{fig:histgram-ica_maintext}.
    The inverse of the variance is $300.000\approx d$ in (a) and $90{,}108.283\approx d^2$ in (b, c) for $d=300$.
    }
\label{fig:histgram-pca}
\end{figure*}

\begin{table*}[t!]
\small
\centering
\begin{subtable}{.33\textwidth}
  \centering
  \begin{adjustbox}{max width=0.93\linewidth}
  \begin{tabular}{rcrrr}
  \toprule
  Axis & Value & $p$-value & Bonferroni \\
  \midrule
{\color{green}\num{152}} & \num{0.197} & \num{3.21e-04} & \num{9.62e-02} \\
{\color{red}\num{80}} & \num{0.172} & \num{1.41e-03} & \num{4.22e-01} \\
{\color{cyan}\num{153}} & \num{0.134} & \num{1.01e-02} & \num{1.00e+00} \\
{\color{orange}\num{92}} & \num{0.129} & \num{1.30e-02} & \num{1.00e+00} \\
{\color{blue}\num{222}} & \num{0.113} & \num{2.52e-02} & \num{1.00e+00} \\
  \bottomrule
  \end{tabular}
  \end{adjustbox}
  \caption{\textit{ultraviolet}}
\end{subtable}
\begin{subtable}{.33\textwidth}
  \centering
  \begin{adjustbox}{max width=0.93\linewidth}
  \begin{tabular}{rcrrr}
  \toprule
  Axis & Value & $p$-value & Bonferroni \\
  \midrule
\num{1} & \num{0.247} & \num{9.12e-06} & \num{2.74e-03} \\
\num{5} & \num{0.216} & \num{9.38e-05} & \num{2.81e-02} \\
\num{61} & \num{0.176} & \num{1.15e-03} & \num{3.44e-01} \\
\num{118} & \num{0.176} & \num{1.16e-03} & \num{3.47e-01} \\
\num{127} & \num{0.146} & \num{5.76e-03} & \num{1.00e+00} \\
  \bottomrule
  \end{tabular}
  \end{adjustbox}
  \caption{\textit{light}}
\end{subtable}
\begin{subtable}{.33\textwidth}
  \centering
  \begin{adjustbox}{max width=0.93\linewidth}
  \begin{tabular}{rcrrr}
  \toprule
  Axis & Value & $p$-value & Bonferroni \\
  \midrule
\num{1} & \num{0.024} & \num{1.09e-04} & \num{3.26e-02} \\
{\color{red}\num{80}} & \num{0.022} & \num{2.21e-04} & \num{6.62e-02} \\
\num{131} & \num{0.020} & \num{3.80e-04} & \num{1.14e-01} \\
{\color{green}\num{152}} & \num{0.019} & \num{4.58e-04} & \num{1.38e-01} \\
\num{60} & \num{0.017} & \num{9.65e-04} & \num{2.90e-01} \\
  \bottomrule
  \end{tabular}
  \end{adjustbox}
  \caption{\textit{ultraviolet} $\odot$ \textit{light}}
\end{subtable}
  \caption{(a, b) show the observed component values of normalized PCA-transformed GloVe embeddings in Fig.~\ref{fig:cosine_pca}, while (c) shows their component-wise products.
The $p$-values and their Bonferroni-corrected values are shown for the top five axes in each table.
}
\label{tab:pvalues_pca}
\end{table*}

Let $\hat S=(\hat S_1,\ldots,\hat S_d), \hat S'=(\hat S_1',\ldots,\hat S_d') \in \mathbb{R}^d$ be two independent random vectors representing normalized ICA-transformed embeddings. We consider $Z_\ell = \hat S_\ell \hat S_\ell'$, the $\ell$-th element of the component-wise product of $\hat S$ and $\hat S'$. Then, for sufficiently large $d$, the probability density function of $Z_\ell$ is
\begin{equation}  \label{eq:dist-product}
    Z_\ell \sim (d/\pi) K_0(d |Z_\ell|),
\end{equation}
where $K_0(\cdot)$ is the modified Bessel function of the second kind of order zero.
This result follows directly from Theorem~2.1 of \citet{NADARAJAH2016201}, assuming that $\hat S_\ell$ and $\hat S_\ell'$ are independently distributed as $\mathcal{N}(0,1/d)$.
The mean of $Z_\ell$ is $\mathbb{E}(Z_\ell) = \mathbb{E}(\hat S_\ell \hat S_\ell') = \mathbb{E}(\hat S_\ell)  \mathbb{E}(\hat S_\ell') = 0 $ and
the variance of $Z_\ell$ is $\mathbb{E}(Z_\ell^2) = \mathbb{E}(\hat S_\ell^2 \hat S_\ell^{\prime2}) = \mathbb{E}(\hat S_\ell^2) \mathbb{E}(\hat S_\ell^{\prime2})=1/d^2$.

\subsection{Histogram of cosine similarity, component values and their products} \label{app:histogram}

\paragraph{Histogram of cosine similarity.}
The observed and theoretical distributions of the cosine similarities are exhibited in Fig.~\ref{fig:cossim_hist}. The histogram and the theoretical curve of the probability density function clearly suggest that the distribution theory in Appendix~\ref{app:dist-cosine-similarity} is strongly supported and that the cosine similarities are normally distributed with mean zero and variance $1/d$. However, when embeddings are centered but not whitened, which violates the assumption in Appendix~\ref{app:dist-cosine-similarity}, the variance should tend to be larger than $1/d$. In such a case, the inverse of the variance would provide an effective dimensionality of the embeddings.

\paragraph{Histogram of component values and component-wise products.}
The observed and theoretical distributions of the components and the component-wise products are exhibited in Fig.~\ref{fig:histgram-ica_maintext} for ICA and Fig.~\ref{fig:histgram-pca} for PCA.
The histograms and the theoretical curves of the probability density functions clearly suggest that the distribution theory in Appendices~\ref{app:dist-components} and \ref{app:dist-products} is strongly supported.
The components are normally distributed with mean zero and variance $1/d$, and the probability density function of the component-wise products is expressed by the modified Bessel function, where the mean is zero and the variance is $1/d^2$.
Interestingly, while the theory was originally provided for ICA-transformed embeddings, the experiments suggest that the theory also applies to the PCA-transformed embeddings.

\paragraph{Comparison between observed values and theoretical distributions.}
In Figs.~\ref{fig:cossim_hist}, \ref{fig:histgram-ica_maintext}, and \ref{fig:histgram-pca}, the observed values, indicated as vertical lines, are compared to their respective distributions to identify the significance of the values. By multiplying the values by the inverse of the standard deviations, these values can be easily interpreted. Thus, $\sqrt{d}\cos(\mathbf{s}, \mathbf{s}')$, $\sqrt{d} \hat s^{(\ell)}$, and $d \hat s^{(\ell)} \hat s^{(\ell)\prime}$ may be used for numerical comparisons.

Similar to Table~\ref{tab:table2_5columns}, Table~\ref{tab:pvalues_pca} presents the top five values from the bar graphs in Fig.~\ref{fig:cosine_pca}, along with their corresponding $p$-values and Bonferroni-corrected values. 
As shown in Fig.~\ref{fig:histgram-pca}, fewer axes are statistically significant with $\alpha=0.05$ compared to those in Table~\ref{tab:table2_5columns}.

\section{Details of ICA Consistency for Contextualized Embeddings}\label{sec:details_cross_embedding}
\begin{table}[t!]
\small
\centering
\begin{tabular}{@{\hspace{0.2em}}l@{\hspace{0.5em}}r@{\hspace{0.5em}}r@{\hspace{0.5em}}r@{\hspace{0.2em}}}
\toprule
Model & Layers & Dimensions & Parameters\\
\midrule
\texttt{bert-base-uncased} & \multirow{4}{*}{12} & \multirow{4}{*}{768} & 110M \\
\texttt{roberta-base} &  &  & 125M\\
\texttt{gpt2} &  & & 117M \\
\texttt{pythia-160m} & & & 160M \\
\bottomrule
\end{tabular}
\caption{
The number of layers, the dimensions, and the parameter size for each model.
}
\label{tab:models}
\end{table}

\begin{table*}[t!]
\centering
\begin{tabular}{l}
\toprule
The {\color{red} \textit{ultraviolet}} {\color{red} \textit{light}} used in tanning salons are now considered as carcinogenic as tobacco and ...\\
\bottomrule
\end{tabular}
\caption{
The sentence that contains both \textit{ultraviolet\_3} and \textit{light\_1} used in Fig.~\ref{fig:cross-embedding_bargraph_ica} in Section~\ref{sec:various_embeddings}.
Note that the sentence is lowercased when the embeddings are computed.
}
\label{tab:uv_light_sentences}
\end{table*}

\renewcommand{\arraystretch}{1.2} 
\begin{table*}[t!]
\small
\centering
\begin{subtable}[t]{\textwidth}
\centering
\begin{adjustbox}{max width=\textwidth}
\begin{tabular}{@{\hspace{0.25em}}c@{\hspace{0.25em}}|@{\hspace{0.25em}}r@{\hspace{0.25em}}|@{\hspace{0.15em}}c@{\hspace{0.15em}}c@{\hspace{0.15em}}c@{\hspace{0.15em}}c@{\hspace{0.15em}}c@{\hspace{0.5em}}c@{\hspace{0.15em}}c@{\hspace{0.15em}}c@{\hspace{0.15em}}c@{\hspace{0.15em}}c@{\hspace{0.15em}}|@{\hspace{0.15em}}c@{\hspace{0.15em}}}
\toprule
 &  Axis &          Top1 &          Top2 &          Top3 &          Top4 &         Top5 &          Top6 &         Top7 &        Top8 &         Top9 &        Top10 &                    Meaning \\
\midrule
\multirow{5}{*}{\rotatebox{90}{BERT-base}} 
 & {\color{red}286} & \textit{greenhouse\_0} & \textit{greenhouse\_2} & \textit{greenhouse\_4} & \textit{greenhouse\_1} & \textit{emissions\_2} & \textit{greenhouse\_3} & \textit{emissions\_5} & \textit{carbon\_5} & \textit{emissions\_6} & \textit{emissions\_0} & {\color{red}\textit{[environment]}} \\
 & {\color{orange}343} & \textit{before\_7} & \textit{shortly\_5} & \textit{shortly\_8} & \textit{before\_1} & \textit{before\_9} & \textit{shortly\_0} & \textit{within\_2} & \textit{before\_6} & \textit{prior\_1} & \textit{soon\_9} & {\color{orange}\textit{[timings]}} \\
 & {\color{green}373} & \textit{hue\_1} & \textit{colors\_4} & \textit{color\_0} & \textit{hue\_2} & \textit{color\_3} & \textit{colors\_2} & \textit{colors\_3} & \textit{color\_2} & \textit{color\_4} & \textit{color\_1} & {\color{green}\textit{[colors]}} \\
 & {\color{cyan}470} & \textit{adhere\_0} & \textit{size\_9} & \textit{soft\_0} & \textit{drop\_5} & \textit{output\_0} & \textit{reserve\_2} & \textit{against\_8} & \textit{band\_0} & \textit{apply\_3} & \textit{flavor\_0} & {\color{cyan}\textit{[properties]}} \\
 & {\color{blue}634} & \textit{afternoon\_1} & \textit{midday\_0} & \textit{afternoon\_6} & \textit{morning\_3} & \textit{midday\_1} & \textit{morning\_8} & \textit{daily\_7} & \textit{daylight\_0} & \textit{morning\_6} & \textit{evenings\_0} & {\color{blue}\textit{[timeframes]}} \\
\midrule
\multirow{5}{*}{\rotatebox{90}{RoBERTa-base}}
 & {\color{red}80} & \textit{greenhouse\_1} & \textit{carbon\_8} & \textit{emissions\_5} & \textit{greenhouse\_2} & \textit{greenhouse\_0} & \textit{emissions\_0} & \textit{climate\_4} & \textit{climate\_2} & \textit{gases\_0} & \textit{emissions\_6} & {\color{red}\textit{[environment]}} \\
 & {\color{orange}219} & \textit{photos\_2} & \textit{images\_2} & \textit{images\_3} & \textit{pictures\_3} & \textit{photos\_1} & \textit{image\_2} & \textit{photos\_0} & \textit{images\_0} & \textit{photos\_3} & \textit{image\_7} & {\color{orange}\textit{[images]}} \\
 & {\color{green}330} & \textit{wireless\_3} & \textit{wireless\_7} & \textit{wireless\_1} & \textit{wireless\_0} & \textit{networks\_3} & \textit{broadband\_1} & \textit{broadband\_2} & \textit{wireless\_4} & \textit{connectivity\_0} & \textit{networks\_0} & {\color{green}\textit{[networks]}} \\
 & {\color{cyan}350} & \textit{green\_5} & \textit{hue\_0} & \textit{blue\_0} & \textit{colors\_0} & \textit{color\_5} & \textit{muted\_0} & \textit{colored\_1} & \textit{gray\_3} & \textit{oured\_0} & \textit{color\_4} & {\color{cyan}\textit{[colors]}} \\
 & {\color{blue}755} & \textit{whole\_3} & \textit{grain\_0} & \textit{light\_1} & \textit{whole\_4} & \textit{guardian\_1} & \textit{grains\_2} & \textit{emb\_0} & \textit{guardian\_3} & \textit{fiber\_0} & \textit{ju\_0} & {\color{blue}\textit{[grains]}} \\
\midrule
\multirow{5}{*}{\rotatebox{90}{GPT-2}} 
 & {\color{red}28} & \textit{to\_0} & \textit{attracts\_0} & \textit{in\_0} & \textit{specialized\_0} & \textit{genes\_0} & \textit{ting\_0} & \textit{cause\_5} & \textit{adopt\_0} & \textit{ocytes\_0} & \textit{produced\_1} & {\color{red}\textit{[biology]}} \\
 & {\color{orange}273} & \textit{arsenic\_0} & \textit{nitrogen\_1} & \textit{phosphorus\_0} & \textit{nitrogen\_0} & \textit{ate\_4} & \textit{dioxide\_0} & \textit{calcium\_0} & \textit{carbon\_9} & \textit{opes\_0} & \textit{fumes\_0} & {\color{orange}\textit{[elements]}} \\
 & {\color{green}349} & \textit{ultraviolet\_4} & \textit{light\_1} & \textit{-\_0} & \textit{light\_0} & \textit{accurate\_1} & \textit{-\_1} & \textit{ultraviolet\_1} & \textit{color\_4} & \textit{ultraviolet\_3} & \textit{ultraviolet\_0} & {\color{green}\textit{[light]}} \\
 & {\color{cyan}512} & \textit{domain\_1} & \textit{computer\_2} & \textit{identify\_1} & \textit{bypass\_0} & \textit{mail\_6} & \textit{web\_7} & \textit{internet\_8} & \textit{allows\_6} & \textit{user\_2} & \textit{mail\_5} & {\color{cyan}\textit{[technology]}} \\
 & {\color{blue}619} & \textit{restaurant\_0} & \textit{proposal\_2} & \textit{project\_3} & \textit{project\_5} & \textit{soldier\_2} & \textit{farmer\_0} & \textit{scheme\_6} & \textit{designer\_1} & \textit{company\_7} & \textit{company\_6} & {\color{blue}\textit{[occupations]}} \\
\midrule
\multirow{5}{*}{\rotatebox{90}{Pythia-160m}} 
 & {\color{red}262} & \textit{light\_1} & \textit{ultraviolet\_3} & \textit{light\_0} & \textit{ultraviolet\_4} & \textit{ultraviolet\_1} & \textit{-\_0} & \textit{ultraviolet\_0} & \textit{-\_1} & \textit{ultraviolet\_2} & \textit{light\_0} & {\color{red}\textit{[light]}} \\
 & {\color{orange}371} & \textit{aud\_6} & \textit{aud\_8} & \textit{aud\_0} & \textit{aud\_2} & \textit{aud\_3} & \textit{aud\_1} & \textit{aud\_7} & \textit{aud\_4} & \textit{aud\_5} & \textit{jan\_4} & {\color{orange}\textit{[subwords]}} \\
 & {\color{green}485} & \textit{for\_2} & \textit{and\_3} & \textit{both\_0} & \textit{and\_2} & \textit{and\_1} & \textit{,\_4} & \textit{of\_2} & \textit{,\_5} & \textit{limited\_0} & \textit{)\_0} & {\color{green}\textit{[function words]}} \\
 & {\color{cyan}619} & \textit{scheme\_6} & \textit{restaurant\_0} & \textit{campaign\_3} & \textit{discovery\_2} & \textit{investor\_1} & \textit{resolution\_2} & \textit{island\_0} & \textit{let\_5} & \textit{designer\_1} & \textit{shoes\_4} & {\color{cyan}\textit{[business]}} \\
 & {\color{blue}660} & \textit{special\_7} & \textit{special\_4} & \textit{special\_3} & \textit{special\_8} & \textit{special\_2} & \textit{special\_6} & \textit{special\_1} & \textit{extraordinary\_2} & \textit{special\_9} & \textit{special\_5} & {\color{blue}\textit{[exceptional]}} \\
\bottomrule
\end{tabular}
\end{adjustbox}
\caption{Normalized ICA-transformed contextualized embeddings}
\label{tab:4models_topwords_ica}
\end{subtable}

\vspace{1em}

\begin{subtable}[t]{\textwidth}
\centering
\begin{adjustbox}{max width=\textwidth}
\begin{tabular}{@{\hspace{0.25em}}c@{\hspace{0.25em}}|@{\hspace{0.25em}}r@{\hspace{0.25em}}|@{\hspace{0.15em}}c@{\hspace{0.15em}}c@{\hspace{0.15em}}c@{\hspace{0.15em}}c@{\hspace{0.15em}}c@{\hspace{0.5em}}c@{\hspace{0.15em}}c@{\hspace{0.15em}}c@{\hspace{0.15em}}c@{\hspace{0.15em}}c@{\hspace{0.15em}}|@{\hspace{0.15em}}c@{\hspace{0.15em}}}
\toprule
 &  Axis &          Top1 &          Top2 &          Top3 &          Top4 &         Top5 &          Top6 &         Top7 &        Top8 &         Top9 &        Top10 &                    Meaning \\
\midrule
\multirow{5}{*}{\rotatebox{90}{BERT-base}} 
 & {\color{red}45} & \textit{150\_0} & \textit{110\_0} & \textit{drilling\_0} & \textit{saddam\_0} & \textit{september\_0} & \textit{10\_1} & \textit{september\_7} & \textit{sunni\_8} & \textit{mist\_4} & \textit{saddam\_1} & {\color{red}\textit{[PC45]}} \\
 & {\color{orange}58} & \textit{bodyguards\_1} & \textit{danger\_0} & \textit{13\_1} & \textit{threatening\_1} & \textit{angeles\_1} & \textit{castillo\_0} & \textit{tackle\_1} & \textit{bodyguards\_0} & \textit{angeles\_7} & \textit{13\_0} & {\color{orange}\textit{[PC58]}} \\
 & {\color{green}85} & \textit{deutschland\_0} & \textit{equipment\_6} & \textit{explosions\_2} & \textit{germany\_8} & \textit{germany\_5} & \textit{germany\_7} & \textit{berlin\_1} & \textit{berlin\_0} & \textit{investors\_5} & \textit{munich\_2} & {\color{green}\textit{[PC85]}} \\
 & {\color{cyan}91} & \textit{jones\_4} & \textit{prior\_1} & \textit{h\_5} & \textit{side\_9} & \textit{evidence\_4} & \textit{present\_3} & \textit{burns\_1} & \textit{surgeon\_1} & \textit{news\_0} & \textit{video\_6} & {\color{cyan}\textit{[PC91]}} \\
 & {\color{blue}473} & \textit{tongue\_1} & \textit{spend\_0} & \textit{matter\_6} & \textit{romney\_1} & \textit{schedule\_1} & \textit{specify\_1} & \textit{weekly\_0} & \textit{finding\_5} & \textit{.\_7} & \textit{era\_7} & {\color{blue}\textit{[PC473]}} \\
\midrule
\multirow{5}{*}{\rotatebox{90}{RoBERTa-base}} 
 & {\color{red}38} & \textit{examples\_0} & \textit{easy\_4} & \textit{expressed\_3} & \textit{offset\_2} & \textit{peaked\_1} & \textit{extreme\_2} & \textit{sketches\_0} & \textit{rare\_4} & \textit{spike\_0} & \textit{expressed\_1} & {\color{red}\textit{[PC38]}} \\
 & {\color{orange}278} & \textit{returned\_7} & \texttt{\textbackslash xEF\textbackslash xBF\textbackslash xBD}\_2 & \texttt{\textbackslash xEF\textbackslash xBF\textbackslash xBD}\_0 & \textit{worry\_3} & \textit{stories\_3} & \textit{returned\_4} & \texttt{\textbackslash xEF\textbackslash xBF\textbackslash xBD}\_3 & \textit{/\_3} & \textit{j\_0} & \textit{ight\_0} & {\color{orange}\textit{[PC278]}} \\
 & {\color{green}517} & \textit{culture\_4} & \textit{worms\_0} & \textit{examined\_0} & \textit{machine\_8} & \textit{crew\_3} & \textit{architecture\_1} & \textit{rising\_9} & \textit{behalf\_0} & \textit{style\_1} & \textit{igator\_0} & {\color{green}\textit{[PC517]}} \\
 & {\color{cyan}682} & \textit{il\_0} & \textit{ung\_6} & \textit{fire\_0} & \textit{he\_8} & \textit{side\_9} & \textit{situation\_1} & \textit{dj\_2} & \textit{colleague\_0} & \textit{association\_6} & \textit{free\_4} & {\color{cyan}\textit{[PC682]}} \\
 & {\color{blue}729} & \textit{pick\_0} & \textit{sn\_4} & \textit{class\_4} & \textit{flo\_0} & \textit{bacon\_0} & \textit{agreement\_4} & \textit{apiece\_0} & \textit{caught\_9} & \textit{defensively\_0} & \textit{mble\_0} & {\color{blue}\textit{[PC729]}} \\
\midrule
\multirow{5}{*}{\rotatebox{90}{GPT-2}} 
 & {\color{red}128} & \textit{airlines\_0} & \textit{ids\_0} & \textit{buy\_7} & \textit{bought\_3} & \textit{pleaded\_2} & \textit{club\_5} & \textit{flights\_4} & \textit{values\_4} & \textit{aff\_2} & \textit{pleading\_0} & {\color{red}\textit{[PC128]}} \\
 & {\color{orange}254} & \textit{exchange\_5} & \textit{21\_0} & \textit{international\_6} & \textit{telephone\_0} & \textit{political\_9} & \textit{mistakes\_0} & \textit{wine\_9} & \textit{partially\_2} & \textit{holder\_0} & \textit{diplomatic\_0} & {\color{orange}\textit{[PC254]}} \\
 & {\color{green}374} & \textit{nine\_3} & \textit{ib\_5} & \textit{divorced\_1} & \textit{14\_8} & \textit{ultraviolet\_1} & \textit{ib\_7} & \textit{interviewed\_1} & \textit{appealing\_0} & \textit{ane\_9} & \textit{of\_1} & {\color{green}\textit{[PC374]}} \\
 & {\color{cyan}560} & \textit{forcing\_7} & \textit{14\_5} & \textit{results\_4} & \textit{az\_0} & \textit{kn\_3} & \textit{mont\_6} & \textit{really\_6} & \textit{fighting\_6} & \textit{rs\_0} & \textit{tre\_0} & {\color{cyan}\textit{[PC560]}} \\
 & {\color{blue}763} & \textit{these\_3} & \textit{company\_1} & \textit{ful\_9} & \textit{here\_4} & \textit{ki\_0} & \textit{said\_1} & \textit{ash\_9} & \textit{ec\_9} & \textit{gg\_2} & \textit{aud\_0} & {\color{blue}\textit{[PC763]}} \\
\midrule
\multirow{5}{*}{\rotatebox{90}{Pythia-160m}}
 & {\color{red}171} & \textit{rev\_2} & \textit{rate\_6} & \textit{government\_2} & \textit{intensity\_0} & \textit{vibration\_0} & \textit{regiment\_0} & \textit{ash\_3} & \textit{pet\_1} & \textit{2\_4} & \textit{steady\_2} & {\color{red}\textit{[PC171]}} \\
 & {\color{orange}264} & \textit{body\_1} & \textit{v\_4} & \textit{radiation\_1} & \textit{victory\_3} & \textit{article\_6} & \textit{article\_0} & \textit{article\_5} & \textit{concrete\_1} & \textit{62\_0} & \textit{yield\_1} & {\color{orange}\textit{[PC264]}} \\
 & {\color{green}270} & \textit{rain\_3} & \textit{ideology\_2} & \textit{possible\_3} & \textit{bucks\_0} & \textit{ula\_2} & \textit{populated\_0} & \textit{damp\_0} & \textit{nick\_3} & \textit{vis\_0} & \textit{vis\_2} & {\color{green}\textit{[PC270]}} \\
 & {\color{cyan}493} & \textit{expect\_8} & \textit{farms\_2} & \textit{sugar\_0} & \textit{and\_0} & \textit{else\_8} & \textit{slightly\_6} & \textit{mel\_1} & \textit{orb\_0} & \textit{sensations\_0} & \textit{both\_8} & {\color{cyan}\textit{[PC493]}} \\
 & {\color{blue}635} & \textit{book\_2} & \textit{selling\_0} & \textit{kil\_3} & \textit{undergoing\_0} & \textit{if\_5} & \textit{if\_4} & \textit{if\_0} & \textit{if\_7} & \textit{if\_1} & \textit{if\_3} & {\color{blue}\textit{[PC635]}} \\
\bottomrule
\end{tabular}
\end{adjustbox}
\caption{Normalized PCA-transformed contextualized embeddings}

\label{tab:4models_topwords_pca}
\end{subtable}
\caption{
The top 10 words for the axes of the top 5 component values of the normalized embeddings of \textit{ultraviolet\_3} in Figs.~\ref{fig:cross-embedding_bargraph_ica},~\ref{fig:cross-embedding_bargraph_ica_bert_roberta}, and~\ref{fig:cross-embedding_bargraph_pca}.
We also provide the semantic interpretations of these axes.
Note that we removed the prefix \#\# for BERT and the prefix Ġ for other models.
Note that all instances of the corrupted character in the top words of RoBERTa in Table~\ref{tab:4models_topwords_pca} were replaced with the Unicode character \texttt{\textbackslash xEF\textbackslash xBF\textbackslash xBD}.
}
\label{tab:4models_topwords}
\end{table*}
\renewcommand{\arraystretch}{1.0}

\subsection{Experimental settings}\label{app:cross-embeds-settings}
To conduct experiments with contextualized embedding models, we used the \texttt{transformers} library~\cite{DBLP:conf/emnlp/WolfDSCDMCRLFDS20}. The models used are listed in Table~\ref{tab:models}.

Following~\citet{DBLP:conf/emnlp/YamagiwaOS23}, we selected the One Billion Word Benchmark~\cite{DBLP:conf/interspeech/ChelbaMSGBKR14} as the corpus for our experiments.
To compute the embeddings, we used the sentences in the first file\footnote{\texttt{news.en-00001-of-00100}} of the dataset's training data, converted to lowercase.
It is important to note that, for computing the embeddings of \textit{ultraviolet} in Fig.~\ref{fig:cross-embedding_bargraph_ica}, we prioritized sentences in the file that contain the word \textit{ultraviolet}\footnote{The word \textit{light} were included without prioritizing specific sentences.}.
Then, embeddings were computed sequentially from the remaining sentences.
To avoid the overrepresentation of high-frequency words during the ICA transformation, we computed up to 10 embeddings for each token, resulting in a total of $50{,}000$ embeddings.
Each token was distinguished as \textit{ultraviolet\_0}, \textit{ultraviolet\_1}, $\ldots$, and so on.

For the ICA transformation of the contextualized embeddings of each model, we set the hyperparameters to the same values used for the GloVe embeddings: a maximum number of iterations of $10{,}000$ and a convergence tolerance of $1\times 10^{-10}$.

\subsection{Details of the heatmaps in Fig.~\ref{fig:cross-embedding_heatmap}}
We explain the details of the heatmaps in Fig.~\ref{fig:cross-embedding_heatmap} using the ICA-transformed embeddings, and apply the same procedure to the PCA-transformed embeddings.

\paragraph{Axis matching.}
As a preprocessing step, for each model's 50,000 tokens, the prefix \#\# is removed for BERT, and the prefix Ġ is removed for other models, if present.
Then, if a word in the GloVe vocabulary is included in the token sets of each model, the word and its corresponding tokens are paired.
If there are multiple corresponding tokens, multiple pairs are created.
Using these pairs, the correlation coefficients between the axes of the ICA-transformed GloVe embeddings and those of the ICA-transformed contextualized embeddings are computed.
The greedy algorithm by \citet{DBLP:conf/emnlp/YamagiwaOS23} for matching these axes is then applied using these correlation coefficients.
For further details, refer to \citet{DBLP:conf/emnlp/YamagiwaOS23}.

\paragraph{Selection of words and tokens.}
After matching these axes, for the axes of GloVe, we selected the top five words with the largest components among the words that are paired with the tokens of all other models. 
For each model, if multiple pairs existed, the token with the largest component on the corresponding axis was selected.

\subsection{Details of the bar graphs by the embeddings of \textit{ultraviolet\_3} and \textit{light\_1}}
\subsubsection{Details of \textit{ultraviolet\_3} and \textit{light\_1}}
Table~\ref{tab:uv_light_sentences} shows the sentence that contains both \textit{ultraviolet\_3} and \textit{light\_1} used in Fig.~\ref{fig:cross-embedding_bargraph_ica} in Section~\ref{sec:various_embeddings}. Among all the pairs of \textit{ultraviolet} and \textit{light} in the corpus (see Appendix~\ref{app:cross-embeds-settings} for details), the pair of \textit{ultraviolet\_3} and \textit{light\_1} had the highest average cosine similarity computed by the ICA-transformed embeddings of the four models.

\begin{figure*}[t!]
    \centering
    \includegraphics[width=\textwidth]{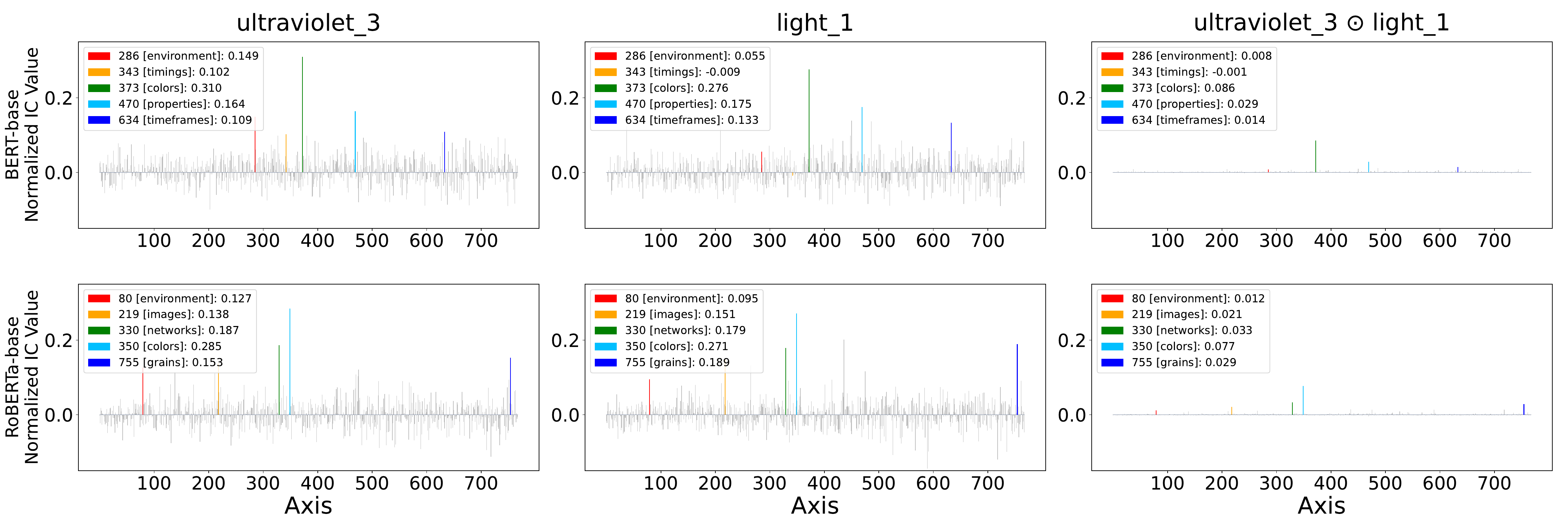}
    \caption{
Similar to Fig.~\ref{fig:cosine_ica}, the component values and their component-wise products for the normalized ICA-transformed contextualized embeddings of \textit{ultraviolet\_3} and \textit{light\_1} are shown in bar graphs. The cosine similarity is 0.532 for BERT-base and 0.608 for RoBERTa-base.
}
\label{fig:cross-embedding_bargraph_ica_bert_roberta}
\end{figure*}

\begin{figure*}[t!]
    \centering
    \includegraphics[width=\textwidth]{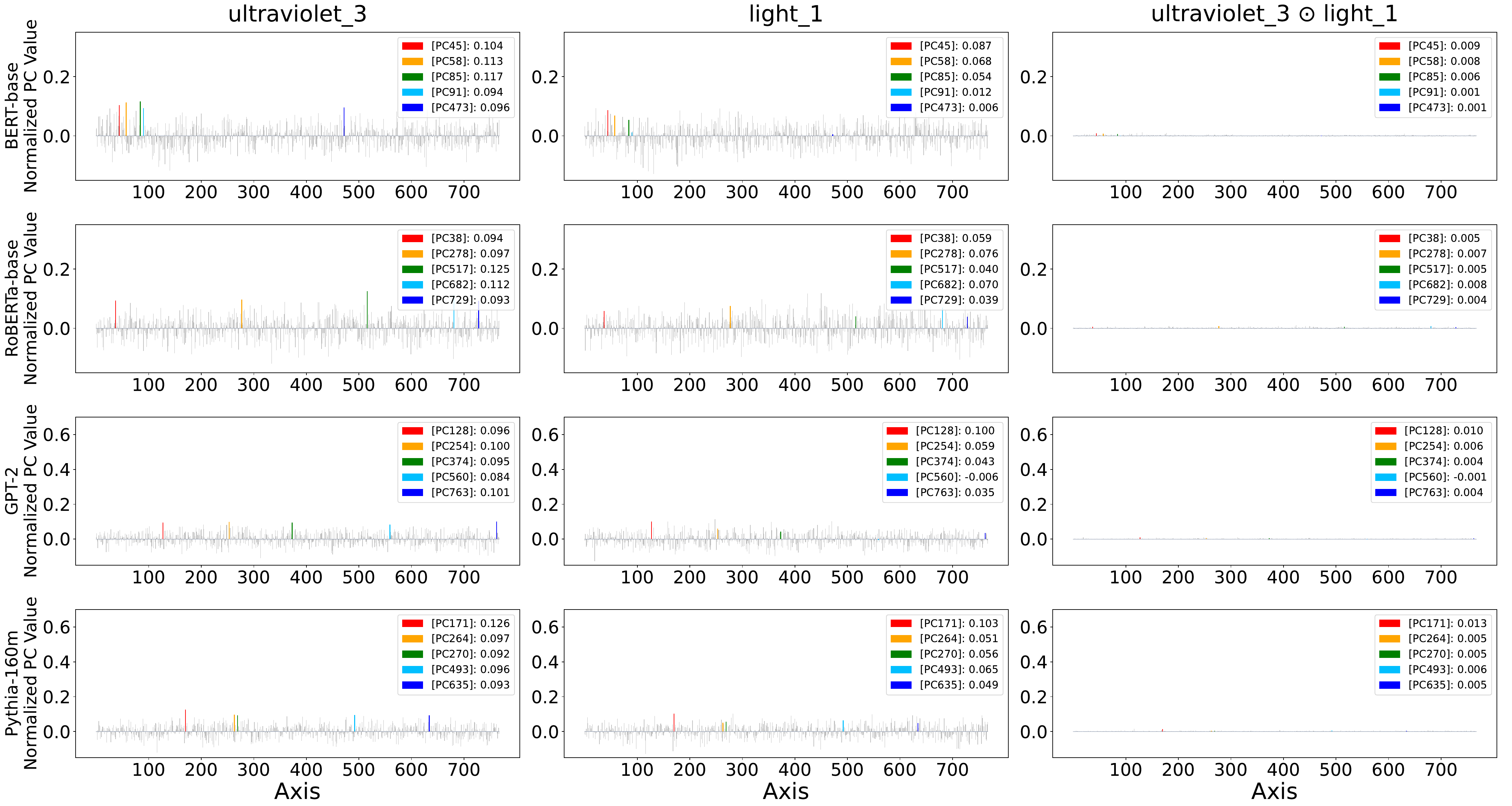}
    \caption{
Similar to Fig.~\ref{fig:cosine_pca}, the component values and their component-wise products for the normalized PCA-transformed contextualized embeddings of \textit{ultraviolet\_3} and \textit{light\_1} are shown in bar graphs. The cosine similarity is 0.532 for BERT-base, 0.608 for RoBERTa-base, 0.576 for GPT-2, and 0.699 for Pythia-160m.
}
\label{fig:cross-embedding_bargraph_pca}
\end{figure*}

\subsubsection{Interpretation of ICA Components for \textit{ultraviolet\_3}, \textit{light\_1}, and \textit{ultraviolet\_3} \texorpdfstring{$\odot$}{odot} \textit{light\_1}}
\paragraph{\textit{ultraviolet\_3}.}
Table~\ref{tab:4models_topwords} presents the top 10 words for the axes of the top 5 component values of the normalized ICA-transformed contextualized embeddings of \textit{ultraviolet\_3}, along with the semantic interpretations of these axes.

\paragraph{\textit{light\_1} and \textit{ultraviolet\_3} \texorpdfstring{$\odot$}{odot} \textit{light\_1}.}
Similar to Table~\ref{tab:4models_topwords}, for the normalized ICA-transformed embeddings of GPT-2 and Pythia-160m, Tables~\ref{tab:gpt2_ica} and~\ref{tab:pythia_ica} provide the interpretation of ICA components for
\textit{light\_1} and \textit{ultraviolet\_3} $\odot$ \textit{light\_1}.
The meaning of each axis in Tables~\ref{tab:table2_5columns_gpt2_ica} and~\ref{tab:table2_5columns_pythia} for $p$-values and their Bonferroni-corrected values is derived from Tables~\ref{tab:4models_topwords_ica},~\ref{tab:gpt2_ica} and~\ref{tab:pythia_ica}.

\subsubsection{Comparison of ICA and PCA}
We also compare ICA and PCA for contextualized embeddings.

First, as in Fig.~\ref{fig:cross-embedding_bargraph_ica} in Section~\ref{sec:various_embeddings}, Fig.~\ref{fig:cross-embedding_bargraph_ica_bert_roberta} shows bar graphs for the normalized ICA-transformed embeddings of BERT-base and RoBERTa-base.
Additionally, Fig.~\ref{fig:cross-embedding_bargraph_pca} shows bar graphs for the normalized PCA-transformed embeddings of the four models.

Then, we compare the ICA bar graphs (Figs.~\ref{fig:cross-embedding_bargraph_ica} and~\ref{fig:cross-embedding_bargraph_ica_bert_roberta}) with the PCA bar graphs (Fig.~\ref{fig:cross-embedding_bargraph_pca}).
As in Fig.~\ref{fig:cosine} in Section~\ref{sec:intro}, for the normalized ICA-transformed embeddings, the component values are sparse overall, and the component-wise products for the common semantic components (for example, \textit{[light]} and \textit{[colors]}) are significantly larger than others.
In contrast, these products are not observed in the normalized PCA-transformed embeddings.

\renewcommand{\arraystretch}{1.7} 
\begin{table*}[t!]
\tiny
\centering
\begin{tabular}{@{\hspace{0.5em}}c@{\hspace{0.5em}}|@{\hspace{0.5em}}r@{\hspace{0.5em}}|@{\hspace{0.5em}}c@{\hspace{0.15em}}c@{\hspace{0.15em}}c@{\hspace{0.15em}}c@{\hspace{0.15em}}c@{\hspace{0.5em}}c@{\hspace{0.15em}}c@{\hspace{0.15em}}c@{\hspace{0.15em}}c@{\hspace{0.15em}}c@{\hspace{0.5em}}|@{\hspace{0.5em}}c@{\hspace{0.5em}}}
\toprule
 & Axis & Top1 & Top2 & Top3 & Top4 & Top5 & Top6 & Top7 & Top8 & Top9 & Top10 & Meaning\\
\midrule
\multirow{5}{*}{\rotatebox{90}{\textit{light\_1}}} 
 & {\color{green}349} & \textit{ultraviolet\_4} & \textit{light\_1} & \textit{-\_0} & \textit{light\_0} & \textit{accurate\_1} & \textit{-\_1} & \textit{ultraviolet\_1} & \textit{color\_4} & \textit{ultraviolet\_3} & \textit{ultraviolet\_0} & {\color{green}\textit{[light]}} \\
 & {\color{orange}273} & \textit{arsenic\_0} & \textit{nitrogen\_1} & \textit{phosphorus\_0} & \textit{nitrogen\_0} & \textit{ate\_4} & \textit{dioxide\_0} & \textit{calcium\_0} & \textit{carbon\_9} & \textit{opes\_0} & \textit{fumes\_0} & {\color{orange}\textit{[elements]}} \\
 & {\color{blue}619} & \textit{restaurant\_0} & \textit{proposal\_2} & \textit{project\_3} & \textit{project\_5} & \textit{soldier\_2} & \textit{farmer\_0} & \textit{scheme\_6} & \textit{designer\_1} & \textit{company\_7} & \textit{company\_6} & {\color{blue}\textit{[occupations]}} \\
 & 402 & \textit{rain\_8} & \textit{rain\_9} & \textit{rain\_2} & \textit{rain\_1} & \textit{rain\_3} & \textit{storm\_1} & \textit{rain\_7} & \textit{rain\_6} & \textit{thunder\_0} & \textit{storm\_3} & \textit{[rainstorm]} \\
 & 284 & \textit{dangers\_2} & \textit{risks\_2} & \textit{dangers\_0} & \textit{dangers\_3} & \textit{capabilities\_1} & \textit{iencies\_1} & \textit{behavior\_1} & \textit{difficulties\_1} & \textit{danger\_2} & \textit{inequalities\_0} & \textit{[challenges]} \\
\midrule
\multirow{5}{*}{\rotatebox{90}{\textit{ultraviolet\_3} $\odot$ \textit{light\_1}}} 
 & {\color{green}349} & \textit{ultraviolet\_4} & \textit{light\_1} & \textit{-\_0} & \textit{light\_0} & \textit{accurate\_1} & \textit{-\_1} & \textit{ultraviolet\_1} & \textit{color\_4} & \textit{ultraviolet\_3} & \textit{ultraviolet\_0} & {\color{green}\textit{[light]}} \\
 & {\color{orange}273} & \textit{arsenic\_0} & \textit{nitrogen\_1} & \textit{phosphorus\_0} & \textit{nitrogen\_0} & \textit{ate\_4} & \textit{dioxide\_0} & \textit{calcium\_0} & \textit{carbon\_9} & \textit{opes\_0} & \textit{fumes\_0} & {\color{orange}\textit{[elements]}} \\
 & {\color{red}28} & \textit{to\_0} & \textit{attracts\_0} & \textit{in\_0} & \textit{specialized\_0} & \textit{genes\_0} & \textit{ting\_0} & \textit{cause\_5} & \textit{adopt\_0} & \textit{ocytes\_0} & \textit{produced\_1} & {\color{red}\textit{[biology]}} \\
 & {\color{blue}619} & \textit{restaurant\_0} & \textit{proposal\_2} & \textit{project\_3} & \textit{project\_5} & \textit{soldier\_2} & \textit{farmer\_0} & \textit{scheme\_6} & \textit{designer\_1} & \textit{company\_7} & \textit{company\_6} & {\color{blue}\textit{[occupations]}} \\
 & 402 & \textit{rain\_8} & \textit{rain\_9} & \textit{rain\_2} & \textit{rain\_1} & \textit{rain\_3} & \textit{storm\_1} & \textit{rain\_7} & \textit{rain\_6} & \textit{thunder\_0} & \textit{storm\_3} & \textit{[rainstorm]} \\
\bottomrule
\end{tabular}
\caption{
For the normalized ICA-transformed GPT-2 embeddings of \textit{light\_1} and the component-wise products \textit{ultraviolet\_3 $\odot$ light\_1} in Table~\ref{tab:table2_5columns_gpt2_ica}, the axes of the top 5 values are focused on, and their top 10 words are shown.
Axes are sorted by component values.
The meanings of the axes are interpreted from these listed words.
}
\label{tab:gpt2_ica}
\end{table*}
\renewcommand{\arraystretch}{1.0}

\renewcommand{\arraystretch}{1.7} 
\begin{table*}[t!]
\tiny
\centering
\begin{tabular}{@{\hspace{0.5em}}c@{\hspace{0.5em}}|@{\hspace{0.5em}}r@{\hspace{0.5em}}|@{\hspace{0.5em}}c@{\hspace{0.15em}}c@{\hspace{0.15em}}c@{\hspace{0.15em}}c@{\hspace{0.15em}}c@{\hspace{0.5em}}c@{\hspace{0.15em}}c@{\hspace{0.15em}}c@{\hspace{0.15em}}c@{\hspace{0.15em}}c@{\hspace{0.5em}}|@{\hspace{0.5em}}c@{\hspace{0.5em}}}
\toprule
 & Axis & Top1 & Top2 & Top3 & Top4 & Top5 & Top6 & Top7 & Top8 & Top9 & Top10 & Meaning\\
\midrule
\multirow{5}{*}{\rotatebox{90}{\textit{light\_1}}} 
 & {\color{red}262} & \textit{light\_1} & \textit{ultraviolet\_3} & \textit{light\_0} & \textit{ultraviolet\_4} & \textit{ultraviolet\_1} & \textit{-\_0} & \textit{ultraviolet\_0} & \textit{-\_1} & \textit{ultraviolet\_2} & \textit{light\_0} & {\color{red}\textit{[light]}} \\
 & {\color{cyan}619} & \textit{scheme\_6} & \textit{restaurant\_0} & \textit{campaign\_3} & \textit{discovery\_2} & \textit{investor\_1} & \textit{resolution\_2} & \textit{island\_0} & \textit{let\_5} & \textit{designer\_1} & \textit{shoes\_4} & {\color{cyan}\textit{[business]}} \\
 & 264 & \textit{carbon\_7} & \textit{carbon\_6} & \textit{carbon\_8} & \textit{carbon\_0} & \textit{carbon\_4} & \textit{carbon\_5} & \textit{nitrogen\_1} & \textit{nitrogen\_0} & \textit{phosphorus\_0} & \textit{arsenic\_0} & \textit{[elements]} \\
 & {\color{green}485} & \textit{for\_2} & \textit{and\_3} & \textit{both\_0} & \textit{and\_2} & \textit{and\_1} & \textit{,\_4} & \textit{of\_2} & \textit{,\_5} & \textit{limited\_0} & \textit{)\_0} & {\color{green}\textit{[function words]}} \\
 & 548 & \textit{fre\_1} & \textit{fre\_4} & \textit{fre\_0} & \textit{fre\_2} & \textit{fre\_3} & \textit{eas\_0} & \textit{gre\_0} & \textit{gre\_1} & \textit{eas\_3} & \textit{eas\_5} & \textit{[subwords]} \\
\midrule
\multirow{5}{*}{\rotatebox{90}{\textit{ultraviolet\_3} $\odot$ \textit{light\_1}}} 
 & {\color{red}262} & \textit{light\_1} & \textit{ultraviolet\_3} & \textit{light\_0} & \textit{ultraviolet\_4} & \textit{ultraviolet\_1} & \textit{-\_0} & \textit{ultraviolet\_0} & \textit{-\_1} & \textit{ultraviolet\_2} & \textit{light\_0} & {\color{red}\textit{[light]}} \\
 & {\color{cyan}619} & \textit{scheme\_6} & \textit{restaurant\_0} & \textit{campaign\_3} & \textit{discovery\_2} & \textit{investor\_1} & \textit{resolution\_2} & \textit{island\_0} & \textit{let\_5} & \textit{designer\_1} & \textit{shoes\_4} & {\color{cyan}\textit{[business]}} \\
 & {\color{green}485} & \textit{for\_2} & \textit{and\_3} & \textit{both\_0} & \textit{and\_2} & \textit{and\_1} & \textit{,\_4} & \textit{of\_2} & \textit{,\_5} & \textit{limited\_0} & \textit{)\_0} & {\color{green}\textit{[function words]}} \\
 & 264 & \textit{carbon\_7} & \textit{carbon\_6} & \textit{carbon\_8} & \textit{carbon\_0} & \textit{carbon\_4} & \textit{carbon\_5} & \textit{nitrogen\_1} & \textit{nitrogen\_0} & \textit{phosphorus\_0} & \textit{arsenic\_0} & \textit{[elements]} \\
 & 336 & \textit{made\_3} & \textit{significant\_9} & \textit{ful\_7} & \textit{making\_5} & \textit{make\_3} & \textit{specific\_4} & \textit{making\_1} & \textit{making\_2} & \textit{grave\_2} & \textit{makes\_0} & \textit{[creation]} \\
\bottomrule
\end{tabular}
\caption{
For the normalized ICA-transformed Pythia-160m embeddings of \textit{light\_1} and the component-wise products \textit{ultraviolet\_3 $\odot$ light\_1} in Table~\ref{tab:table2_5columns_pythia}, the axes of the top 5 values are focused on, and their top 10 words are shown.
Axes are sorted by component values.
The meanings of the axes are interpreted from these listed words.
}
\label{tab:pythia_ica}
\end{table*}
\renewcommand{\arraystretch}{1.0}

\subsection{Comparison of normalization of embeddings in Appendix~\ref{app:normalization}}\label{app:comp_cross_embeds_normalization}
Based on the normalization experiments described in Appendix~\ref{app:normalization}, we also conducted comparative experiments on normalization using the four contextualized embeddings after ICA and PCA transformations.

\begin{figure*}[t!]
\centering
\begin{subfigure}{\textwidth}
\centering
    \includegraphics[width=\textwidth]{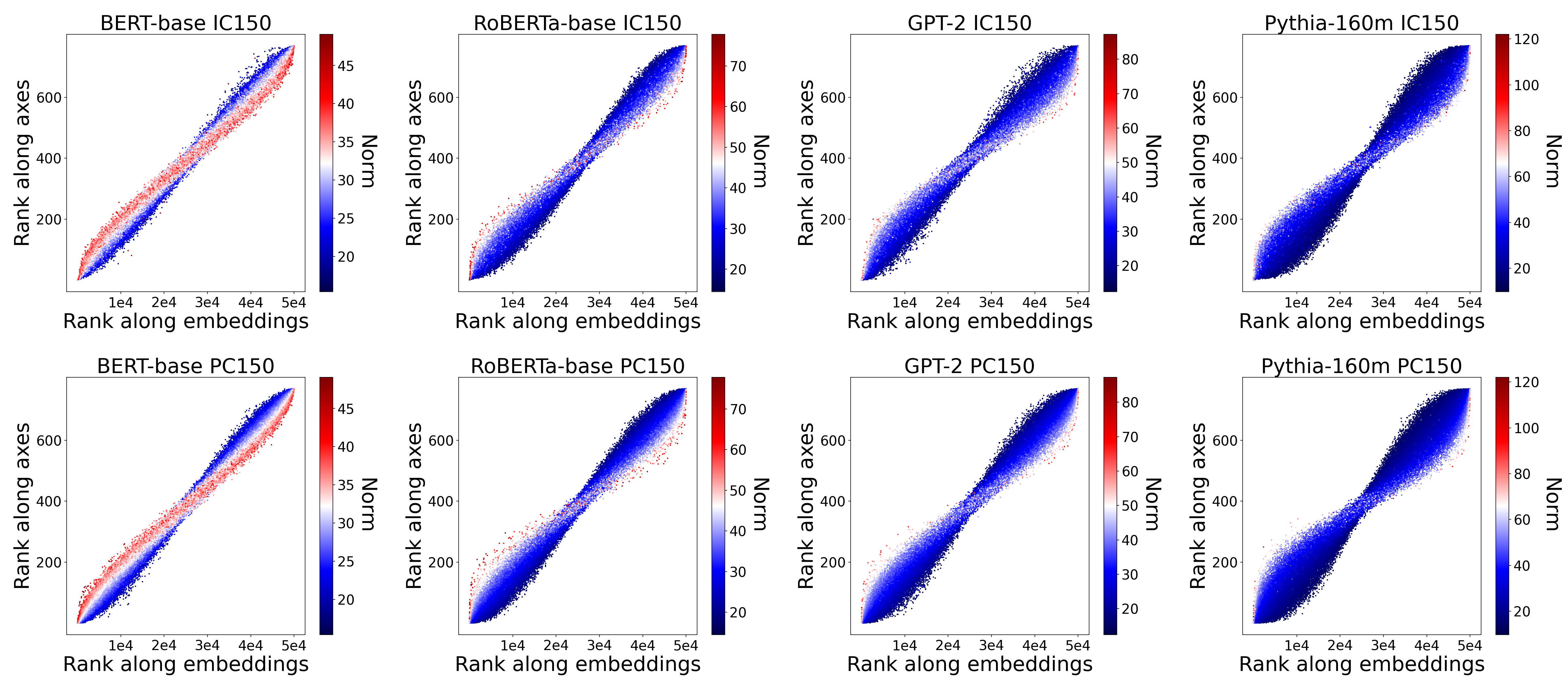}
    \subcaption{Before Normalization}
    \label{fig:inner_rank_before_149_4models}
\end{subfigure}
\par\bigskip
\begin{subfigure}{\textwidth}
\centering
    \includegraphics[width=\textwidth]{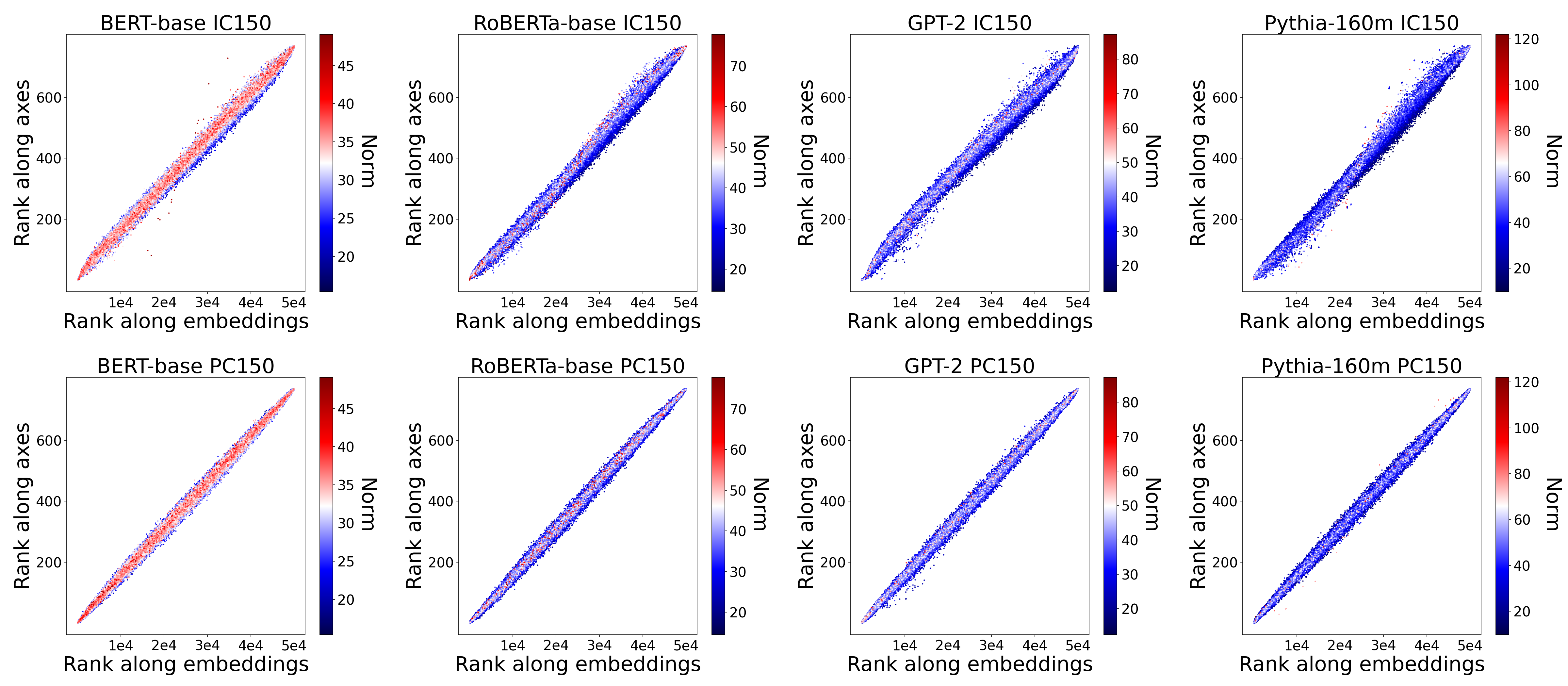}
    \subcaption{After Normalization}
    \label{fig:inner_rank_after_149_4models}
\end{subfigure}
    \caption{
Scatterplots for the 150th axis of the (top) ICA-transformed and (bottom) PCA-transformed contextualized embeddings, (a) before and (b) after normalization, showing the rankings of component values along the embeddings and those along the axes, colored by the norms.
The larger the norm of an embedding, the more it is plotted in the foreground.
    }
\label{fig:inner_rank_149_4models}
\end{figure*}

Focusing on the component values of the 150th axis of each embedding, Fig.~\ref{fig:inner_rank_149_4models} shows scatterplots of the ranks of component values along the embeddings and those along the axes, both before and after normalization.
Similar to the norm-derived artifacts observed in Fig.~\ref{fig:inner_rank_149} in Appendix~\ref{app:normalization}, artifacts are seen in Fig.~\ref{fig:inner_rank_before_149_4models} before normalization.
However, these artifacts disappear in Fig.~\ref{fig:inner_rank_after_149_4models} after normalization.

\subsection{Comparison of component values of ICA and PCA in Section~\ref{sec:ica-vs-pca}}\label{app:ica-vs-pca}
We also compare the ICA-transformed and PCA-transformed contextualized embeddings after normalization, based on Section~\ref{sec:ica-vs-pca}. 
These embeddings have a size of $50{,}000$ and a dimensionality of $768$.

\begin{figure*}[t!]
\centering
\begin{subfigure}{\textwidth}
\centering
    \includegraphics[width=\textwidth]{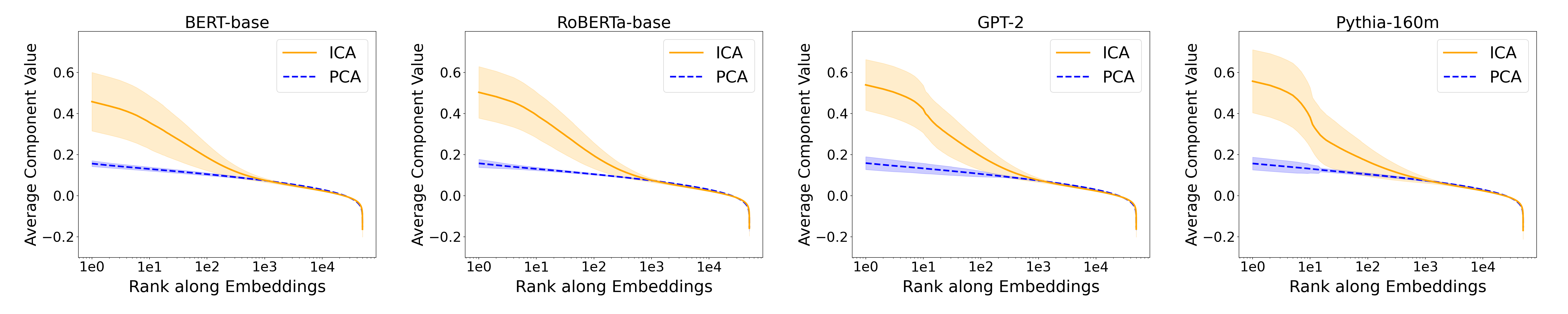}
    \subcaption{Sorted along embeddings}
    \label{fig:sorted_along_embeddings_4models}
\end{subfigure}
\par\bigskip
\begin{subfigure}{\textwidth}
\centering
    \includegraphics[width=\textwidth]{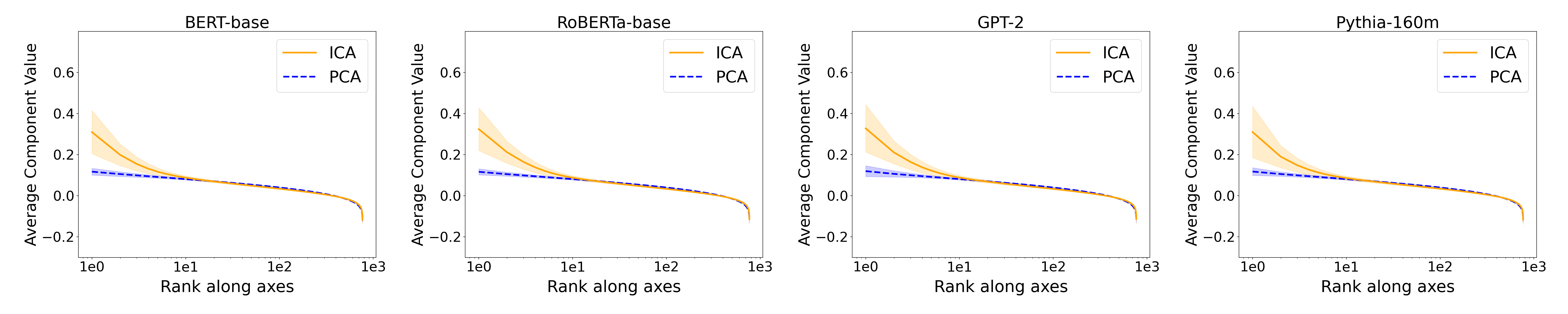}
    \subcaption{Sorted along axes}
    \label{fig:sorted_along_axes_4models}
\end{subfigure}
    \caption{
Comparison of component values of normalized ICA-transformed and PCA-transformed contextualized embeddings. 
The component values are averaged after being sorted in descending order (a) along embeddings for each axis, and  (b) along axes for each embedding. The range of $\pm1\sigma$ is shown, where $\sigma$ is the standard deviation of the component values.
    }
\label{fig:sorted_along_4models}
\end{figure*}

Figure~\ref{fig:sorted_along_4models} shows the sorted component values of these embeddings. 
In Fig.~\ref{fig:sorted_along_embeddings_4models}, the ICA components are larger than the PCA components for up to approximately $1{,}000$ embeddings. 
Additionally, in Fig.~\ref{fig:sorted_along_axes_4models}, the ICA components are larger than the PCA components for up to approximately the 10th axis. 
These results show a trend similar to that shown by GloVe in Figs.~\ref{fig:sort_emb_component} and~\ref{fig:sort_axis_component} in Section~\ref{sec:ica-vs-pca}.

\section{Details of Word Intrusion Task in Section~\ref{sec:intruder}} \label{app:intruder}

\paragraph{Selection of the intruder word.} 
Our objective is to evaluate the interpretability of the word embeddings $\mathbf{Y}\in \mathbb{R}^{n\times d}$, where each row vector $\mathbf{y}_i\in \mathbb{R}^d$ corresponds to a word $w_i$.
In order to select the intruder word for the set of top $k\,(=5)$ words of each $\ell$-th axis ($\ell\in\{1,\ldots,d\}$), denoted as $\mathrm{top}_{k}(\ell)$, we randomly chose a word from a pool of words that satisfy both of the following two criteria simultaneously: (i) the word ranks in the lower $50\%$ in terms of the component value on the $\ell$-th axis, and (ii) it ranks in the top $10\%$ in terms of the component value on some axis other than the $\ell$-th axis.
For each axis, $L\,(=10)$ intruder words are randomly selected, and $W_{\mathrm{int}}(\ell)$ denotes the set of these $L$ intruder words.

\paragraph{Evaluation metric.}
We adopted the metric by \citet{DBLP:conf/ijcai/SunGLXC16}, but we mitigated the effect of outliers by using the median instead of the mean in the final score computation described below:
\begin{flalign*}
\mathrm{IntruderScore} = 
\underset{\ell=1,\ldots,d}{\mathrm{median}} \,  \frac{\mathrm{InterDist}(\ell)}{\mathrm{IntraDist}(\ell)},  &&
\end{flalign*}
\begin{flalign*}
    \mathrm{IntraDist}(\ell) &= \sum_{\substack{w_i, w_j \in \mathrm{top}_{k}(\ell) \\ w_i \neq w_j}} \frac{\mathrm{dist}(w_i, w_j)}{k(k-1)}, &&\\
    \mathrm{InterDist}(\ell) &= \underset{w \in W_{\mathrm{int}}(\ell)}{\mathrm{mean}} \sum_{w_i \in \mathrm{top}_{k}(\ell)} \frac{\mathrm{dist}(w_i, w)}{k}. &&
\end{flalign*}
In this formula, we defined $\mathrm{dist}(w_i, w_j) = \| \mathbf{y}_i - \mathbf{y}_j\|$.
Here, $\mathrm{IntraDist}(\ell)$ denotes the average distance between the top $k$ words, and $\mathrm{InterDist}(\ell)$ represents the average distance between the top words and the intruder word. The score is higher when the intruder word is further away from the set $\mathrm{top}_{k}(\ell)$. Therefore, this score serves as a quantitative measure of the ability to identify the intruder word; thus, it is used as a measure of the consistency of the meaning of the top $k$ words and the interpretability of axes.

\section{Details of the Downstream Tasks in Section~\ref{sec:downstream}}\label{app:downstream}
We performed analogy and word similarity tasks following the settings of~\citet{DBLP:journals/corr/abs-2401-06112}, using the Word Embedding Benchmark~\cite{DBLP:journals/corr/JastrzebskiLC17}\footnote{\url{https://github.com/kudkudak/word-embeddings-benchmarks}}.
We explain the details of the tasks using the ICA-transformed embeddings, and apply the same procedure to the PCA-transformed embeddings. 
We define $\llbracket d \rrbracket:=\{1,\ldots,d\}$ and let $p\,(\leq d)$ be the number of non-zero product components.

\subsection{Word similarity tasks}\label{app:wordsim}

\subsubsection{Settings}
We used several datasets, including MEN~\cite{ws-MEN}, MTurk~\cite{ws-MTurk}, RG65~\cite{ws-RG65}, RW~\cite{ws-RW}, SimLex999~\cite{ws-SimLex999}, WS353~\cite{ws-WS353}, WS353R (WS353 Relatedness), and WS353S (WS353 Similarity).
In these tasks, we compute the cosine similarity $\cos{(w_i, w_j)}$ between words $w_i$ and $w_j$ and compare it with human-rated similarity scores. 
Spearman's rank correlation is used as the evaluation metric.

\subsubsection{Details of \texorpdfstring{$p$}{p} non-zero products}
We explain the top $p$ component-wise products used for the similarity tasks. 

As seen in (\ref{eq:cosine}), since the cosine similarity between words $w_i$ and $w_j$ is expressed as the sum of the component-wise products, we consider the index set of the top $p$ component-wise products:
\begin{align}
\text{Top}_p:=\mathop{\argmax}_{\ell\in \llbracket d \rrbracket}{}_p\,\hat{s}_i^{(\ell)}\,\hat{s}_j^{(\ell)}\subset\llbracket d \rrbracket.\label{eq:Top-p}
\end{align}
If $p=d$, then $\text{Top}_p=\llbracket d \rrbracket$.

Then, based on the cosine similarity expression in (\ref{eq:cosine}), we express the top $p$ component-wise products as
\begin{align}
\sum_{\ell\in\text{Top}_p} \hat{s}_{i}^{(\ell)}\,\hat{s}_j^{(\ell)}.\label{eq:cosine-p}
\end{align}

\begin{table*}[!t]
\centering
\begin{adjustbox}{width=\linewidth}
\begin{tabular}{clrrrrrrrrrrrr}
\toprule
& & \multicolumn{2}{c}{$p=1$} & \multicolumn{2}{c}{$p=5$} & \multicolumn{2}{c}{$p=10$} & \multicolumn{2}{c}{$p=50$} & \multicolumn{2}{c}{$p=100$} & \multicolumn{2}{c}{$p=300$} \\
\cmidrule(rr){3-4}\cmidrule(rr){5-6}\cmidrule(rr){7-8}\cmidrule(rr){9-10}\cmidrule(rr){11-12}\cmidrule(rr){13-14}
& Tasks & PCA & ICA & PCA & ICA & PCA & ICA & PCA & ICA & PCA & ICA & PCA & ICA\\
\midrule
\multirow{9}{*}{Similarity} & MEN & 0.11 & 0.45 & 0.31 & 0.59 & 0.45 & 0.63 & 0.69 & 0.72 & 0.73 & 0.74 & 0.75 & 0.75\\
 & WS353 & -0.02 & 0.19 & 0.12 & 0.45 & 0.25 & 0.49 & 0.49 & 0.55 & 0.54 & 0.56 & 0.57 & 0.57\\
 & WS353R & 0.04 & 0.16 & 0.15 & 0.44 & 0.22 & 0.46 & 0.43 & 0.49 & 0.48 & 0.49 & 0.51 & 0.51\\
 & WS353S & 0.01 & 0.31 & 0.17 & 0.56 & 0.34 & 0.60 & 0.62 & 0.67 & 0.66 & 0.70 & 0.69 & 0.69\\
 & SimLex999 & 0.00 & 0.10 & 0.11 & 0.21 & 0.19 & 0.25 & 0.35 & 0.35 & 0.38 & 0.37 & 0.40 & 0.40\\
 & RW & 0.08 & 0.13 & 0.15 & 0.20 & 0.18 & 0.22 & 0.27 & 0.30 & 0.30 & 0.32 & 0.34 & 0.34\\
 & RG65 & 0.42 & 0.42 & 0.53 & 0.57 & 0.61 & 0.64 & 0.71 & 0.76 & 0.74 & 0.79 & 0.78 & 0.78\\
 & MTurk & 0.22 & 0.42 & 0.36 & 0.58 & 0.49 & 0.61 & 0.64 & 0.65 & 0.66 & 0.66 & 0.64 & 0.64\\
\cmidrule(lr){2-14}
 & Average & 0.11 & \textbf{0.27} & 0.24 & \textbf{0.45} & 0.34 & \textbf{0.49} & 0.53 & \textbf{0.56} & 0.56 & \textbf{0.58} & 0.59 & 0.59\\
\midrule
\multirow{31}{*}{Analogy} 
 & capital-common-countries & 0.00 & 0.22 & 0.01 & 0.51 & 0.24 & 0.62 & 0.93 & 0.94 & 0.95 & 0.94 & 0.95 & 0.95\\
 & capital-world & 0.01 & 0.05 & 0.04 & 0.15 & 0.21 & 0.29 & 0.91 & 0.90 & 0.95 & 0.93 & 0.95 & 0.95\\
 & city-in-state & 0.00 & 0.00 & 0.00 & 0.13 & 0.05 & 0.21 & 0.49 & 0.55 & 0.59 & 0.60 & 0.67 & 0.67\\
 & currency & 0.00 & 0.00 & 0.00 & 0.00 & 0.00 & 0.03 & 0.06 & 0.11 & 0.11 & 0.12 & 0.12 & 0.12\\
 & family & 0.00 & 0.00 & 0.01 & 0.24 & 0.14 & 0.36 & 0.74 & 0.74 & 0.84 & 0.84 & 0.88 & 0.88\\
 & gram1-adjective-to-adverb & 0.00 & 0.00 & 0.00 & 0.00 & 0.02 & 0.02 & 0.16 & 0.16 & 0.19 & 0.20 & 0.21 & 0.21\\
 & gram2-opposite & 0.00 & 0.00 & 0.00 & 0.00 & 0.02 & 0.02 & 0.18 & 0.18 & 0.24 & 0.23 & 0.26 & 0.26\\
 & gram3-comparative & 0.00 & 0.00 & 0.02 & 0.19 & 0.17 & 0.42 & 0.79 & 0.82 & 0.84 & 0.86 & 0.88 & 0.88\\
 & gram4-superlative & 0.00 & 0.00 & 0.04 & 0.15 & 0.14 & 0.28 & 0.57 & 0.63 & 0.66 & 0.69 & 0.69 & 0.69\\
 & gram5-present-participle & 0.00 & 0.00 & 0.01 & 0.15 & 0.07 & 0.36 & 0.64 & 0.68 & 0.69 & 0.70 & 0.69 & 0.69\\
 & gram6-nationality-adjective & 0.01 & 0.22 & 0.14 & 0.37 & 0.52 & 0.49 & 0.91 & 0.92 & 0.92 & 0.92 & 0.93 & 0.93\\
 & gram7-past-tense & 0.00 & 0.00 & 0.03 & 0.09 & 0.09 & 0.21 & 0.50 & 0.52 & 0.56 & 0.57 & 0.60 & 0.60\\
 & gram8-plural & 0.00 & 0.00 & 0.01 & 0.15 & 0.13 & 0.27 & 0.70 & 0.69 & 0.75 & 0.74 & 0.76 & 0.76\\
 & gram9-plural-verbs & 0.00 & 0.00 & 0.02 & 0.20 & 0.12 & 0.37 & 0.48 & 0.57 & 0.57 & 0.60 & 0.58 & 0.58\\
& jj\_jjr & 0.00 & 0.00 & 0.03 & 0.13 & 0.14 & 0.29 & 0.53 & 0.56 & 0.60 & 0.63 & 0.66 & 0.66\\
 & jj\_jjs & 0.00 & 0.00 & 0.03 & 0.11 & 0.10 & 0.22 & 0.41 & 0.49 & 0.49 & 0.54 & 0.51 & 0.51\\
 & jjr\_jj & 0.00 & 0.00 & 0.01 & 0.03 & 0.05 & 0.10 & 0.45 & 0.46 & 0.53 & 0.54 & 0.54 & 0.54\\
 & jjr\_jjs & 0.00 & 0.00 & 0.03 & 0.08 & 0.09 & 0.20 & 0.38 & 0.49 & 0.49 & 0.56 & 0.55 & 0.55\\
 & jjs\_jj & 0.00 & 0.00 & 0.02 & 0.05 & 0.05 & 0.10 & 0.36 & 0.36 & 0.43 & 0.44 & 0.48 & 0.48\\
 & jjs\_jjr & 0.00 & 0.01 & 0.01 & 0.14 & 0.09 & 0.29 & 0.55 & 0.59 & 0.60 & 0.63 & 0.63 & 0.63\\
 & nn\_nnpos & 0.00 & 0.02 & 0.00 & 0.13 & 0.06 & 0.22 & 0.35 & 0.35 & 0.39 & 0.39 & 0.42 & 0.42\\
 & nn\_nns & 0.00 & 0.03 & 0.00 & 0.24 & 0.10 & 0.39 & 0.62 & 0.63 & 0.69 & 0.68 & 0.74 & 0.74\\
 & nnpos\_nn & 0.00 & 0.06 & 0.00 & 0.21 & 0.05 & 0.29 & 0.41 & 0.41 & 0.43 & 0.42 & 0.45 & 0.45\\
 & nns\_nn & 0.00 & 0.06 & 0.01 & 0.24 & 0.11 & 0.36 & 0.56 & 0.55 & 0.61 & 0.59 & 0.64 & 0.64\\
 & vb\_vbd & 0.00 & 0.01 & 0.03 & 0.10 & 0.13 & 0.23 & 0.55 & 0.53 & 0.56 & 0.54 & 0.58 & 0.58\\
 & vb\_vbz & 0.00 & 0.00 & 0.02 & 0.16 & 0.11 & 0.34 & 0.66 & 0.72 & 0.74 & 0.76 & 0.76 & 0.76\\
 & vbd\_vb & 0.00 & 0.01 & 0.02 & 0.09 & 0.08 & 0.27 & 0.60 & 0.64 & 0.66 & 0.67 & 0.69 & 0.69\\
 & vbd\_vbz & 0.00 & 0.00 & 0.03 & 0.11 & 0.10 & 0.29 & 0.56 & 0.60 & 0.63 & 0.63 & 0.63 & 0.63\\
 & vbz\_vb & 0.00 & 0.00 & 0.02 & 0.09 & 0.15 & 0.27 & 0.77 & 0.75 & 0.80 & 0.79 & 0.82 & 0.82\\
 & vbz\_vbd & 0.01 & 0.00 & 0.04 & 0.05 & 0.14 & 0.14 & 0.57 & 0.51 & 0.59 & 0.56 & 0.55 & 0.55\\
\cmidrule(lr){2-14}
 & Average & 0.00 & \textbf{0.02} & 0.02 & \textbf{0.14} & 0.12 & \textbf{0.26} & 0.55 & \textbf{0.57} & 0.60 & \textbf{0.61} & 0.63 & 0.63\\
\bottomrule
\end{tabular}
\end{adjustbox}
\caption{
The performance of ICA-transformed and PCA-transformed embeddings when reducing the number of non-zero normalized component-wise products to $p$ and then computing cosine similarity. The values in the table correspond to Top 1 accuracy for analogy tasks and Spearman's rank correlation for word similarity tasks.
}
\label{tab:downstream}
\end{table*}

\subsection{Analogy tasks}
\subsubsection{Settings}\label{app:analogy-setting}
We used the Google Analogy Test Set~\cite{analogy-google}, which consists of 14 types of analogy tasks, and the Microsoft Research Syntactic Analogies Dataset~\cite{analogy-msr}, which consists of 16 types. 
In these tasks, if $w_{i_1}$ corresponds to $w_{i_2}$, then we predict $w_{i_4}$ to which $w_{i_3}$ corresponds.
To do this, the vector $\mathbf{s}_{i_2}-\mathbf{s}_{i_1}+\mathbf{s}_{i_3}$ is computed, and if the index of the closest word embedding to the vector in terms of cosine similarity,
\begin{align}
\mathop{\argmax}_{i\in \llbracket n \rrbracket}{}\,\cos{(\mathbf{s}_{i_2}-\mathbf{s}_{i_1}+\mathbf{s}_{i_3}, \mathbf{s}_i)},\label{eq:analogy}
\end{align}
is $i_4$, then it is considered correct (top 1 accuracy).

\begin{figure*}[!t]
    \centering
    \begin{minipage}{0.33\linewidth}
        \centering
        \includegraphics[width=\linewidth]{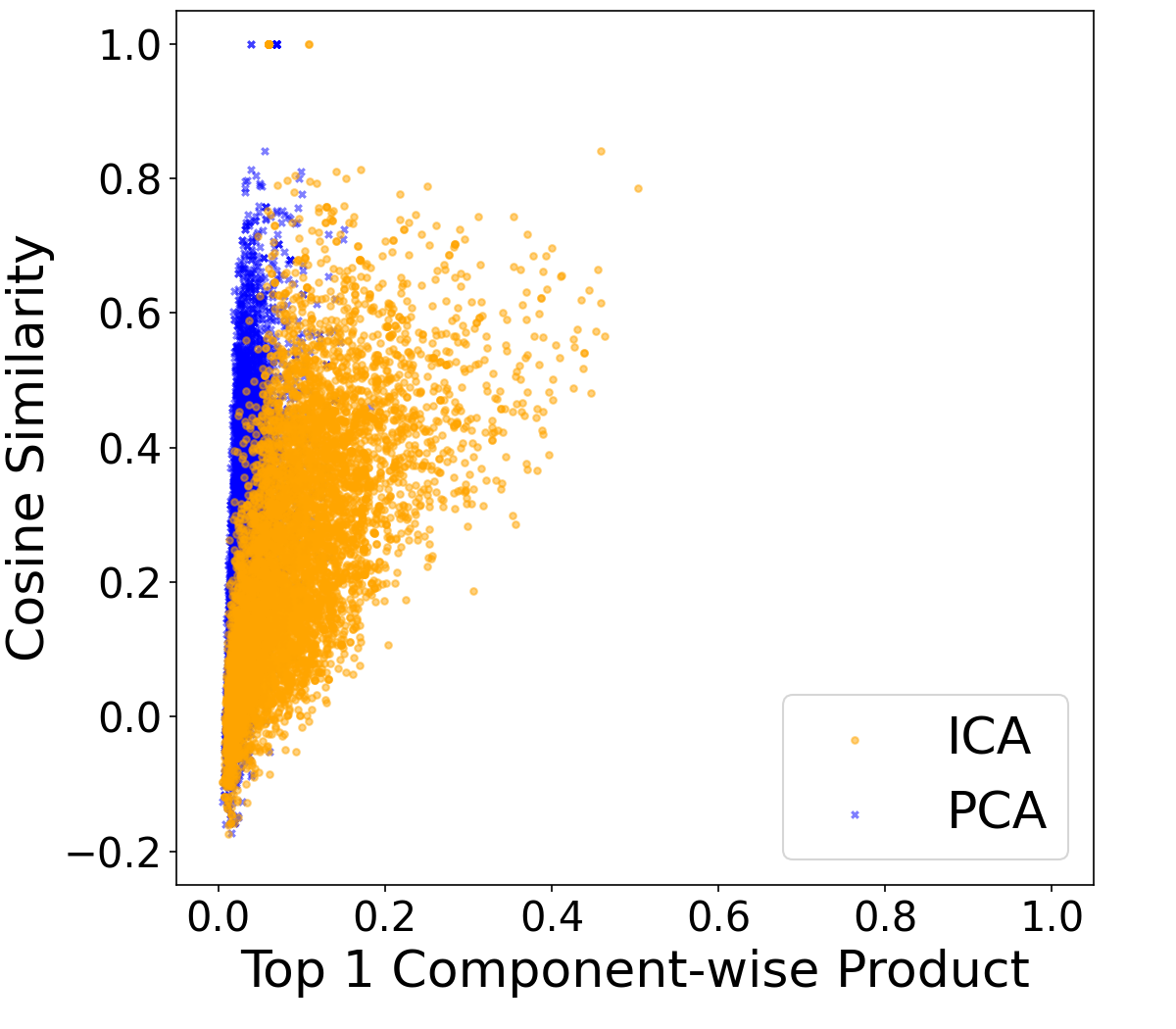}
        \subcaption{$p=1$}
        \label{fig:scatter_p1}
    \end{minipage}\hfill
    \begin{minipage}{0.33\linewidth}
        \centering
        \includegraphics[width=\linewidth]{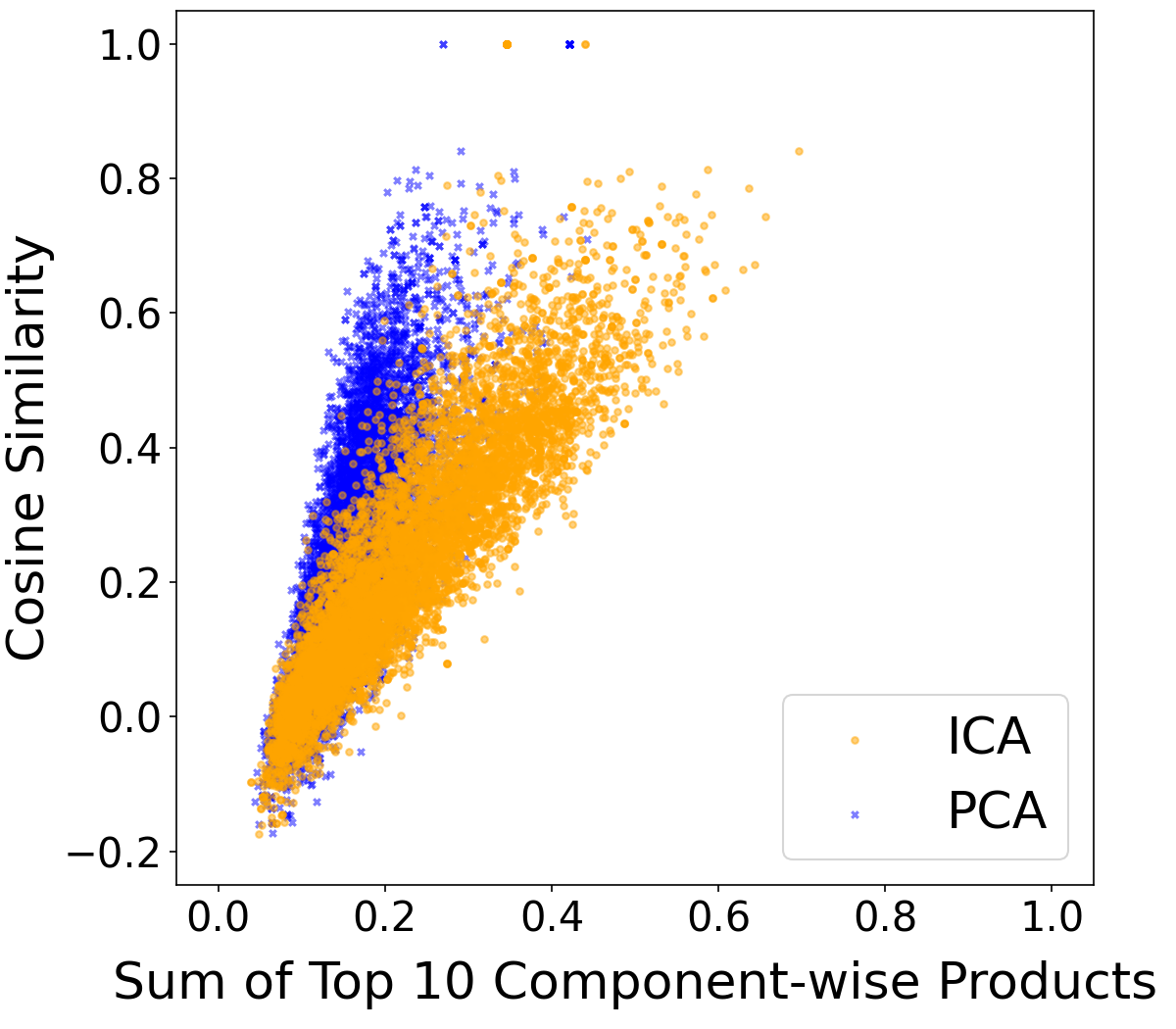}
        \subcaption{$p=10$}
        \label{fig:scatter_p10}
    \end{minipage}
    \begin{minipage}{0.33\linewidth}
        \centering
        \includegraphics[width=\linewidth]{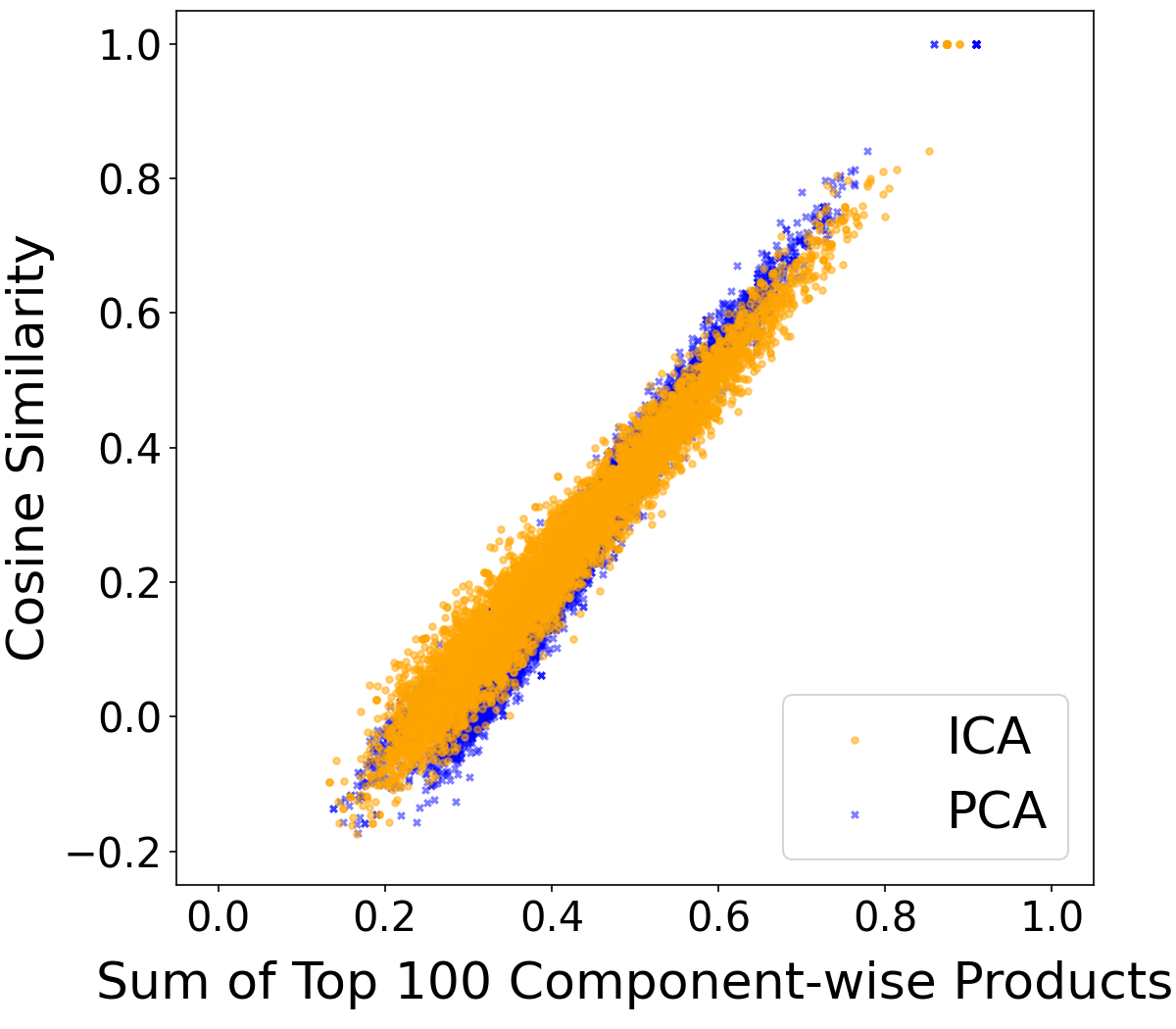}
        \subcaption{$p=100$}
        \label{fig:scatter_p100}
    \end{minipage}
    \caption{
Comparison of the scatterplots for the sum of the top $p$ component-wise products and the cosine similarity for the normalized ICA-transformed and PCA-transformed embeddings at (a) $p=1$, (b) $10$, and (c) $100$. 
We used word pairs from the word similarity tasks. 
For $p=1$, $10$, and $100$, the correlation coefficients for ICA were $0.619$, $0.876$, and $0.979$, respectively, while for PCA they were $0.432$, $0.761$, and $0.982$. 
See Fig.~\ref{fig:cos_toppsum} for the correlation coefficients for other values of $p$.
}
    \label{fig:scatter_p}
\end{figure*}

\begin{figure}[t!]
    \centering
    \includegraphics[width=0.9\columnwidth]{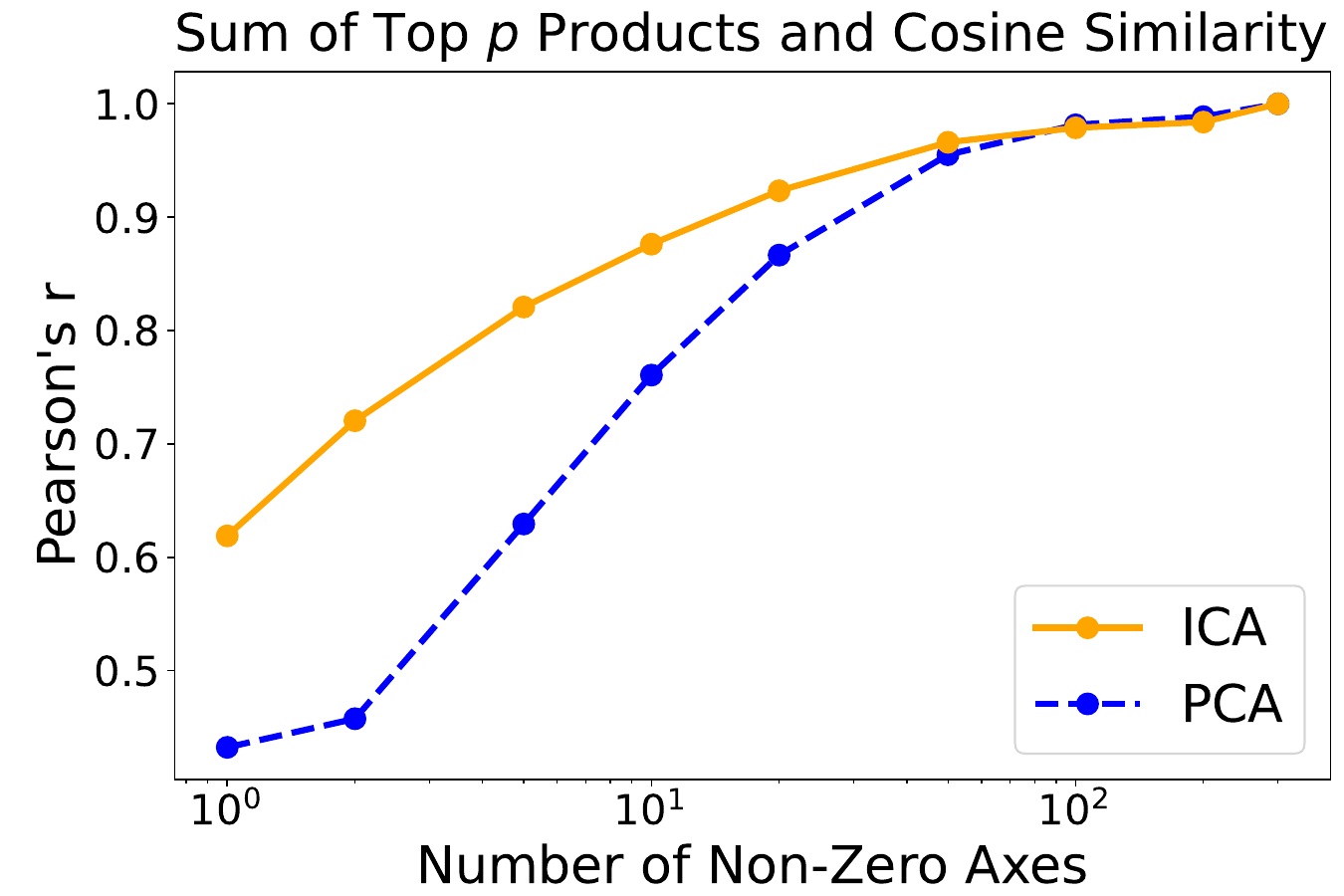}
\caption{
Comparison of correlation coefficients between the sum of the top $p$ component-wise products and the cosine similarity for normalized ICA-transformed and PCA-transformed embeddings.
We used word pairs from the word similarity tasks.
}
    \label{fig:cos_toppsum}
\end{figure}

\begin{figure*}[t!]
    \centering
    \includegraphics[width=\textwidth]{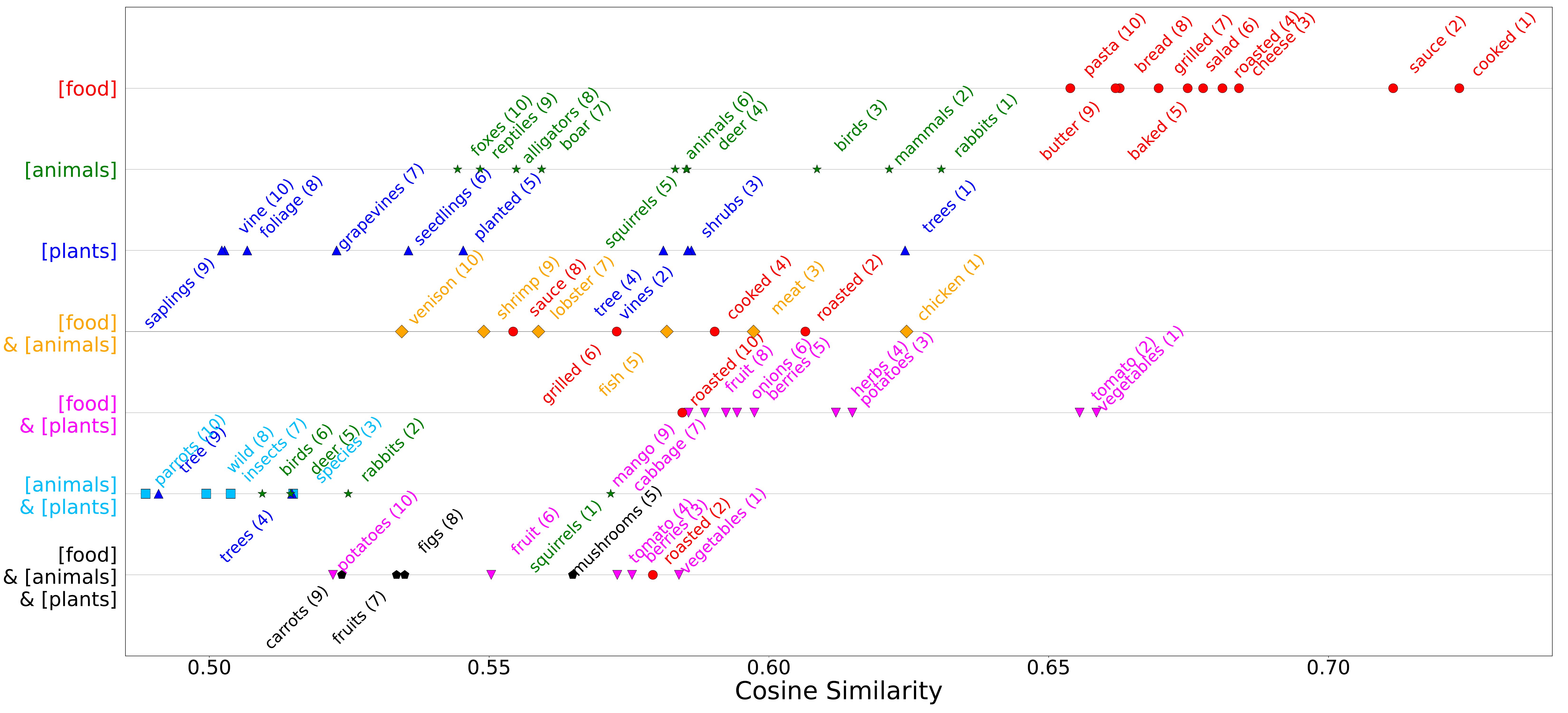}
    \caption{
Search results for seven ideal embeddings derived from ICA-transformed GloVe embeddings, containing only the semantic components of \textit{[food]}, \textit{[animals]}, and \textit{[plants]}.
For each embedding, the top $10$ words and their cosine similarities are displayed. 
The search is performed sequentially from top to bottom for each combination of semantic components. 
Each word has its rank and is assigned a color corresponding to the combination of semantic components where it first appears.
}
\label{fig:ideal-weight}
\end{figure*}

\subsubsection{Details of \texorpdfstring{$p$}{p} non-zero products}
We explain the top $p$ component-wise products used for the analogy tasks. 

Based on Appendix~\ref{app:analogy-setting}, we define $\mathbf{s}_{i_1,i_2,i_3}:=\mathbf{s}_{i_2}-\mathbf{s}_{i_1}+\mathbf{s}_{i_3}$ and normalize it as
\begin{align}
    \hat{\mathbf{s}}_{i_1,i_2,i_3}:= \frac{\mathbf{s}_{i_1,i_2,i_3}}{\|\mathbf{s}_{i_1,i_2,i_3}\|} = \left(\hat{s}_{i_1,i_2,i_3}^{(\ell)}\right)_{\ell=1}^d\in\mathbb{R}^d.\label{eq:i1-i2-i3}
\end{align}
Then, following the cosine similarity expression in (\ref{eq:cosine}), (\ref{eq:analogy}) can be rewritten as
\begin{align}
    \mathop{\argmax}_{i\in \llbracket n \rrbracket}{}\,\sum_{\ell=1}^d \hat{s}_{i_1,i_2,i_3}^{(\ell)}\,\hat{s}_i^{(\ell)}.\label{eq:analogy-cos}
\end{align}

Based on (\ref{eq:analogy-cos}), we consider the index set of top $p$ component-wise products for each word $w_i$:
\begin{align}
\text{Top}_p^i:=\mathop{\argmax}_{\ell\in \llbracket d \rrbracket}{}_p\,\hat{s}_{i_1,i_2,i_3}^{(\ell)}\,\hat{s}_i^{(\ell)}\subset\llbracket d \rrbracket.\label{eq:Top-p-i}
\end{align}
If $p=d$, then $\text{Top}_p^i=\llbracket d \rrbracket$.

Then, following the expression in (\ref{eq:analogy-cos}), we use the top $p$ component-wise products and predict the index of the answer word as follows:
\begin{align}
\mathop{\argmax}_{i\in \llbracket n \rrbracket}{}\,\sum_{\ell\in\text{Top}_p^i}\hat{s}_{i_1,i_2,i_3}^{(\ell)}\,\hat{s}_i^{(\ell)}.\label{eq:analogy-p}
\end{align}

\subsection{Results}
Table~\ref{tab:downstream} shows the results for the tasks with $p=1$, $5$, $10$, $50$, $100$, and $300$.
The performance of the ICA-transformed embeddings is better than that of the PCA-transformed embeddings.
As seen in Fig.~\ref{fig:downstream}, the difference in performance between ICA and PCA is larger for the word similarity tasks than for the analogy tasks.
This is probably because in the word similarity tasks, the semantic components are used directly, as shown in (\ref{eq:cosine-p}).
On the other hand, in the analogy tasks, the component values of the normalized vector $\hat{\mathbf{s}}_{i_1,i_2,i_3}$ are used, as shown in (\ref{eq:analogy-p}).

\subsection{Relation between the sum of the top \texorpdfstring{$p$}{p} products and cosine similarity}
As seen in Appendix~\ref{app:wordsim}, in the word similarity tasks, we compared performance using the sum of the top $p$ component-wise products for the normalized ICA-transformed and PCA-transformed GloVe embeddings.
In this section, we examine the relation between this sum and the original cosine similarity.

Figure~\ref{fig:scatter_p} shows the comparison of the scatterplots for the sum and the cosine similarity for the normalized ICA-transformed and PCA-transformed embeddings.
For smaller values of $p$, such as $p=1$ or $p=10$, the correlation of the scatterplots is stronger for ICA than for PCA.
At $p=100$, the scatterplot shows strong correlations for both ICA and PCA.

Figure~\ref{fig:cos_toppsum} compares the correlation coefficients at different values of $p$ for these normalized embeddings. 
As seen in Fig.~\ref{fig:scatter_p}, the correlation coefficients for ICA are stronger than those for PCA at smaller values of $p$. 
As $p$ increases, the correlation coefficients for both ICA and PCA increase and the difference between them decreases. 
These results indicate the favorable sparsity of the component-wise products for the normalized ICA-transformed embeddings.

\section{Retrieval of Embeddings with Selected Semantic Components}\label{app:ideal-embeddings}
In this section, we define ideal embeddings only with specific semantic components and investigate words with high cosine similarity to these embeddings. 
This analysis aims to provide a deeper understanding of the sparsity properties of ICA-transformed embeddings and the interpretation of cosine similarity presented in this study.

\subsection{Settings}
The ideal embedding $\hat{\mathbf{q}}=(\hat{q}^{(\ell)})_{\ell=1}^d\in\mathbb{R}^d$ with the semantic components at indices $\ell_1,\ldots,\ell_m$ is defined as follows\footnote{This study considers only uniform weights for simplicity.}:
\begin{align}
    \hat{q}^{(\ell)} := \begin{cases} 
1/\sqrt{m} & \text{if }\ell={\ell_1,\ldots,\ell_m}, \\
   0 & \text{otherwise},
   \end{cases}
\end{align}
where $\|\hat{\mathbf{q}}\|=1$. Note that searching using an ideal embedding with only one semantic component is the same as examining the top words of the axis of the normalized ICA-transformed embeddings.
Using the ideal embedding $\hat{\mathbf{q}}$, we search for the top $k$ words by the inner product, i.e., cosine similarity:
\begin{align}
\mathop{\argmax}_{i\in \llbracket n \rrbracket}{}_k\,\hat{\mathbf{q}}^\top\hat{\mathbf{s}}_i,
\end{align}
where $\llbracket n \rrbracket := \{1, \ldots, n\}$.

\subsection{Specific examples}
As an example, we use ideal embeddings derived from ICA-transformed GloVe embeddings.
We focus on the semantic components of \textit{[food]}, \textit{[animals]}, and \textit{[plants]}, which yield insightful results.
Figure~\ref{fig:ideal-weight} shows these results. 
The results for one semantic component show that the meanings of the selected axes represent \textit{[food]}, \textit{[animals]}, and \textit{[plants]}.
For the combinations of two semantic components, top words include \textit{chicken}, \textit{meat}, and \textit{fish} for \textit{[food]} \& \textit{[animals]}; \textit{vegetables}, \textit{tomato}, and \textit{potatoes} for \textit{[food]} \& \textit{[plants]}; and \textit{species}, \textit{insects}, and \textit{wild} for \textit{[animals]} \& \textit{[plants]}.
In addition, for \textit{[food]} \& \textit{[animals]} \& \textit{[plants]}, top words include words such as \textit{mushrooms}, \textit{fruits}, and \textit{figs}, which are plant-based food favored by animals. 
As the number of semantic components increases, the ambiguity increases, resulting in lower cosine similarity values.

\renewcommand{\arraystretch}{1.2} 
\begin{table*}[t!]
\small
\centering
\begin{subtable}[t]{.48\textwidth}
\centering
\begin{adjustbox}{max width=0.93\linewidth}
\begin{tabular}{lll|r|r}
\toprule
$w_i$ & \textbf{M} & \textbf{F} &  $\hat{\mathbf{s}}_{i_\text{\textit{woman}}}{}^\top\hat{\mathbf{s}}_i$ & $\hat{\mathbf{s}}_{i_\text{\textit{woman}}\ominus\ell_{\text{[female]}}}{}^\top\hat{\mathbf{s}}_i$ \\
\midrule
\textit{girl} & & \checkmark  & $0.691$ & $0.595$ ({\color{red}$\downarrow0.096$}) \\
\textit{man}& \checkmark &  & $0.679$ & 0.720 ({\color{blue}$\uparrow0.041$}) \\
\textit{mother} & & \checkmark  & $0.632$ & 0.467 ({\color{red}$\downarrow0.165$}) \\
\textit{person}& \checkmark & \checkmark  & $0.579$ & 0.576 ({\color{red}$\downarrow0.003$}) \\
\textit{female} & & \checkmark  & $0.575$ & 0.518 ({\color{red}$\downarrow0.057$}) \\
\textit{she} & & \checkmark  & $0.568$ & 0.422 ({\color{red}$\downarrow0.146$}) \\
\textit{herself} & & \checkmark  & $0.567$ & 0.429 ({\color{red}$\downarrow0.138$}) \\
\textit{wife} & & \checkmark  & $0.553$ & 0.392 ({\color{red}$\downarrow0.161$}) \\
\textit{women} & & \checkmark  & $0.544$ & 0.468 ({\color{red}$\downarrow0.076$}) \\
\textit{daughter} & & \checkmark  & $0.535$ & 0.382 ({\color{red}$\downarrow0.153$}) \\
\bottomrule
\end{tabular}
\end{adjustbox}
\caption{\textit{woman} top $10$ words}
\label{tab:woman_top}
\end{subtable}
\begin{subtable}[t]{.48\textwidth}
\centering
\begin{adjustbox}{max width=0.93\linewidth}
\begin{tabular}{lll|r|r}
\toprule
$w_i$ & \textbf{M} & \textbf{F} &  $\hat{\mathbf{s}}_{i_\text{\textit{woman}}}{}^\top\hat{\mathbf{s}}_i$ & $\hat{\mathbf{s}}_{i_\text{\textit{woman}}\ominus\ell_{\text{[female]}}}{}^\top\hat{\mathbf{s}}_i$ \\
\midrule
\textit{man} & \checkmark & & $0.679$ & $0.720$ ({\color{blue}$\uparrow0.041$}) \\
\textit{girl} & & \checkmark & $0.691$ & $0.595$ ({\color{red}$\downarrow0.096$}) \\
\textit{person} & \checkmark & \checkmark & $0.579$ & $0.576$ ({\color{red}$\downarrow0.003$}) \\
\textit{female} & & \checkmark & $0.575$ & $0.518$ ({\color{red}$\downarrow0.057$}) \\
\textit{boy} & \checkmark & & $0.497$ & $0.509$ ({\color{blue}$\uparrow0.012$}) \\
\textit{someone} & \checkmark & \checkmark & $0.490$ & $0.482$ ({\color{red}$\downarrow0.008$}) \\
\textit{teenager} & \checkmark & \checkmark & $0.524$ & $0.481$ ({\color{red}$\downarrow0.044$}) \\
\textit{men} & \checkmark & & $0.432$ & $0.471$ ({\color{blue}$\uparrow0.039$}) \\
\textit{victim} & \checkmark & \checkmark & $0.507$ & $0.471$ ({\color{red}$\downarrow0.036$}) \\
\textit{women} & & \checkmark & $0.544$ & $0.468$ ({\color{red}$\downarrow0.076$}) \\
\bottomrule
\end{tabular}
\end{adjustbox}
\caption{\textit{woman}$\ominus$\textit{[female]} top $10$ words}
\label{tab:woman_female_top}
\end{subtable}
\caption{
The top $10$ words for (a) \textit{woman} and (b) \textit{woman}$\ominus$\textit{[female]} are shown with their inner product values against both \textit{woman} and \textit{woman}$\ominus$\textit{[female]}.
Check marks indicate whether words are masculine (\textbf{M}) or feminine (\textbf{F}).
The differences between their inner product values for \textit{woman} and \textit{woman}$\ominus$\textit{[female]}, 
computed for each word $w_i$ as $\hat{\mathbf{s}}_{i_\text{\textit{woman}}\ominus\ell_{\text{[female]}}}{}^\top\hat{\mathbf{s}}_i - \hat{\mathbf{s}}_{i_\text{\textit{woman}}}{}^\top\hat{\mathbf{s}}_i$, 
are indicated with {\color{blue}$\uparrow$} for increases and {\color{red}$\downarrow$} for decreases, with the magnitude shown as the absolute value.
}
\label{tab:woman_top_and_woman_female_top}
\end{table*}
\renewcommand{\arraystretch}{1.0} 

\begin{figure*}[t!]
    \centering
    \includegraphics[width=\textwidth]{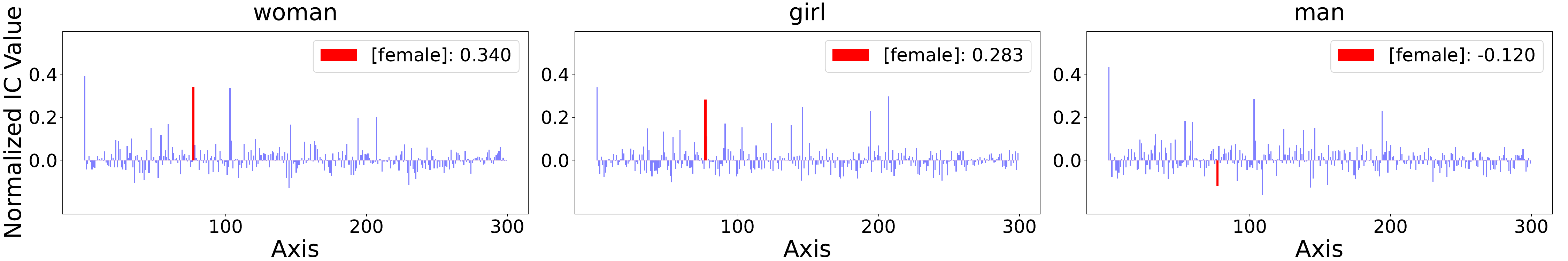}
\caption{
Bar graphs of the normalized ICA-transformed GloVe embeddings of \textit{woman}, \textit{girl}, and \textit{man}, showing each component value of \textit{[female]}.
}
    \label{fig:woman_girl_man_bar}
\end{figure*}

\section{Retrieval of Embeddings after Ablating a Semantic Component}\label{app:woman-girl-man}
In this section, we ablate one semantic component from a normalized ICA-transformed embedding, effectively setting the semantic similarity on the axis to zero. We then examine how the words with high similarity change before and after this ablation.
This analysis aims to explore the impact of changes in a single semantic similarity and to provide an illustrative example for interpreting cosine similarity as the sum of semantic similarities.

\subsection{Settings}
By setting the $\ell_\ast$-th semantic component to zero for the normalized ICA-transformed embeddings, we define the ideal embedding of the $i_\ast$-th word as
\begin{align}
    \hat{\mathbf{s}}_{i_\ast\ominus\ell_\ast} := (\hat{s}_{i_\ast}^{(1)},\ldots,\hat{s}_{i_\ast}^{(\ell_\ast-1)},0,\hat{s}_{i_\ast}^{(\ell_\ast+1)}\ldots,\hat{s}_{i_\ast}^{(d)})\label{eq:ablation}.
\end{align}
Since $\|\hat{\mathbf{s}}_{i_\ast}\|=1$, it follows that $\|\hat{\mathbf{s}}_{i_\ast\ominus\ell_\ast}\|\leq1$. 
For $\hat{\mathbf{s}}_{i_\ast}$ and $ \hat{\mathbf{s}}_{i_\ast\ominus\ell_\ast}$, based on (\ref{eq:cosine}), we search for the top $k$ words by their inner products as follows:
\begin{align}
&\mathop{\argmax}_{i\in\llbracket n \rrbracket\setminus\{i_\ast\}}{}_k\,\hat{\mathbf{s}}_{i_\ast}{}^\top\hat{\mathbf{s}}_i,\label{eq:woman1}\\
&\mathop{\argmax}_{i\in\llbracket n \rrbracket\setminus\{i_\ast\}}{}_k\,\hat{\mathbf{s}}_{i_\ast\ominus\ell_\ast}{}^\top\hat{\mathbf{s}}_i.\label{eq:woman2}
\end{align}
Note that the original $i_\ast$-th word is excluded from the candidates.
In particular, as evident from the definition of $\hat{\mathbf{s}}_{i_\ast\ominus\ell_\ast}$ in (\ref{eq:ablation}), (\ref{eq:woman2}) considers the situation where the $\ell_\ast$-th semantic similarity is set to zero in the cosine similarity in (\ref{eq:cosine-sem}).

\subsection{Specific examples}
As an example, consider the ablation of a semantic component in the normalized ICA-transformed GloVe embedding of \textit{woman}. 
We focused on one of the axes with large component values in the embedding, where the top $5$ words were \textit{her}, \textit{wife}, \textit{mother}, \textit{daughter}, and \textit{actress}. 
Therefore, we interpret the meaning of this axis as \textit{[female]}.
We then set the semantic component of \textit{[female]} in the embedding to zero and define the ideal embedding. 
We call the corresponding ideal word for this ideal embedding \textit{woman}$\ominus$\textit{[female]}.

\subsubsection{Top words and their inner products}
Based on (\ref{eq:woman1}), we first searched for the top $10$ words of \textit{woman}, and Table~\ref{tab:woman_top} shows their inner product values with both \textit{woman} and \textit{woman}$\ominus$\textit{[female]}.
Words such as \textit{mother}, \textit{wife}, and \textit{daughter}, which have high inner product values with \textit{woman}, have lower values with \textit{woman}$\ominus$\textit{[female]}. 
Conversely, \textit{man} has higher values with \textit{woman}$\ominus$\textit{[female]} than with \textit{woman}.

Next, based on (\ref{eq:woman2}), for the top $10$ words of \textit{woman}$\ominus$\textit{[female]}, Table~\ref{tab:woman_female_top} shows their inner product values with both \textit{woman} and \textit{woman}$\ominus$\textit{[female]}.
Compared to the top $10$ words of \textit{woman}, that of \textit{woman}$\ominus$\textit{[female]} includes masculine words such as \textit{boy} and \textit{men}, as well as words like \textit{someone} and \textit{teenager} that are both masculine and feminine.
Furthermore, for the masculine words, the inner product values with \textit{woman}$\ominus$\textit{[female]} are larger than those with \textit{woman}.
On the other hand, for feminine words (including those that are both masculine and feminine), the opposite trend is observed.

\begin{figure}[t!]
    \centering
    \begin{minipage}{0.48\linewidth}
        \centering
        \includegraphics[width=\linewidth]{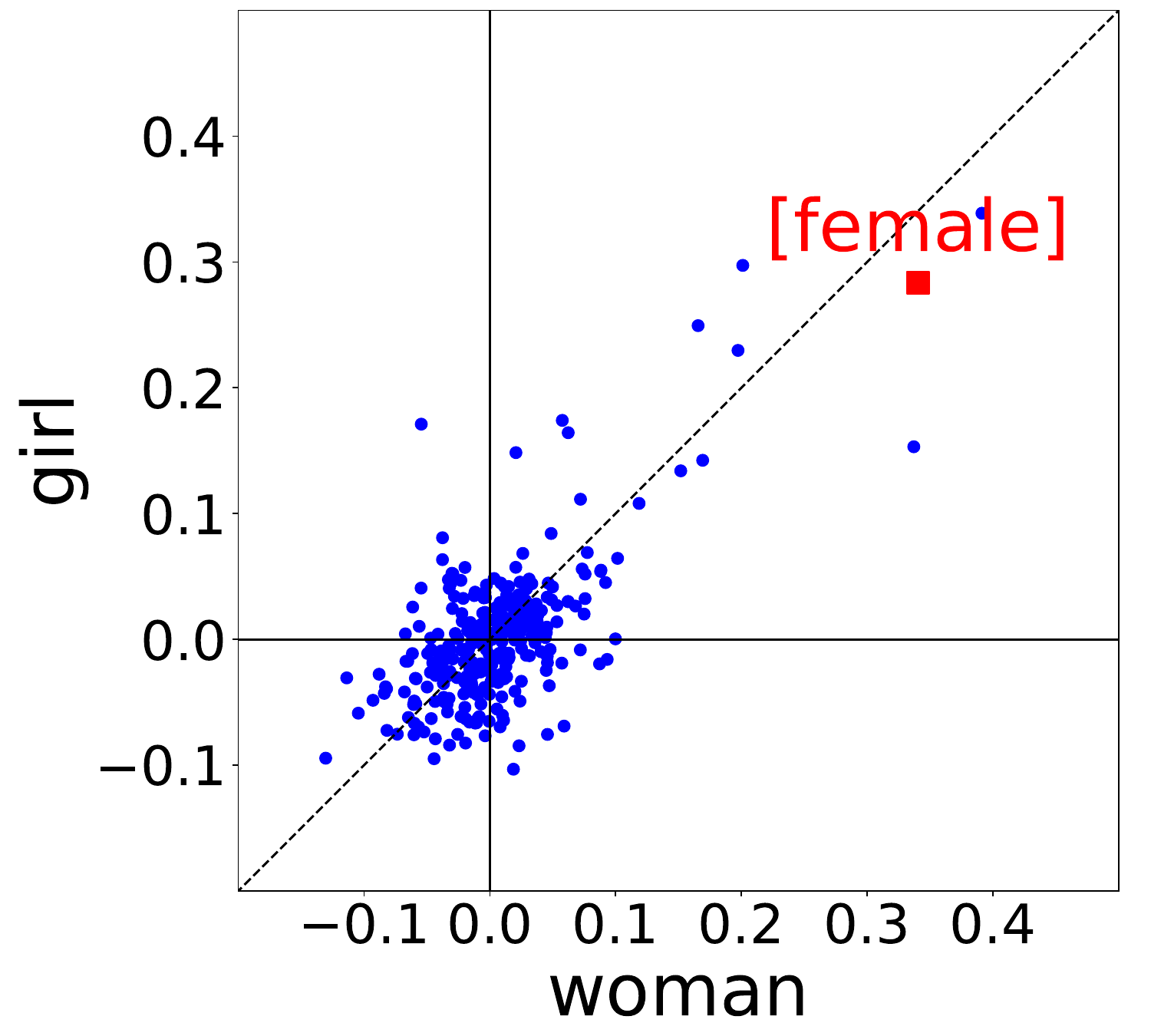}
        \subcaption{\textit{woman} and \textit{girl}}
        \label{fig:woman_girl}
    \end{minipage}\hfill
    \begin{minipage}{0.48\linewidth}
        \centering
        \includegraphics[width=\linewidth]{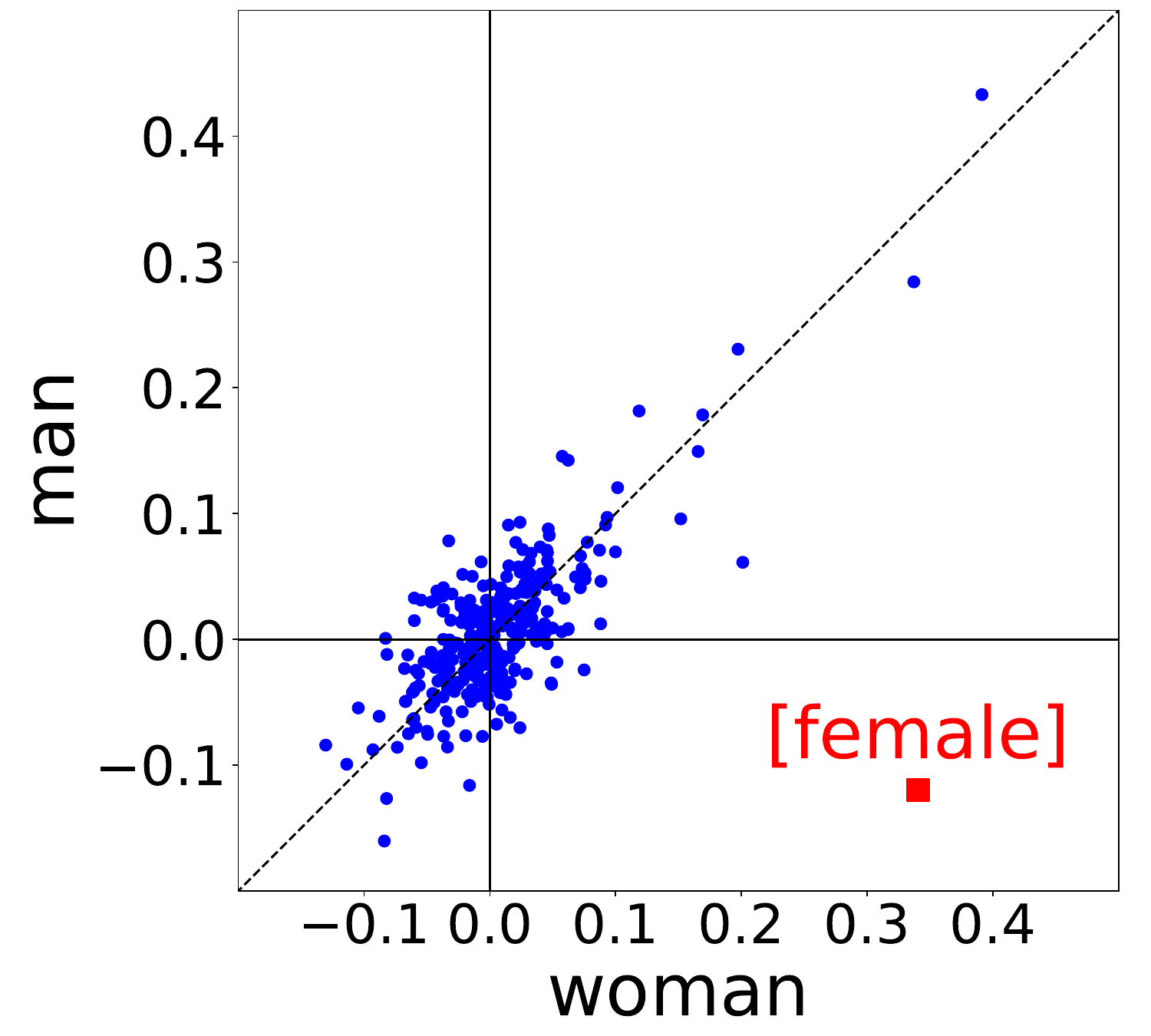}
        \subcaption{\textit{woman} and \textit{man}}
        \label{fig:woman_man}
    \end{minipage}
    \caption{
Scatterplots of component values of normalized ICA-transformed GloVe embeddings for (a) the pair \textit{woman} and \textit{girl} and (b) the pair \textit{woman} and \textit{man}. 
The semantic component of \textit{[female]} is marked with {\color{red}$\blacksquare$} and others with {\color{blue}$\bullet$}. 
Figure~\ref{fig:woman_girl_man_bar} in Appendix~\ref{app:woman-girl-man} shows bar graphs for each embedding.
}
    \label{fig:woman_girl_man}
\end{figure}

\begin{figure}[t!]
    \centering
    \includegraphics[width=\columnwidth]{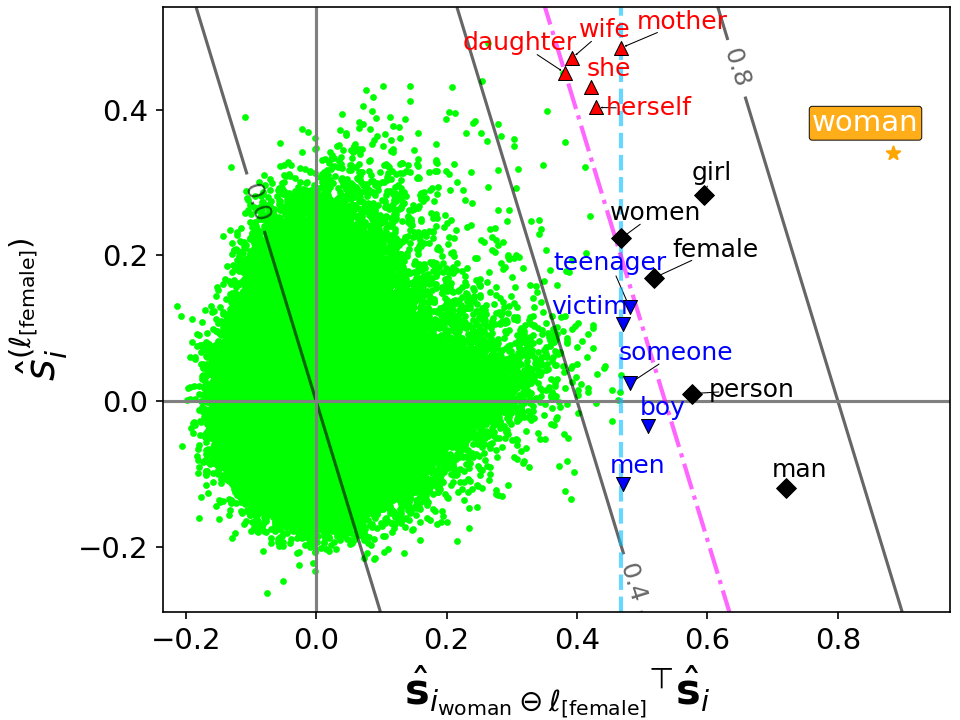}
    \caption{
Scatterplot of $\hat{\mathbf{s}}_{i_\textit{woman}\ominus\ell_{\textit{[female]}}}{}^\top\hat{\mathbf{s}}_i$ and $\hat{s}_i^{(\ell_{\textit{[female]}})}$ for each $i\in\llbracket n \rrbracket$. 
The position of \textit{woman} is marked with {\color{orange}$\star$}. 
Words from Table~\ref{tab:woman_top} (\textit{woman} top 10 words) and Table~\ref{tab:woman_female_top} (\textit{woman}$\ominus$\textit{[female]} top 10 words) are categorized as follows:
Words that appear in the top 10 for both \textit{woman} and \textit{woman}$\ominus$\textit{[female]} are marked with {\color{black}$\blacklozenge$}, those that appear only in the top 10 for \textit{woman} with {\color{red}$\blacktriangle$}, those that appear only in the top 10 for \textit{woman}$\ominus$\textit{[female]} with {\color{blue}$\blacktriangledown$}. Other words are marked with {\color{green}$\bullet$}. 
The value of $\hat{\mathbf{s}}_{i_\textit{woman}}{}^\top\hat{\mathbf{s}}_i$, which represents the cosine similarity $\cos{(\textit{woman}, w_i)}$ between \textit{woman} and the word $w_i$, is plotted as contours. 
The contour for the 10th closest word \textit{daughter} to \textit{woman} is shown as a \pink{dash-dotted} line. 
For the 10th closest word \textit{women} to \textit{woman}$\ominus$\textit{[female]}, their inner product $\hat{\mathbf{s}}_{i_\textit{woman}\ominus\ell_{\textit{[female]}}}{}^\top\hat{\mathbf{s}}_{i_\textit{women}}$ is shown as a {\color{cyan}dashed} line.
}
\label{fig:prediction_change}
\end{figure}

\subsection{Bar graphs and scatterplots of the embeddings for \textit{woman}, \textit{girl}, and \textit{man}}
In Table~\ref{tab:woman_top_and_woman_female_top}, \textit{girl} has a high inner product value with \textit{woman} but a low inner product value with \textit{woman}$\ominus$\textit{[female]}. 
Conversely, \textit{man} exhibits the opposite trend.
Therefore, in this section, we take a closer look at the normalized ICA-transformed GloVe embeddings of \textit{woman}, \textit{girl}, and \textit{man}.

Figure~\ref{fig:woman_girl_man_bar} shows bar graphs of these three embeddings.
The graphs for \textit{woman} and \textit{girl} are similar in shape.
On the other hand,  the graphs for \textit{woman} and \textit{man} show differences in the semantic component of \textit{[female]}, although other parts are broadly similar.

In addition, Fig.~\ref{fig:woman_girl_man} shows scatterplots of the normalized ICA-transformed GloVe embeddings for the pair \textit{woman} and \textit{girl}, and the pair \textit{woman} and \textit{man}. 
For the semantic component of \textit{[female]}, both the embeddings of \textit{woman} and \textit{girl} have positive values, while the embedding of \textit{man} has negative values. 
Similar to Fig.~\ref{fig:woman_girl_man_bar}, these results also show that other semantic components for both pairs are roughly correlated.

\subsection{Analysis of changes in top words}
The changes in the top words before and after ablating the semantic component of \textit{[female]} are shown in Table~\ref{tab:woman_top_and_woman_female_top}.
In this section, we examine these changes in detail.

Figure~\ref{fig:prediction_change} shows a scatterplot of the inner product with \textit{woman}$\ominus$\textit{[female]} and the semantic component of \textit{[female]} for every word.
Words in the upper right are close to \textit{woman}, while those with higher values on the horizontal axis are closer to \textit{woman}$\ominus$\textit{[female]}. 
The transition from \pink{dash-dotted} to {\color{cyan}dashed} lines indicates that the ablation for \textit{[female]} replaces the top words for \textit{woman} such as \textit{mother}, \textit{she}, \textit{herself}, \textit{wife}, and \textit{daughter} with the top words for \textit{woman}$\ominus$\textit{[female]} such as \textit{boy}, \textit{someone}, \textit{teenager}, \textit{men}, and \textit{victim}.
These results show that setting one semantic similarity to zero can significantly alter the cosine similarity in (\ref{eq:cosine-sem}).

As related work, \citet{DBLP:conf/acl/IshibashiSYN20} proposed a method to invert the meaning of a word by mirroring its embedding across a hyerplane.
In contrast, we invert the meaning by simply ablating the component from a particular axis.

\section{Case Study of ICA-transformed embeddings}\label{app:case_study}
In this section, we introduce interesting case studies of ICA-transformed embeddings using the embeddings published by \citet{DBLP:conf/emnlp/YamagiwaOS23}.

\subsection{Comparison of the noun and verb senses of \textit{shore} in ICA-transformed BERT Embeddings}\label{app:bert}

\begin{figure*}[p]
    \centering
    \includegraphics[width=\textwidth]{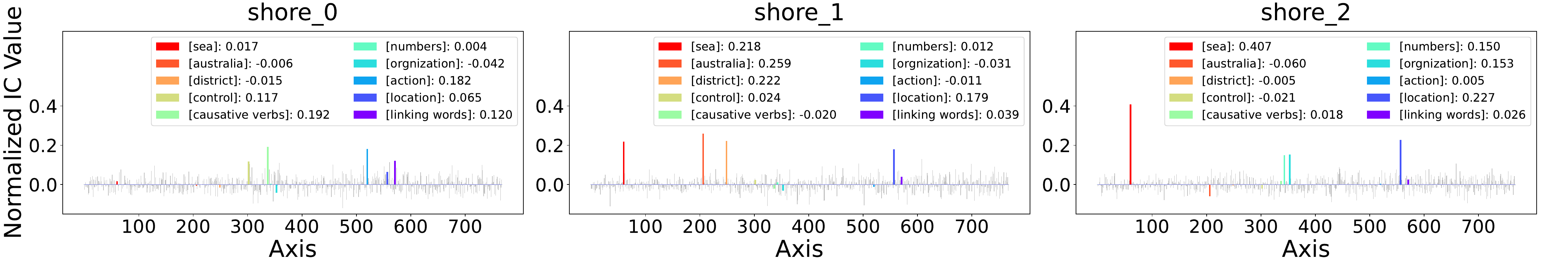}
    \caption{
Bar graphs of the normalized ICA-transformed BERT embeddings for \textit{shore\_0}, \textit{shore\_1}, and \textit{shore\_2}.
The ten axes whose component values are large for these \textit{shore} are interpreted and colored.
While \textit{shore\_0} is a verb, \textit{shore\_1} and \textit{shore\_2} are nouns.
See Table~\ref{tab:bert-topwords} for the top words of the axes and Table~\ref{tab:shore_sentences} for the sentences containing each \textit{shore}.
}
\label{fig:shore_examples}
\end{figure*}

\citet{DBLP:conf/emnlp/YamagiwaOS23} observed that the noun \textit{shore} has a large component value of \textit{[sea]}, while that of the verb \textit{shore} is small. We use their embeddings to examine the semantic components across all axes. Figure~\ref{fig:shore_examples} shows the normalized ICA-transformed embeddings of these three \textit{shore} instances as bar graphs.
For the embeddings of \textit{shore\_1} and \textit{shore\_2}, the semantic components of \textit{[sea]} and \textit{[location]} are large.
For the embedding of \textit{shore\_0}, these are small, but those of \textit{[control]} and \textit{[causative verbs]} are large. These results explain the large and small relations in the cosine similarity: $\cos(\text{\textit{shore\_1}},\text{\textit{shore\_2}})=0.299$, while $\cos(\text{\textit{shore\_0}},\text{\textit{shore\_1}})=0.054$ and $\cos(\text{\textit{shore\_0}},\text{\textit{shore\_2}})=0.128$.
Table~\ref{tab:bert-topwords} shows the top 10 words of the 10 axes selected by choosing the axes of the top 4 component values for each \textit{shore}, excluding duplicates. 

Additionally, the sentences containing these tokens are shown in Table~\ref{tab:shore_sentences}.
While \textit{shore\_0} is a verb, both \textit{shore\_1} and \textit{shore\_2} are nouns. Comparing the nouns \textit{shore\_1} and \textit{shore\_2}, we find that since \textit{shore\_1} is specifically part of \textit{Sydney's North Shore}, the normalized ICA-transformed BERT embedding of \textit{shore\_1} in Fig.~\ref{fig:shore_examples} has a large semantic component of \textit{[australia]}. This result also shows that the BERT embeddings are well contextualized.

\subsection{Comparison of ICA-transformed fastText embeddings in multiple languages}\label{app:crosslingual}

\citet{DBLP:conf/emnlp/YamagiwaOS23} showed that common semantic axes exist between ICA-transformed embeddings of different languages, and matched these axes by their correlations.
They used the fastText~\cite{DBLP:journals/tacl/BojanowskiGJM17} embeddings for their experiments, and their ICA-transformed and PCA-transformed fastText embeddings are published\footnote{\url{https://github.com/shimo-lab/Universal-Geometry-with-ICA}}.

\begin{figure*}[p]
\centering
\begin{subfigure}{0.33\linewidth}
\includegraphics[width=\linewidth]{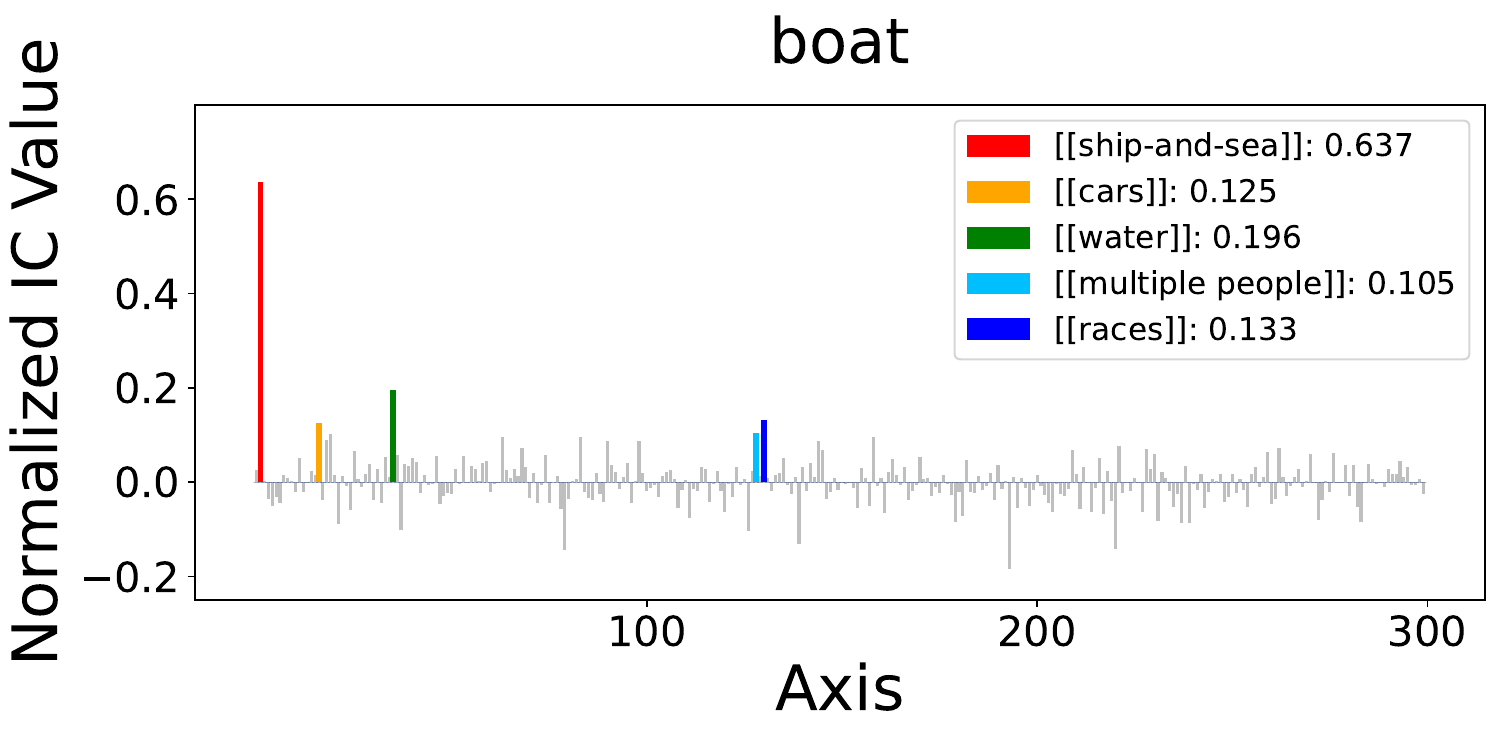} 
\caption{English}
\label{fig:en_ica}
\end{subfigure}
\par\smallskip
\begin{subfigure}{0.33\linewidth}
\includegraphics[width=\linewidth]{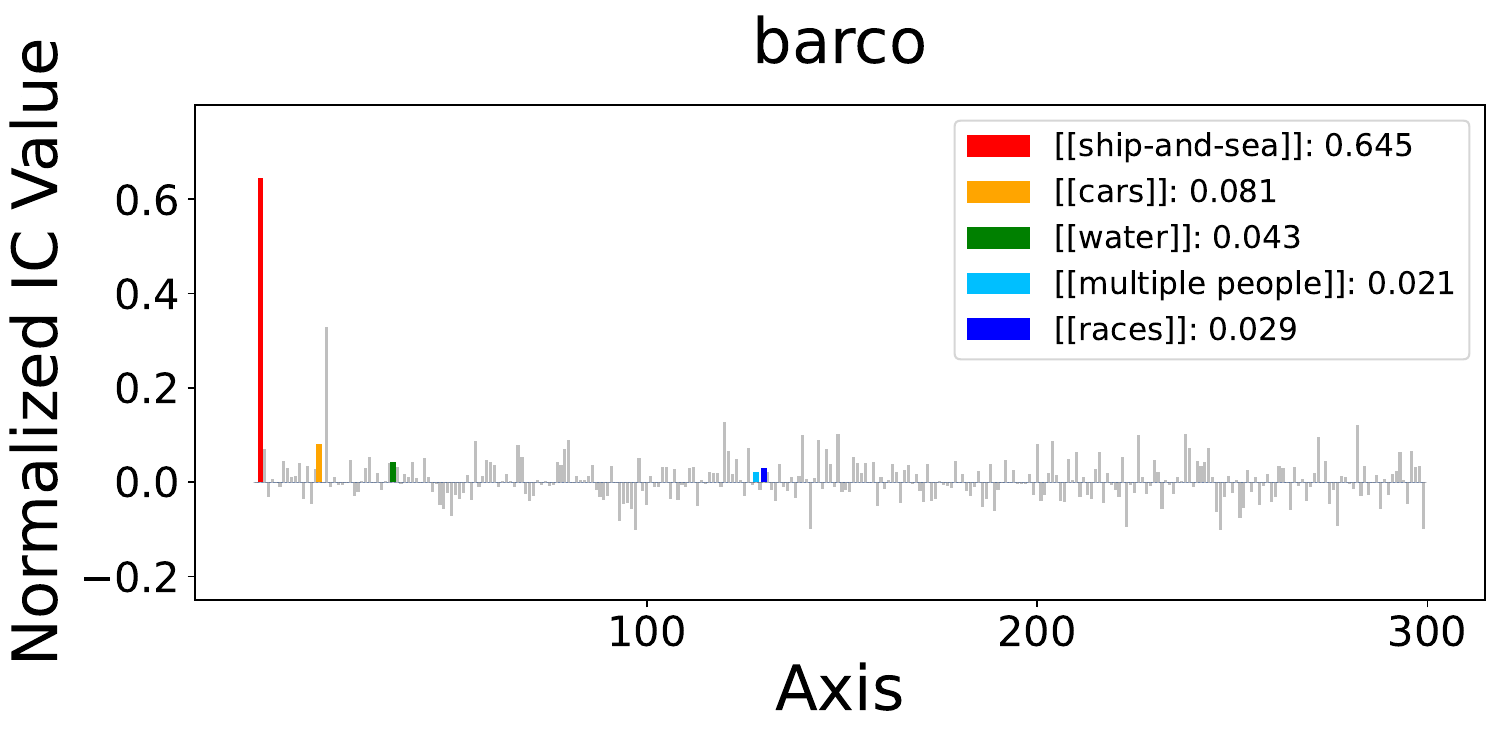}
\caption{Spanish}
\label{fig:es_ica}
\end{subfigure}\hfill
\begin{subfigure}{0.33\linewidth}
\includegraphics[width=\linewidth]{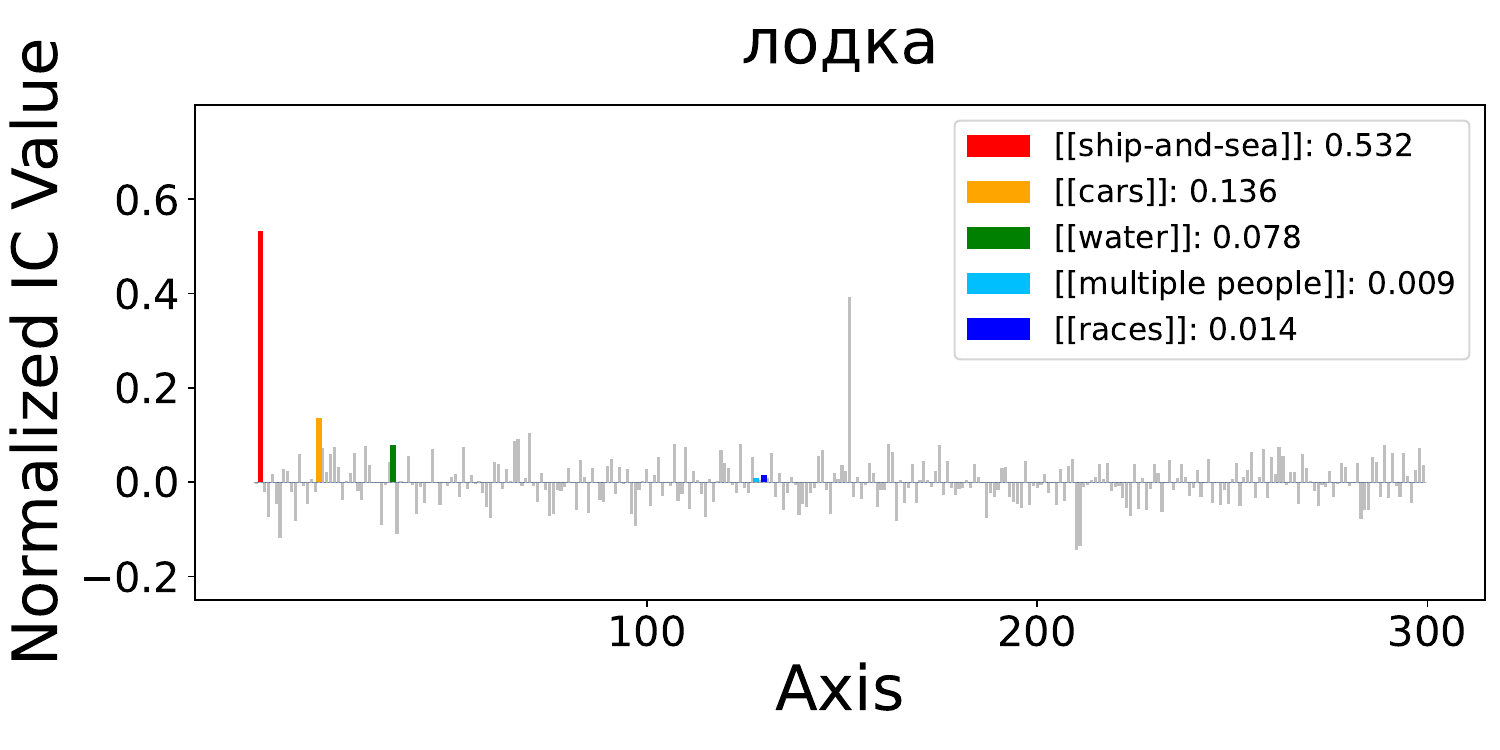}
\caption{Russian}
\label{fig:ru_ica}
\end{subfigure}\hfill
\begin{subfigure}{0.33\linewidth}
\includegraphics[width=\linewidth]{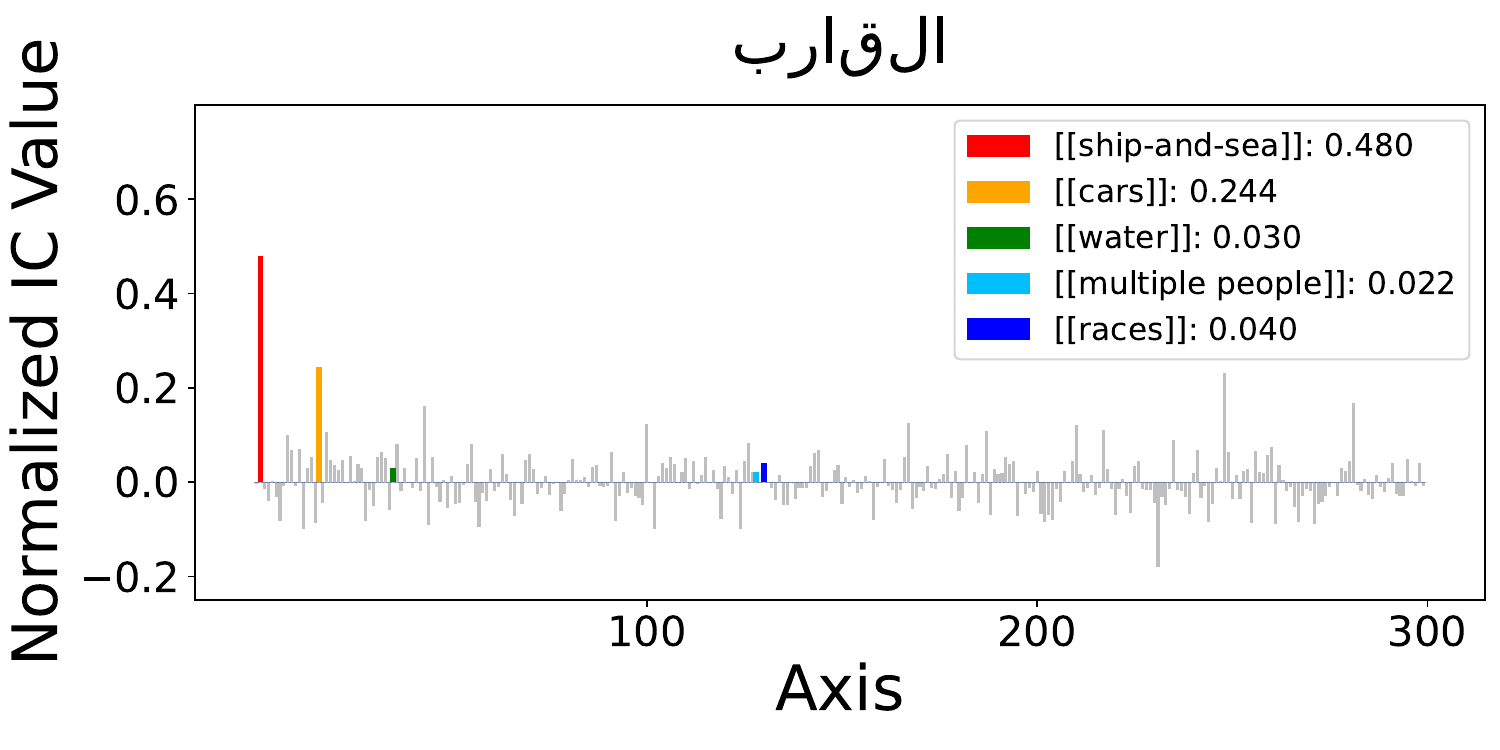}
\caption{Arabic}
\label{fig:ar_ica}
\end{subfigure}
\par\smallskip
\begin{subfigure}{0.33\linewidth}
\includegraphics[width=\linewidth]{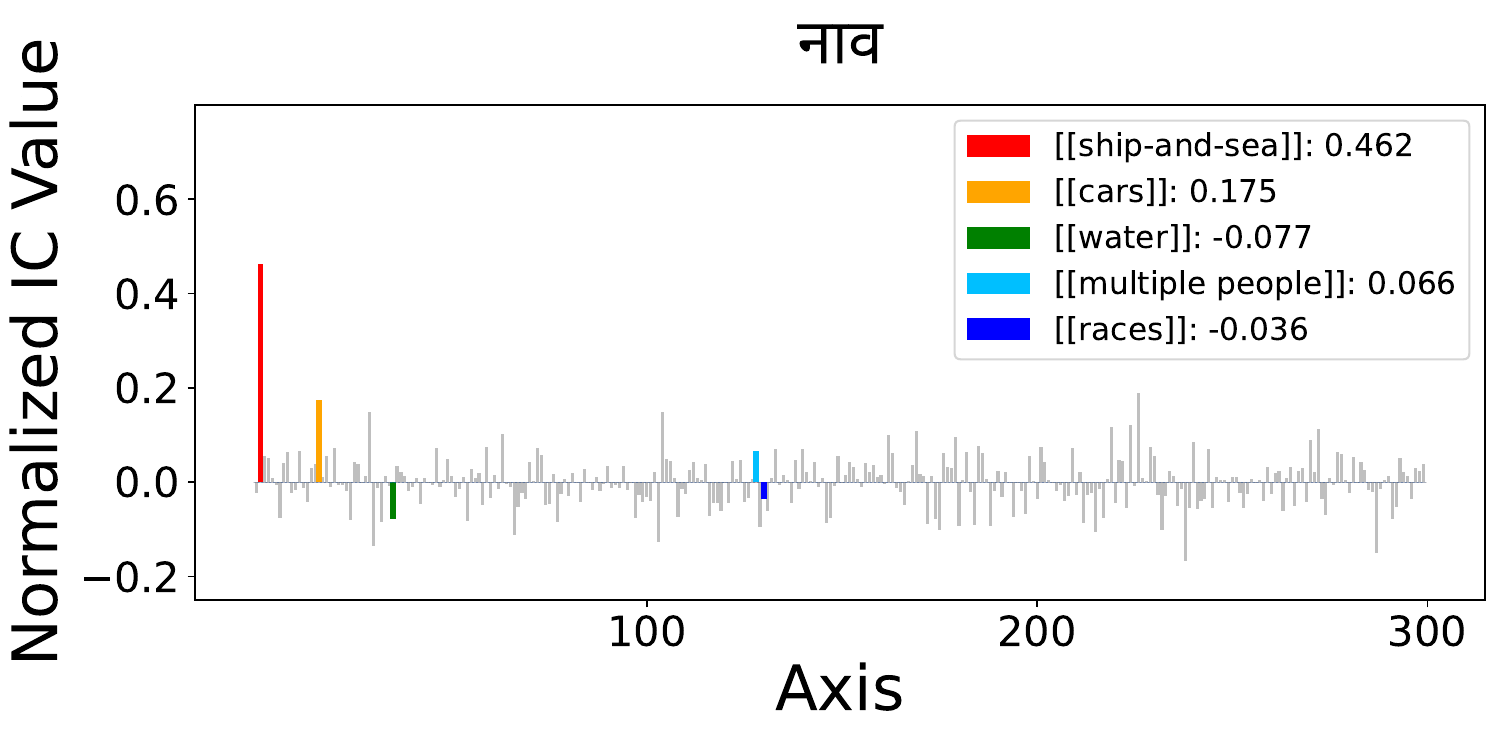}
\caption{Hindi}
\label{fig:hi_ica}
\end{subfigure}\hfill
\begin{subfigure}{0.33\linewidth}
\includegraphics[width=\linewidth]{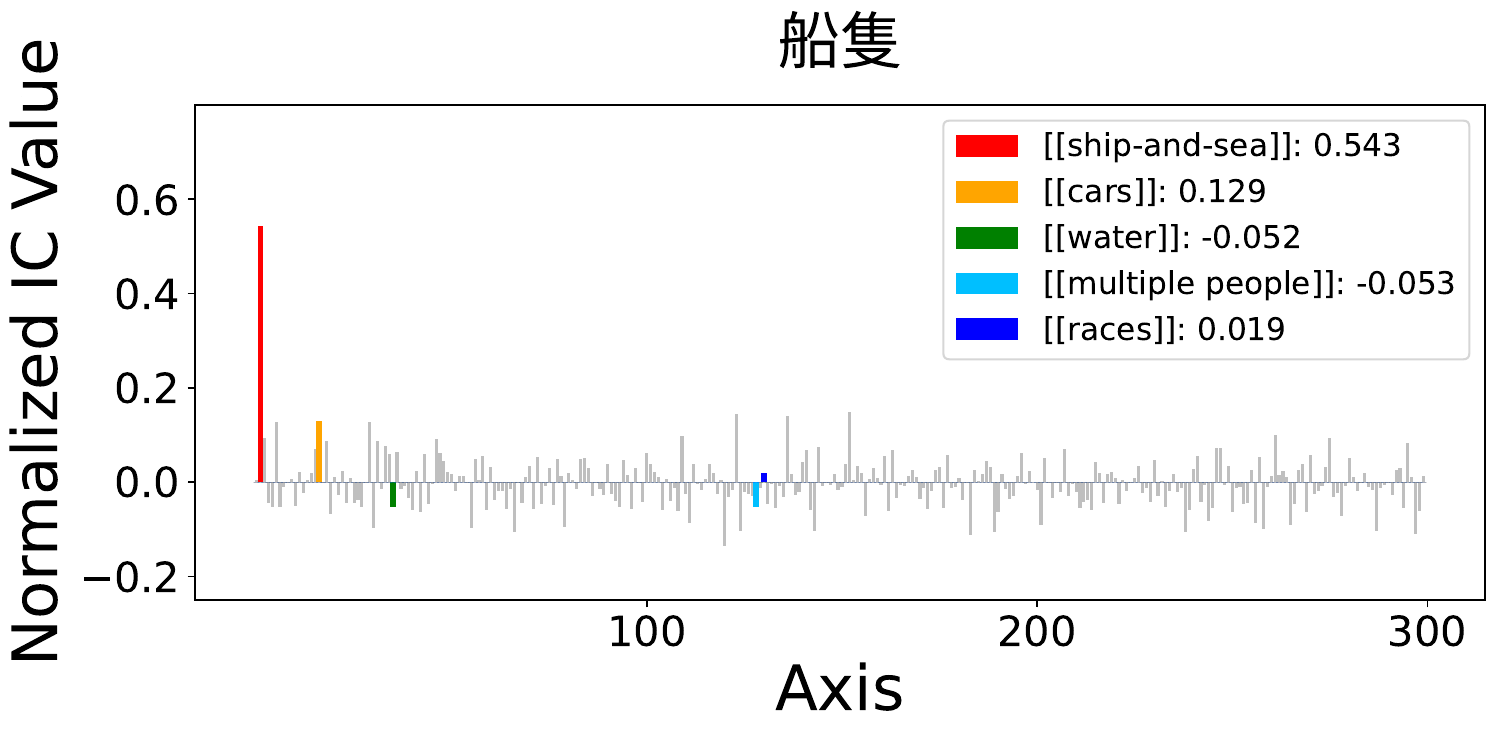}
\caption{Chinese}
\label{fig:zh_ica}
\end{subfigure}\hfill
\begin{subfigure}{0.33\linewidth}
\includegraphics[width=\linewidth]{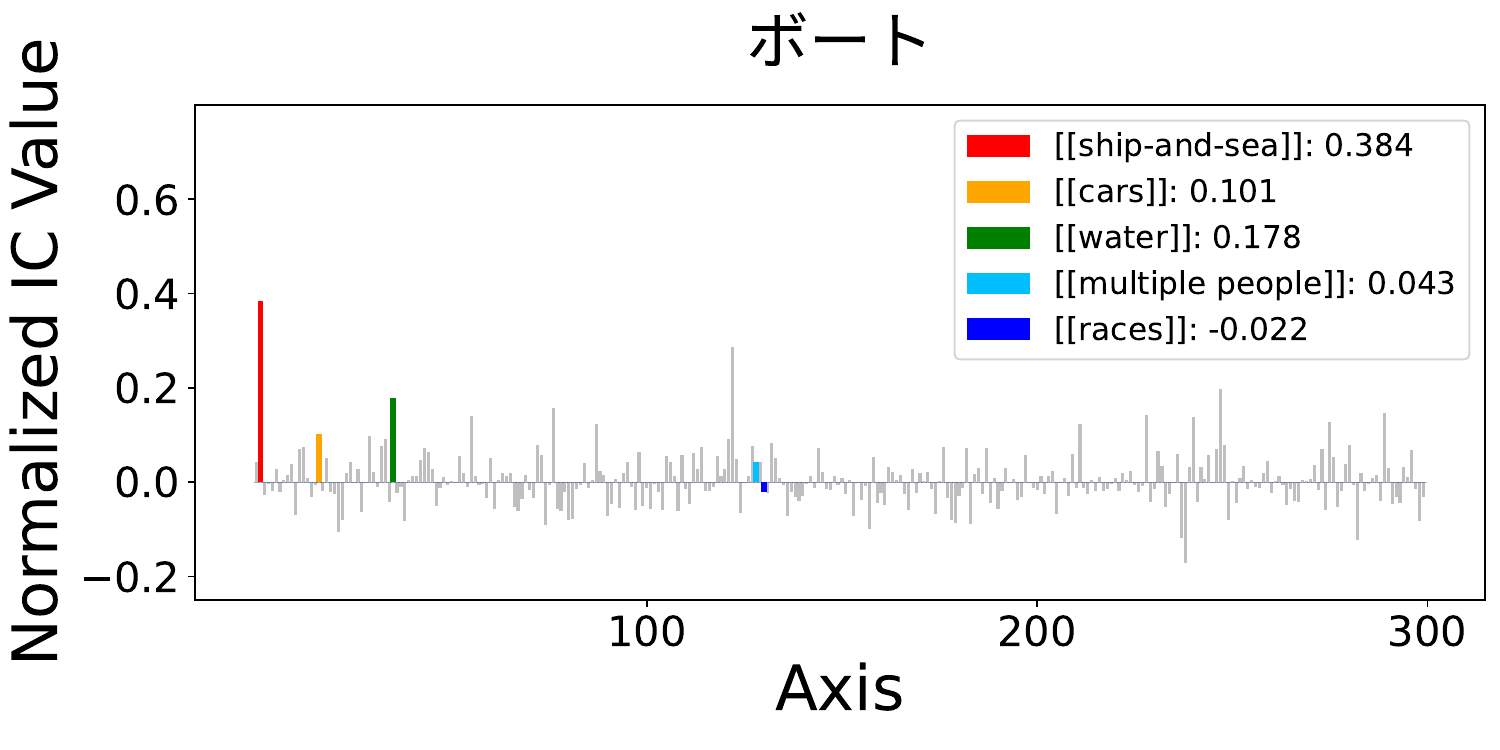}
\caption{Japanese}
\label{fig:ja_ica}
\end{subfigure}
\caption{
For \textit{boat} and its translations, the normalized ICA-transformed fastText embeddings are shown as bar graphs.
These axes are aligned by the correlation coefficients between English and the other languages.
The axes of the top 5 component values in English are highlighted with their meanings.
The component values of these axes are shown in the bar graphs for each language.
See Table~\ref{tab:boat-topwords} for the top 10 words of these axes.
}
\label{fig:ica-lang7}
\end{figure*}

\begin{figure*}[p]
\centering
\begin{subfigure}{0.33\linewidth}
\includegraphics[width=\linewidth]{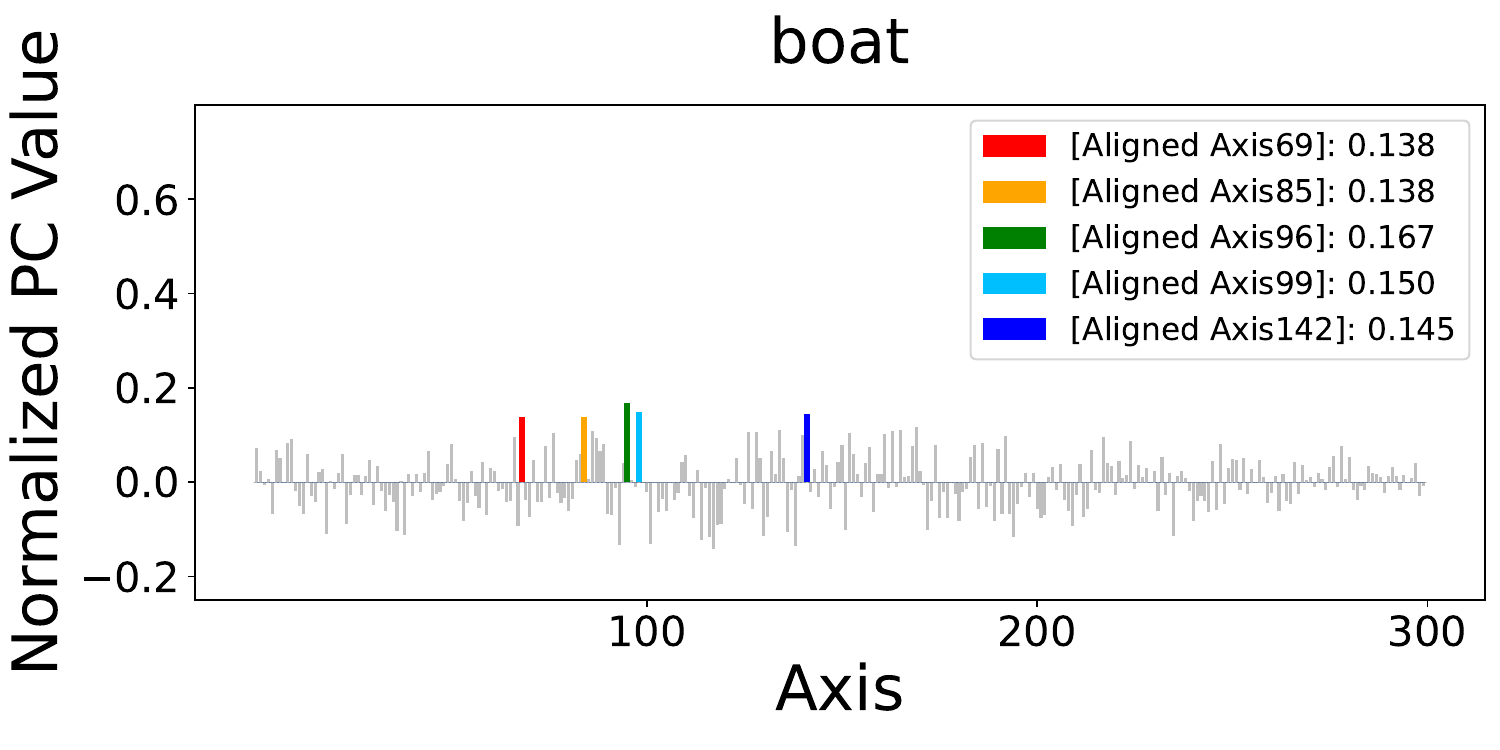} 
\caption{English}
\label{fig:en_pca}
\end{subfigure}
\par\smallskip
\begin{subfigure}{0.33\linewidth}
\includegraphics[width=\linewidth]{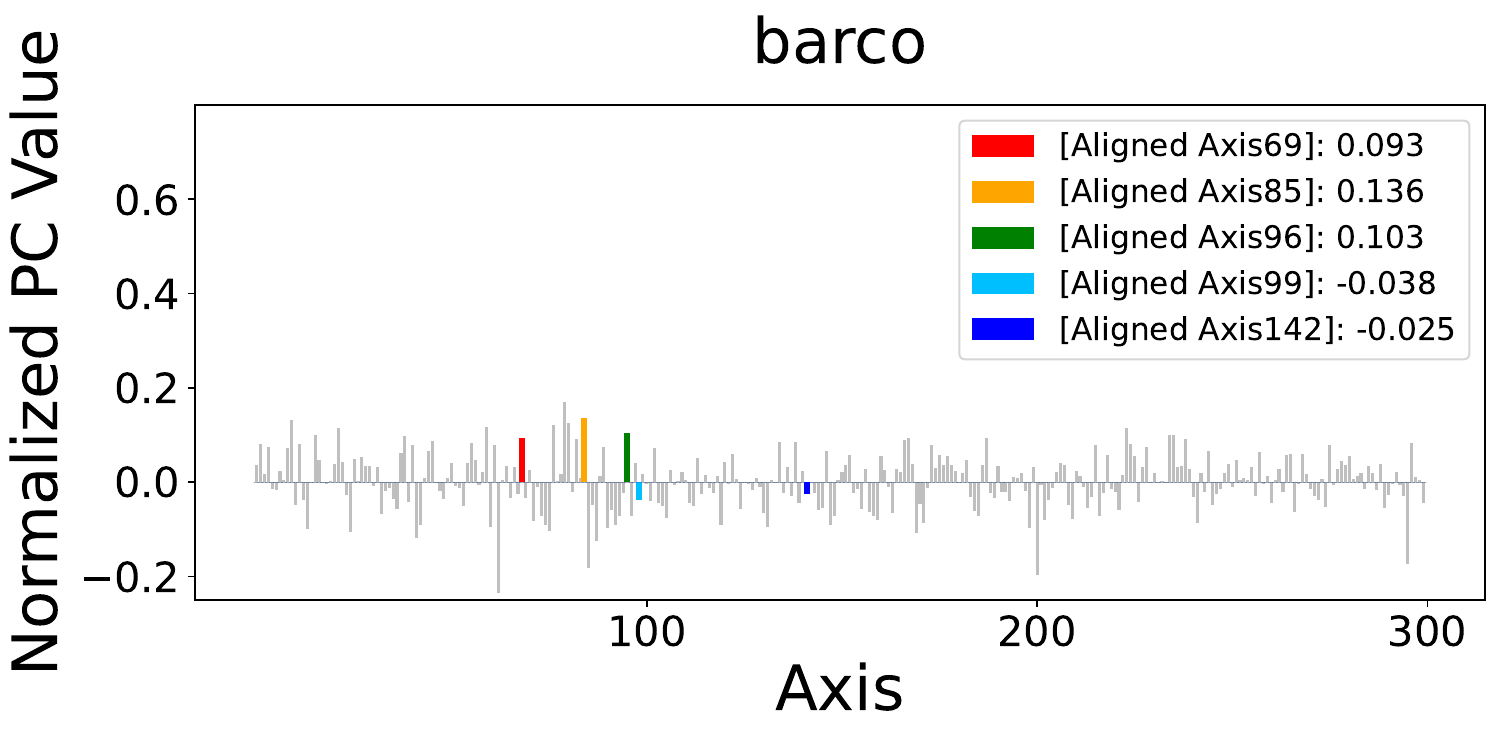}
\caption{Spanish}
\label{fig:es_pca}
\end{subfigure}\hfill
\begin{subfigure}{0.33\linewidth}
\includegraphics[width=\linewidth]{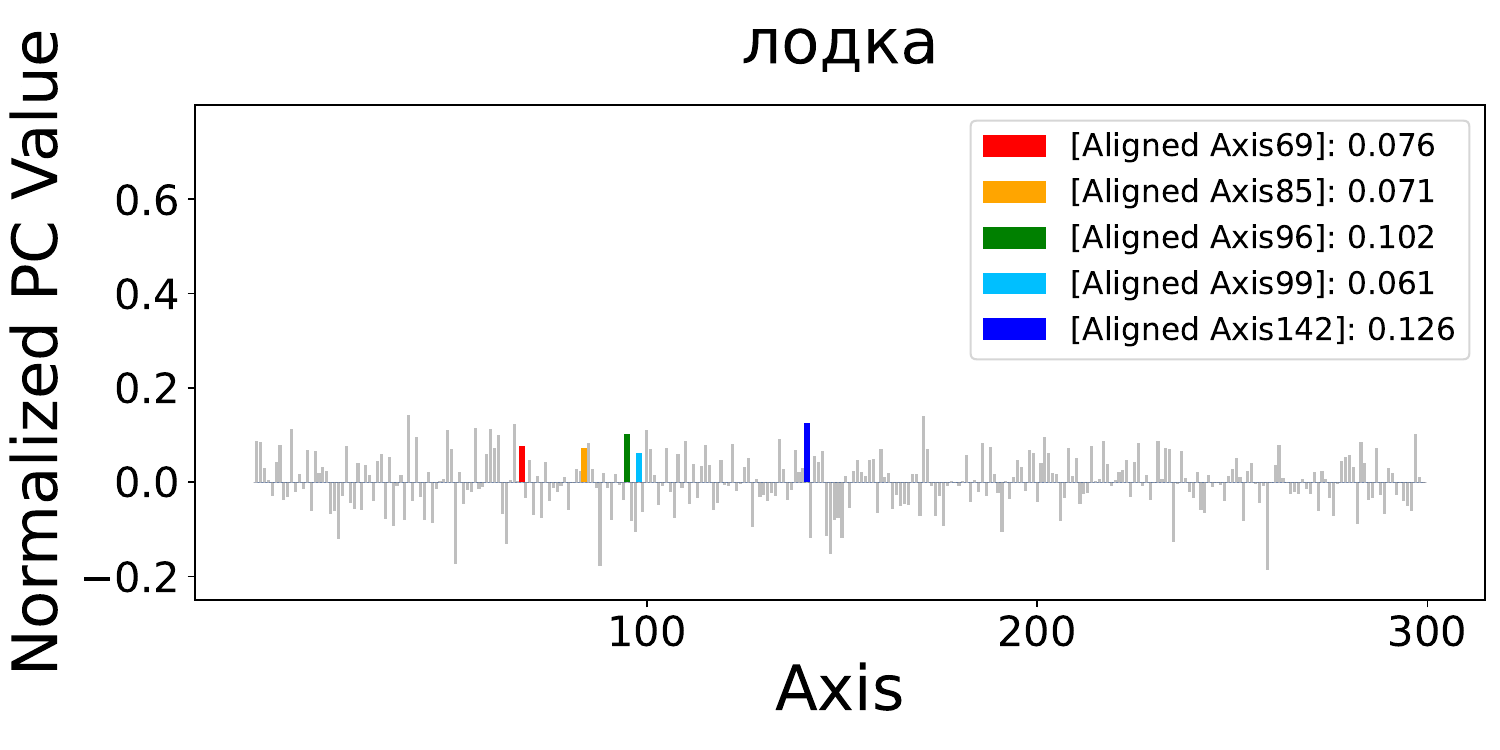}
\caption{Russian}
\label{fig:ru_pca}
\end{subfigure}\hfill
\begin{subfigure}{0.33\linewidth}
\includegraphics[width=\linewidth]{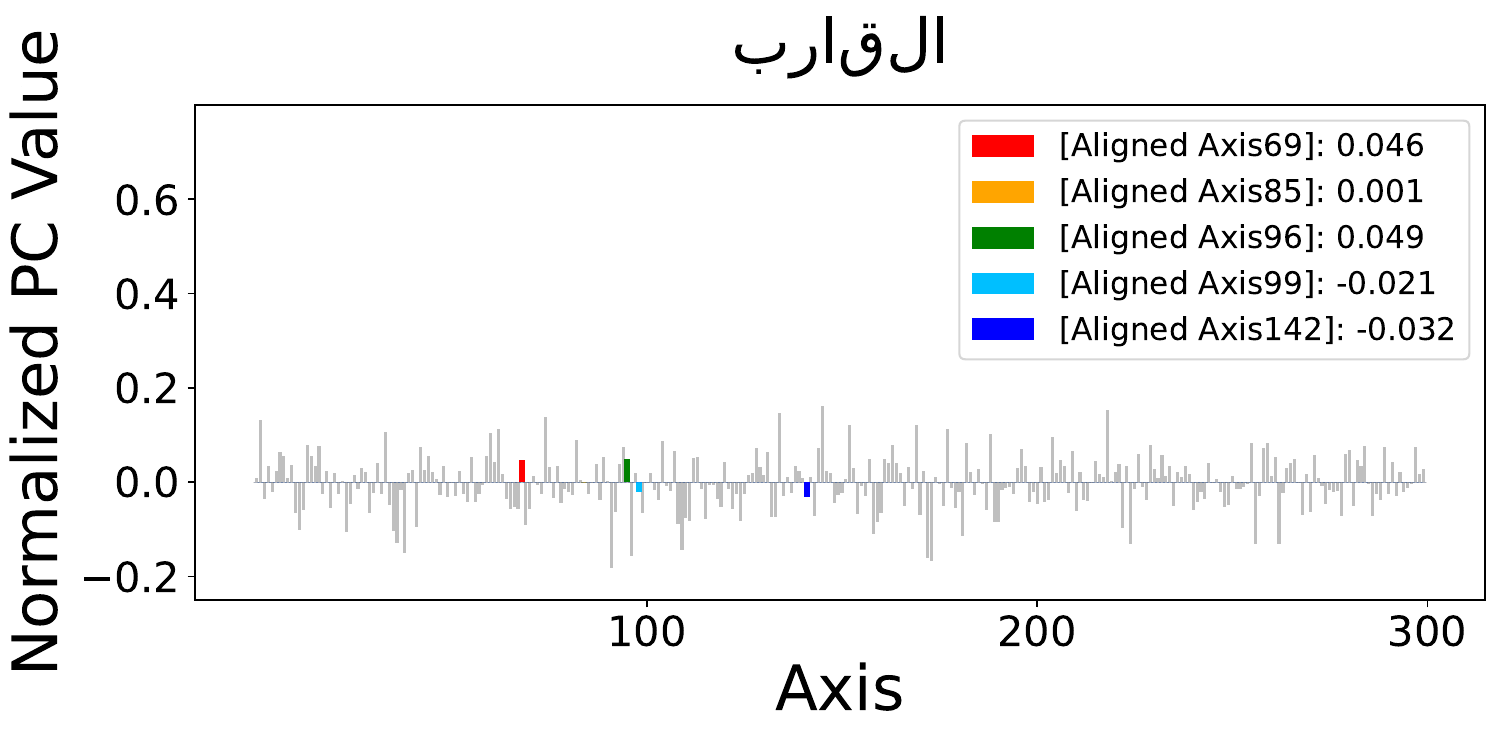}
\caption{Arabic}
\label{fig:ar_pca}
\end{subfigure}
\par\smallskip
\begin{subfigure}{0.33\linewidth}
\includegraphics[width=\linewidth]{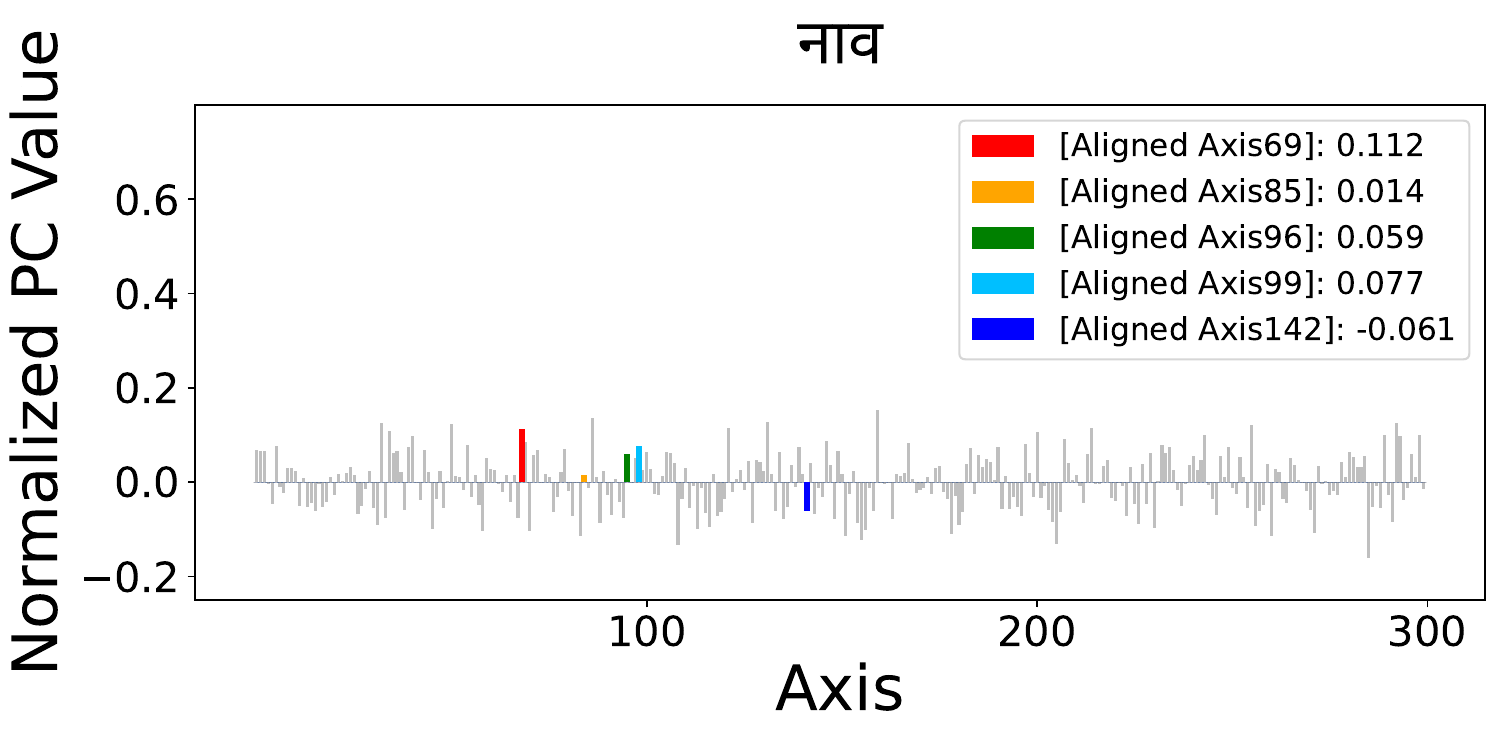}
\caption{Hindi}
\label{fig:hi_pca}
\end{subfigure}\hfill
\begin{subfigure}{0.33\linewidth}
\includegraphics[width=\linewidth]{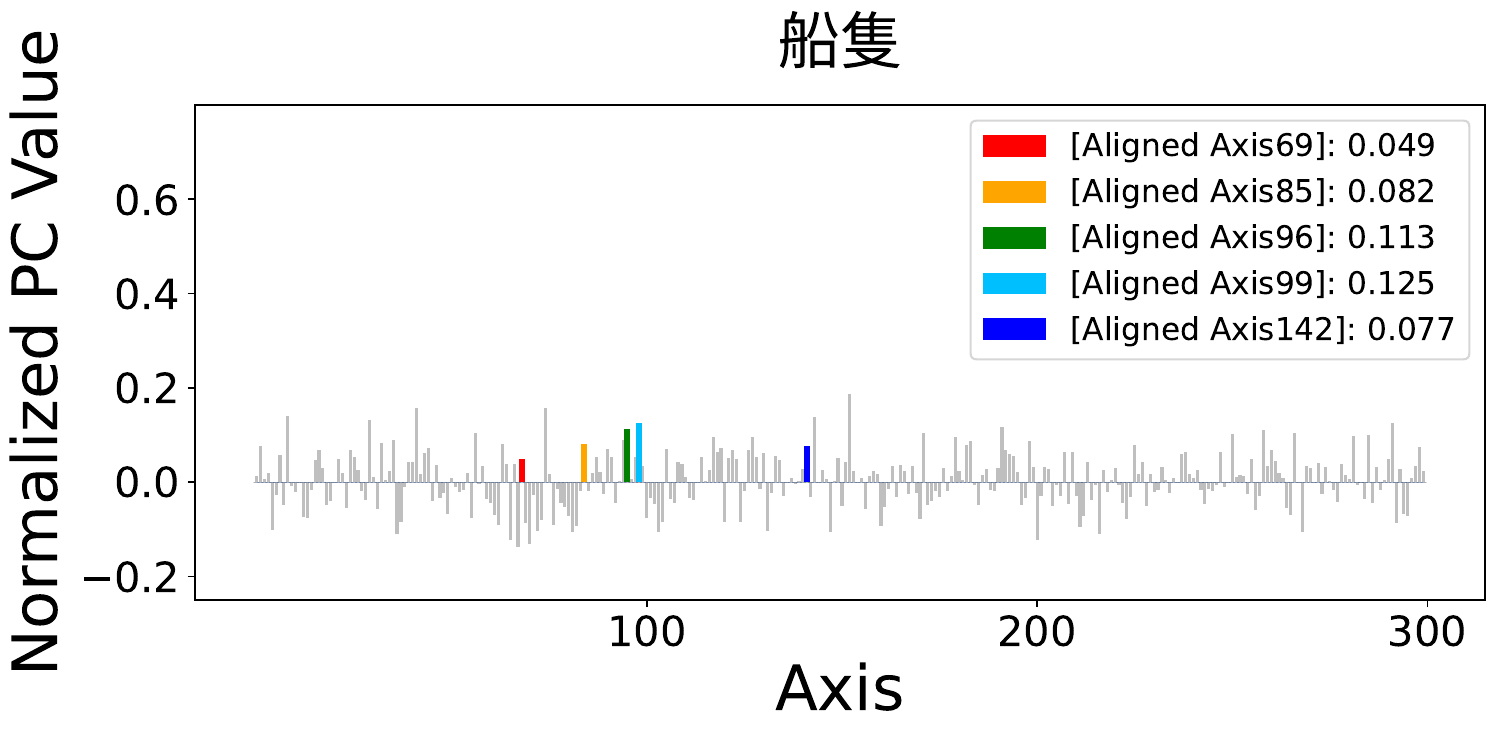}
\caption{Chinese}
\label{fig:zh_pca}
\end{subfigure}\hfill
\begin{subfigure}{0.33\linewidth}
\includegraphics[width=\linewidth]{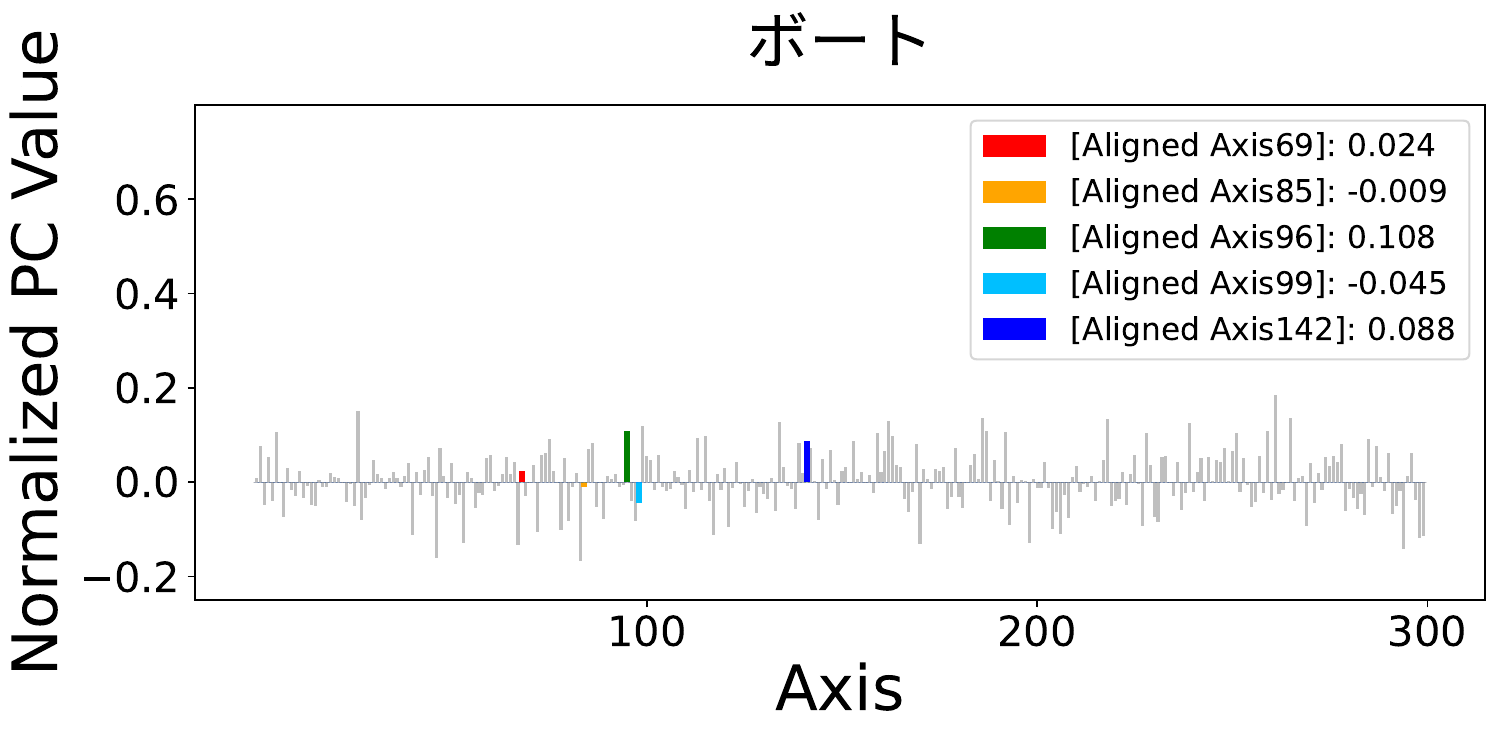}
\caption{Japanese}
\label{fig:ja_pca}
\end{subfigure}
\caption{
For \textit{boat} and its translations, the normalized PCA-transformed fastText embeddings are shown as bar graphs.
These axes are aligned by the correlation coefficients between English and the other languages.
The axes of the top 5 component values in English are highlighted with their meanings.
The component values of these axes are shown in the bar graphs for each language.
See Table~\ref{tab:boat-topwords} for the top 10 words of these axes.
}
\label{fig:pca-lang7}
\end{figure*}

As an example, we analyzed the embedding of \textit{boat} and compared it with those of its translations in Spanish, Russian, Arabic, Hindi, Chinese, and Japanese\footnote{We chose \textit{boat} as the example word, which is the top word of the second most correlated axis. Note that while the meaning of the most correlated axis is \textit{[first name]}, first names such as \textit{mike} are the same across languages such as Spanish.}. 
Figures~\ref{fig:ica-lang7} and~\ref{fig:pca-lang7} show the bar graphs of the normalized ICA-transformed and PCA-transformed embeddings, respectively, for these languages. 
Table~\ref{tab:boat-topwords} shows the top 10 words of the axes of the top 5 component values in each embedding of \textit{boat}\footnote{``海'' in the top words of the second axis of the normalized ICA-transformed embeddings is the Chinese character for \textit{sea}.}.
These results show that while the semantic component of \textit{[ship-and-sea]} is the largest for all normalized ICA-transformed embeddings in Fig.~\ref{fig:ica-lang7}, there is no such semantic component for the normalized PCA-transformed embeddings in Fig.~\ref{fig:pca-lang7}.

\renewcommand{\arraystretch}{1.2}
\begin{table*}[p]
\tiny
\centering
\begin{tabular}{@{\hspace{0.75em}}r@{\hspace{0.75em}}|@{\hspace{0.75em}}c@{\hspace{0.75em}}c@{\hspace{0.75em}}c@{\hspace{0.75em}}c@{\hspace{0.75em}}c@{\hspace{0.75em}}c@{\hspace{0.75em}}c@{\hspace{0.75em}}c@{\hspace{0.75em}}c@{\hspace{0.75em}}c@{\hspace{0.75em}}|@{\hspace{0.75em}}c@{\hspace{0.75em}}}
\toprule
Axis & Top1 & Top2 & Top3 & Top4 & Top5 & Top6 & Top7 & Top8 & Top9 & Top10 & Meaning\\
\midrule
\colora{61} & \textit{naval\_2} & \textit{sea\_1} & \textit{marine\_2} & \textit{vessels\_2} & \textit{sea\_2} & \textit{sea\_6} & \textit{vessel\_0} & \textit{naval\_1} & \textit{ships\_1} & \textit{ship\_2} & \colora{\textit{[sea]}} \\
\colorb{207} & \textit{australian\_1} & \textit{australian\_2} & \textit{new\_1} & \textit{wales\_0} & \textit{australia\_0} & \textit{sydney\_1} & \textit{south\_0} & \textit{queensland\_0} & \textit{sydney\_0} & \textit{australia\_3} & \colorb{\textit{[australia]}} \\
\colorc{250} & \textit{neighborhood\_3} & \textit{street\_12} & \textit{drag\_0} & \textit{of\_209} & \textit{central\_7} & \textit{district\_12} & \textit{street\_29} & \textit{street\_8} & \textit{heart\_5} & \textit{in\_680} & \colorc{\textit{[district]}} \\
\colord{303} & \textit{curb\_2} & \textit{prevent\_0} & \textit{avoid\_0} & \textit{reduce\_12} & \textit{reducing\_1} & \textit{\#\#d\_39} & \textit{reduce\_3} & \textit{control\_2} & \textit{prevent\_7} & \textit{prevent\_10} & \colord{\textit{[control]}} \\
\colore{338} & \textit{makes\_2} & \textit{allowed\_7} & \textit{forcing\_1} & \textit{bring\_11} & \textit{triggered\_1} & \textit{lead\_2} & \textit{illustrated\_0} & \textit{make\_65} & \textit{triggered\_2} & \textit{force\_8} & \colore{\textit{[causative verbs]}} \\
\colorf{344} & \textit{the\_1179} & \textit{the\_1976} & \textit{of\_1927} & \textit{were\_41} & \textit{were\_89} & \textit{the\_1593} & \textit{also\_64} & \textit{the\_1180} & \textit{been\_154} & \textit{been\_81} & \colorf{\textit{[numbers]}} \\
\colorg{354} & \textit{agency\_17} & \textit{agency\_1} & \textit{strategy\_2} & \textit{company\_59} & \textit{country\_7} & \textit{group\_41} & \textit{law\_7} & \textit{charity\_1} & \textit{company\_13} & \textit{pact\_0} & \colorg{\textit{[orgnization]}} \\
\colorh{521} & \textit{bail\_2} & \textit{bail\_8} & \textit{pay\_23} & \textit{walk\_0} & \textit{\#\#avi\_1} & \textit{roll\_0} & \textit{walk\_5} & \textit{cop\_1} & \textit{go\_28} & \textit{turn\_5} & \colorh{\textit{[action]}} \\
\colori{558} & \textit{ground\_10} & \textit{side\_2} & \textit{front\_10} & \textit{side\_18} & \textit{corner\_2} & \textit{scene\_3} & \textit{sides\_6} & \textit{trail\_0} & \textit{hand\_8} & \textit{hand\_19} & \colori{\textit{[location]}} \\
\colorj{572} & \textit{more\_167} & \textit{a\_244} & \textit{\#\#some\_0} & \textit{with\_434} & \textit{and\_1493} & \textit{good\_38} & \textit{and\_246} & \textit{with\_318} & \textit{more\_148} & \textit{the\_2200} & \colorj{\textit{[linking words]}} \\
\bottomrule
\end{tabular}
\caption{
For the normalized ICA-transformed BERT embeddings of \textit{shore\_0}, \textit{shore\_1}, and \textit{shore\_2} in Fig.~\ref{fig:shore_examples}, the axes of the top 4 component values for each embedding are focused.
The number of axes is 10, excluding duplicates, and the top 10 words of the axes are shown.
}
\label{tab:bert-topwords}
\end{table*}
\renewcommand{\arraystretch}{1.0}

\begin{table*}[p]
\tiny
\centering
\begin{tabular}{l|l}
\toprule
\textit{shore\_0} & Last month , the two companies sliced their dividends and sold billions of dollars of special stock to raise capital and {\color{red}\textit{shore}} up their finances .\\
& \\
\textit{shore\_1} & Working for a Sydney newspaper , my daughter covered a dreadful 1994 fire where , on one of the suburban streets of Sydney 's North {\color{red}\textit{Shore}} , \\
& the fire jumped the road and , for some terrifying seconds , took all the oxygen with it . \\
& \\
\textit{shore\_2} & Coastguards from Clevedon and Weston searched the {\color{red}\textit{shore}} while two lifeboats and two helicopters were also involved . \\
\bottomrule
\end{tabular}
\caption{
Sentences for \textit{shore\_0}, \textit{shore\_1}, and \textit{shore\_2} in Fig.~\ref{fig:shore_examples}. Note that \textit{shore\_0} is a verb, while \textit{shore\_1} and \textit{shore\_2} are nouns.
}
\label{tab:shore_sentences}
\end{table*}

\renewcommand{\arraystretch}{1.2}
\begin{table*}[p]
\tiny
\centering
\begin{tabular}{@{\hspace{0.75em}}c@{\hspace{0.75em}}|@{\hspace{0.75em}}r@{\hspace{0.75em}}|@{\hspace{0.75em}}c@{\hspace{0.75em}}c@{\hspace{0.75em}}c@{\hspace{0.75em}}c@{\hspace{0.75em}}c@{\hspace{0.75em}}c@{\hspace{0.75em}}c@{\hspace{0.75em}}c@{\hspace{0.75em}}c@{\hspace{0.75em}}c@{\hspace{0.75em}}|@{\hspace{0.75em}}c@{\hspace{0.75em}}}
\toprule
 & Axis & Top1 & Top2 & Top3 & Top4 & Top5 & Top6 & Top7 & Top8 & Top9 & Top10 & Meaning\\
\midrule
\multirow{5}{*}{\rotatebox{90}{Normalized ICA}}
 & {\color{red}2} & \textit{boat} & \textit{sailing} & \textit{sail} & \textit{ship} & \textit{boats} & \textit{sea} & \textit{ships} & \textit{海} & \textit{open-sea} & \textit{ocean} & {\color{red}\textit{[ship-and-sea]}} \\
 & {\color{orange}17} & \textit{car.} & \textit{car} & \textit{bmw} & \textit{4-door} & \textit{car--} & \textit{v-6} & \textit{2-dr} & \textit{u.s.-market} & \textit{car--and} & \textit{2-door} & {\color{orange}\textit{[cars]}} \\
 & {\color{green}36} & \textit{water} & \textit{rivers} & \textit{reservoir} & \textit{water--the} & \textit{river-water} & \textit{water.} & \textit{water--} & \textit{black-water} & \textit{basin} & \textit{de-water} & {\color{green}\textit{[water]}} \\
 & {\color{cyan}129} & \textit{12-man} & \textit{five-man} & \textit{five-person} & \textit{six-member} & \textit{seven-man} & \textit{14-member} & \textit{12-person} & \textit{seven-person} & \textit{12-member} & \textit{three-person} & {\color{cyan}\textit{[multiple people]}} \\
 & {\color{blue}131} & \textit{race} & \textit{races} & \textit{racing} & \textit{race.} & \textit{racer} & \textit{race.-} & \textit{.race} & \textit{rider} & \textit{laps} & \textit{race-like} & {\color{blue}\textit{[races]}} \\
\midrule
\multirow{5}{*}{\rotatebox{90}{Normalized PCA}}
 & {\color{red}69} & \textit{bit-field} & \textit{torn} & \textit{out-of-round} & \textit{unused} & \textit{final-} & \textit{3-space} & \textit{bad.2.} & \textit{cup} & \textit{too-large} & \textit{.language} & {\color{red}\textit{[Aligned Axis69]}} \\
 & {\color{orange}85} & \textit{2-the} & \textit{name-called} & \textit{accusations} & \textit{t-head} & \textit{down.1.} & \textit{1-the} & \textit{relata} & \textit{attacked} & \textit{flew} & \textit{two-place} & {\color{orange}\textit{[Aligned Axis85]}} \\
 & {\color{green}96} & \textit{government-run} & \textit{trading} & \textit{state-run} & \textit{military-run} & \textit{kids-only} & \textit{trade} & \textit{floating} & \textit{e-a} & \textit{sirven} & \textit{trade.} & {\color{green}\textit{[Aligned Axis96]}} \\
 & {\color{cyan}99} & \textit{white-red-white} & \textit{-green} & \textit{white-blue} & \textit{red-green} & \textit{white-red} & \textit{.hair} & \textit{color} & \textit{'k} & \textit{voz} & \textit{poles} & {\color{cyan}\textit{[Aligned Axis99]}} \\
 & {\color{blue}142} & \textit{business-process} & \textit{source-to-pay} & \textit{time-to-market} & \textit{menudo} & \textit{s-u} & \textit{5-6-11} & \textit{pulse} & \textit{cost} & \textit{news-} & \textit{time-to-value} & {\color{blue}\textit{[Aligned Axis142]}} \\
\bottomrule
\end{tabular}
\caption{
For the normalized ICA and PCA transformed fastText embeddings of \textit{boat} in Figs~\ref{fig:ica-lang7} and~\ref{fig:pca-lang7}, the axes of the top 5 component values are focused and their top 10 words are shown.
For PCA, since it is difficult to interpret the meanings of the axes, they are simply labeled such as \textit{[Aligned Axis69]}. 
Similar to GloVe, the meanings of the axes of the ICA-transformed fastText embeddings are interpretable.
}
\label{tab:boat-topwords}
\end{table*}
\renewcommand{\arraystretch}{1.0}

\end{document}